\documentclass{article}

\usepackage{times}
\usepackage{graphicx}
\usepackage{sidecap}
\usepackage{subfigure}
\usepackage[small]{caption}

\usepackage{multicol}

\usepackage{setspace}
\AtBeginDocument{
  \addtolength\abovedisplayskip{-0.25\baselineskip}
  \addtolength\belowdisplayskip{-0.25\baselineskip}
  \addtolength\abovedisplayshortskip{-0.25\baselineskip}
  \addtolength\belowdisplayshortskip{-0.25\baselineskip}
}

\usepackage{pifont}
\usepackage{xspace}
\newcommand{\circone}{\ding{172}\xspace}
\newcommand{\circtwo}{\ding{173}\xspace}
\newcommand{\circthree}{\ding{174}\xspace}

\usepackage[round]{natbib}

\usepackage{appendix}

\newcommand{\rightcomment}[1]{\(\triangleright\) {\small \it #1}}
\newcommand{\eqcomment}[1]{\tag*{\rightcomment{#1}\quad\addtocounter{equation}{1}(\theequation)}}
\usepackage{suffix}
\WithSuffix\newcommand\eqcomment*[1]{\tag*{\rightcomment{#1}}}

\usepackage{algorithm}

\usepackage[hidelinks]{hyperref}

\usepackage[final]{nips_2017}

\usepackage[noend]{algpseudocode}

\algrenewcommand{\algorithmiccomment}[1]{\hfill \rightcomment{#1}}
\algnewcommand{\LineComment}[1]{\State \rightcomment{#1}}
\makeatletter

\newcommand{\algmargin}{\the\ALG@thistlm}
\makeatother
\algnewcommand{\Statepar}[1]{\State\parbox[t]{\dimexpr\linewidth-\algmargin}{\strut #1\strut}}
\newcommand{\pluseq}{\mathrel{+\!\!=}}

\usepackage{xpatch}
\xapptocmd\normalsize{
 \abovedisplayskip=11pt plus 3pt minus 9pt
 \abovedisplayshortskip=0pt plus 3pt
 \belowdisplayskip=11pt plus 3pt minus 9pt
 \belowdisplayshortskip=6.5pt plus 3.5pt minus 3pt
}{}{}

\usepackage{amsmath}
\usepackage{amssymb}
\usepackage{mathtools, cuted}
\usepackage{tabularx, booktabs}
\newcolumntype{C}{>{\centering\arraybackslash}X}
\newcolumntype{R}{>{\raggedleft\arraybackslash}X}
\newcolumntype{S}{>{\raggedleft\arraybackslash\hsize=.5\hsize}X}
\usepackage{latexsym}
\usepackage{url}
\usepackage{xspace}
\usepackage{bm,array}
\usepackage{amsfonts}
\usepackage{enumitem}
\usepackage{mathtools}
\usepackage{bm}
\usepackage{esvect}
\usepackage[noabbrev,capitalize]{cleveref}
\crefname{equation}{equation}{equations}
\crefname{section}{section}{sections}
\crefname{footnote}{footnote}{footnotes}

\usepackage{bbm}
\renewcommand{\vec}[1]{{\boldsymbol{\mathbf{#1}}}}
\newcommand{\defn}[1]{\textbf{#1}}
\newcommand{\defeq}{\mathrel{\stackrel{\mbox{\tiny def}}{=}}}
\DeclareMathOperator*{\argmin}{argmin}
\DeclareMathOperator*{\argmax}{argmax}
\newcommand{\E}[2][]{\mathbb{E}_{{#1}}[#2]}
\newcommand{\Real}{\mathbb{R}}
\newcommand{\Uniform}{\mathrm{Unif}}
\newcommand{\Exp}{\mathrm{Exp}}
\renewcommand{\th}{\textsuperscript{th}\xspace}
\newcommand{\bos}{\textsc{bos}\xspace}
\newcommand{\eos}{\textsc{eos}\xspace}

\usepackage[usenames,dvipsnames,svgnames,table]{xcolor}
\usepackage[disable]{todonotes}

\newcommand{\Fixme}[2][]{\noindent}

\newcommand{\Jason}[2][]{\noindent}
\newcommand{\Hongyuan}[2][]{\noindent}

\newcommand{\cutforspace}[1]{}

\usepackage{verbatim}

\title{
The Neural Hawkes Process:
A Neurally Self-Modulating Multivariate Point Process
}

\author{
  Hongyuan Mei	 \ \ \ \ \ \ Jason Eisner \\
  Department of Computer Science, Johns Hopkins University \\
  3400 N. Charles Street, Baltimore, MD 21218 U.S.A \\
  \texttt{\{hmei,jason\}@cs.jhu.edu} \\
}

\begin{document}

\maketitle

\vspace{-18pt}
\begin{abstract}
\vspace{-6pt}
Many events occur in the world. Some event types are stochastically excited or inhibited---in the sense of having their probabilities elevated or decreased---by patterns in the sequence of previous events. Discovering such patterns can help us predict {\em which type} of event will happen next and {\em when}. We model streams of discrete events in continuous time, by constructing a \defn{neurally self-modulating multivariate point process} in which the intensities of multiple event types evolve according to a novel \defn{continuous-time LSTM}.  This generative model allows past events to influence the future in complex and realistic ways, by conditioning future event intensities on the hidden state of a recurrent neural network that has consumed the stream of past events.  Our model  has desirable qualitative properties.  It achieves competitive likelihood and predictive accuracy on real and synthetic datasets, including under missing-data conditions.
\end{abstract}

\vspace{-23pt}
\section{Introduction}
\label{sec:introduction}
\vspace{-3pt}

Some events in the world are correlated.  A single event, or a pattern of events, may help to cause or prevent future events.  We are interested in learning the distribution of sequences of events (and in future work, the causal structure of these sequences).  The ability to discover correlations among events is crucial to accurately predict the future of a sequence given its past, i.e., which events are likely to happen next and when they will happen.

We specifically focus on sequences of discrete events in continuous time (``event streams'').
Modeling such sequences seems natural and useful in many applied domains:
\begin{itemize}
\item {\em Medical events.}  Each patient has a sequence of acute incidents, doctor's visits, tests, diagnoses, and medications.  By learning from previous patients what sequences tend to look like, we could predict a new patient's future from their past.
\item {\em Consumer behavior.}  Each online consumer has a sequence of online interactions.  By modeling the distribution of sequences, we can learn purchasing patterns.  Buying cookies may temporarily depress purchases of all desserts, yet increase the probability of buying milk.\looseness=-1
\item {\em ``Quantified self'' data.}  Some individuals use cellphone apps to record their behaviors---eating, traveling, working, sleeping, waking.  By anticipating behaviors, an app could perform helpful supportive actions, including issuing reminders and placing advance orders.
\item {\em Social media actions.}  Previous posts, shares, comments, messages, and likes by a set of users are predictive of their future actions.
\item Other event streams arise in {\em news}, {\em animal behavior}, {\em dialogue}, {\em music}, etc.
\end{itemize}

A basic model for event streams is the \defn{Poisson process}~\citep{palm-43}, which assumes that events occur independently of one another.  In a \defn{non-homogenous Poisson process}, the (infinitesimal) probability of an event happening at time $t$ may vary with $t$, but it is still independent of other events.  A \defn{Hawkes process} \citep{hawkes-71,liniger-09-hawkes}
supposes that past events can temporarily {\em raise} the probability of future events, assuming that such excitation is \circone positive, \circtwo additive over the past events, and \circthree exponentially decaying with time.

\begin{figure}[t]
\begin{center}
\centerline{\includegraphics[width=\columnwidth]{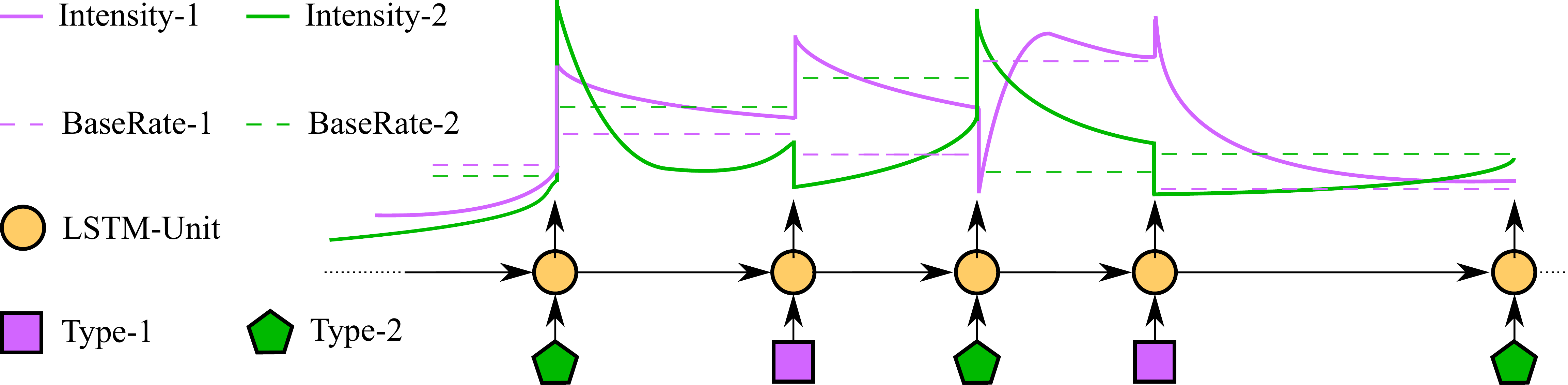}}
\caption{
	Drawing an event stream from a neural Hawkes process.
        An LSTM reads the sequence of past events (polygons) to arrive at a hidden state (orange).  That state determines the future ``intensities'' of the two types of events---that is, their time-varying instantaneous probabilities.
        The intensity functions are continuous parametric curves (solid lines) determined by the most recent LSTM state, with dashed lines showing the steady-state asymptotes that they would eventually approach.
	In this example, events of type 1 excite type 1 but inhibit type 2.  Type 2 excites itself, and excites or inhibits type 1 according to whether the count of type 2 events so far is odd or even. Those are immediate effects, shown by the sudden jumps in intensity. The events also have longer-timescale effects, shown by the shifts in the asymptotic dashed lines.
}\label{fig:event-example}
\end{center}
\end{figure}

However, real-world patterns often seem to violate these assumptions.  For example, \circone is violated if one event inhibits another rather than exciting it: cookie consumption inhibits cake consumption.  \circtwo is violated when the combined effect of past events is not additive. Examples abound: The 20th advertisement does not increase purchase rate as much as the first advertisement did, and may even drive customers away.  Market players may act based on their own complex analysis of market history.  Musical note sequences follow some intricate language model that considers melodic trajectory, rhythm, chord progressions, repetition, etc.  \circthree is violated when, for example, a past event has a delayed effect, so that the effect starts at 0 and increases sharply before decaying.

We generalize the Hawkes process by determining the event intensities (instantaneous probabilities) from the hidden state of a recurrent neural network.  This state is a deterministic function of the past history.  It plays the same role as the state of a deterministic finite-state automaton.  However, the recurrent network enjoys a continuous and infinite state space (a high-dimensional Euclidean space), as well as a learned transition function.  In our network design, the state is updated discontinuously with each successive event occurrence and also evolves continuously as time elapses between events.

Our main motivation is that our model can capture effects that the Hawkes process misses.  The combined effect of past events on future events can now be superadditive, subadditive, or even subtractive, and can depend on the sequential ordering of the past events.  Recurrent neural networks already capture other kinds of complex sequential dependencies when applied to language modeling---that is, generative modeling of linguistic word sequences, which are governed by syntax, semantics, and habitual usage~\citep{mikolov-10-rnnlm,sundermeyer-12-lstm,karpathy-15-visualize}. We wish to extend their success \citep{chelba-13-billion} to sequences of events in {\em continuous} time.

\label{sec:missing}
Another motivation for a more expressive model than the Hawkes process is to cope with missing data.
Even in a domain where Hawkes might be appropriate, it is hard to apply Hawkes when sequences are only partially observed.  Real datasets may {\em systematically} omit some types of events (e.g., illegal drug use, or offline purchases) which, in the true generative model, would have a strong influence on the future.  They may also have {\em stochastically} missing data, where the missingness mechanism---the probability that an event is not recorded---can be complex and data-dependent (MNAR).  In this setting, we can fit our model directly to the observation sequences, and use it to predict observation sequences that were generated in the same way (using the same complete-data distribution and the same missingness mechanism).  Note that if one knew the true complete-data distribution---perhaps Hawkes---and the true missingness mechanism, one would optimally predict the incomplete future from the incomplete past in Bayesian fashion, by integrating over possible completions (imputing the missing events and considering their influence on the future).  Our hope is that the neural model is expressive enough that it can learn to approximate this true predictive distribution.  Its hidden state after observing the past should implicitly encode the Bayesian posterior, and its update rule for this hidden state should emulate the ``observable operator'' that updates the posterior upon each new observation.  See \cref{sec:missing_data_details} for further discussion.\looseness=-1

A final motivation is that one might wish to {\em intervene} in a medical, economic, or social event stream so as to improve the future course of events.  \Cref{sec:future} discusses our plans to deploy our model family as an environment model within reinforcement learning, where an agent controls some events.

\section{Notation}
\label{sec:notation}

We are interested in constructing distributions over \defn{event streams} $(k_1,t_1), (k_2, t_2), \ldots$, where each $k_i \in \{1,2,\ldots,K\}$ is an event type and $0 < t_1 < t_2 < \cdots$ are times of occurrence.\footnote{\label{fn:immediate}More generally, one could allow $0 \leq t_1 \leq t_2 \leq \cdots$, where $t_i$ is a \defn{immediate event} if $t_{i-1}=t_i$ and a \defn{delayed event} if $t_{i-1} < t_i$.  It is not too difficult to extend our model to assign positive probability to immediate events, but we will disallow them here for simplicity.}  That is, there are $K$ types of events, tokens of which are observed to occur in continuous time.

For any distribution $P$ in our proposed family, an event stream is almost surely infinite.  However, when we observe the process only during a time interval $[0,T]$, the number $I$ of observed events is almost surely finite.  The {\em log-likelihood} $\ell$ of the model $P$ given these $I$ observations is
\begin{equation}\label{eqn:loglik-orig}
    \Big( \sum_{i=1}^I \log P\!\left( \left(k_i,t_i\right) \mid \mathcal{H}_{i}, \Delta t_i \right) \Big) + \log P(t_{I+1} > T \mid \mathcal{H}_I)
\end{equation}
where the \defn{history} $\mathcal{H}_{i}$ is the prefix sequence $(k_1,t_1)$, $(k_2, t_2), \ldots, (k_{i-1}, t_{i-1})$, and $\Delta t_i \defeq t_i - t_{i-1}$, and
$P((k_i, t_i) \mid \mathcal{H}_i, \Delta t_i)\,dt$ is the probability that the {\em next} event occurs at time $t_i$ and has type $k_i$.\looseness=-1

Throughout the paper, the subscript $i$ usually denotes quantities that affect the distribution of the next event $(k_i,t_i)$. These quantities depend only on the history ${\cal H}_i$.

We use (lowercase) Greek letters for parameters related to the classical Hawkes process,
and Roman letters for other quantities, including hidden states and affine transformation parameters.
We denote vectors by bold lowercase letters such as $\vec{s}$ and $\vec{\mu}$, and matrices by bold capital Roman letters such as $\vec{U}$. Subscripted bold letters denote distinct vectors or matrices (e.g., $\vec{w}_k$).
Scalar quantities, including vector and matrix elements such as $s_k$ and $\alpha_{j,k}$, are written without bold. Capitalized scalars represent upper limits on lowercase scalars, e.g., $1 \leq k \leq K$.  Function symbols are notated like their return type.
All $\Real \rightarrow \Real$ functions are extended to apply elementwise to vectors and matrices.\looseness=-1

\section{The Model}
\label{sec:model}

In this section, we first review Hawkes processes, and then introduce our model one step at a time.

Formally, generative models of event streams are \defn{multivariate point processes}.  A (temporal) point process is a probability distribution over $\{0,1\}$-valued functions on a given time interval (for us, $[0,\infty)$).  A multivariate point process is formally a distribution over $K$-tuples of such functions.  The $k$\th function indicates the times at which events of type $k$ occurred, by taking value 1 at those times.

\subsection{Hawkes Process: A Self-Exciting Multivariate Point Process (SE-MPP)}
\label{sec:poisson_process}
\label{sec:hawkes_process}
\label{sec:SE-MPP}

A basic model of event streams is the \defn{non-homogeneous multivariate Poisson process}.  It assumes that an event of type $k$ occurs at time $t$---more precisely, in the infinitesimally wide interval $[t,t+dt)$---with probability $\lambda_k(t) dt$.  The value $\lambda_k(t) \geq 0$ can be regarded as a rate per unit time, just like the parameter $\lambda$ of an ordinary Poisson process.  $\lambda_k$ is known as the \defn{intensity function}, and the total intensity of all event types is given by $\lambda(t) = \sum_{k=1}^K \lambda_k(t)$.

A well-known generalization that captures interactions is the \defn{self-exciting multivariate point process (SE-MPP)},
or \defn{Hawkes process} \citep{hawkes-71,liniger-09-hawkes}, in which past events $h$ from the history conspire to {\em raise} the intensity of each type of event.
Such excitation is positive, additive over the past events, and exponentially decaying with time:
\begin{align}\label{eqn:hawkes}
    \lambda_k(t)
    &= \mu_k + \sum_{h: t_h < t} \alpha_{k_h,k} \exp (-\delta_{k_h,k} (t-t_h) )
\end{align}
where \mbox{$\mu_{k} \geq 0$} is the base intensity of event type $k$, \mbox{$\alpha_{j,k} \geq 0$} is the degree to which an event of type $j$ initially excites type $k$, and \mbox{$\delta_{j,k} > 0$} is the decay rate of that excitation.
When an event occurs, all intensities are elevated to various degrees, but then will decay toward their base rates $\vec{\mu}$.

\subsection{Self-Modulating Multivariate Point Processes}
\label{sec:self_modulate}

The positivity constraints in the Hawkes process
limit its expressivity.
First, the positive interaction parameters $\alpha_{j,k}$
fail to capture inhibition effects, in which past events {\em reduce} the intensity of future events. Second, the positive base rates $\vec{\mu}$ fail to capture the inherent inertia of some events, which are unlikely until their cumulative excitation by past events crosses some threshold.  To remove such limitations, we introduce two {\em self-modulating} models.  Here the intensities of future events are stochastically {\em modulated} by the past history, where the term ``modulation'' is meant to encompass both excitation and inhibition.  The intensity $\lambda_k(t)$ can even fluctuate non-monotonically between successive events, because the competing excitatory and inhibitory influences may decay at different rates.\looseness=-1

\subsubsection{Hawkes Process with Inhibition: A Decomposable Self-Modulating MPP (D-SM-MPP)}
\label{sec:D-SM-MPP}

Our first move is to enrich the Hawkes model's expressiveness while still maintaining its decomposable structure.
We relax the positivity constraints on $\alpha_{j,k}$ and $\mu_k$, allowing them to range over $\Real$, which allows {\em inhibition} ($\alpha_{j,k} < 0$) and {\em inertia} ($\mu_k < 0$).  However, the resulting total activation could now be negative.  We therefore pass it through a non-linear \defn{transfer function} $f_k: \Real \rightarrow \Real_+$ to obtain a positive intensity function as required:

\begin{subequations} \label{eqn:hawkes_inhib}
\begin{minipage}[t]{0.29\textwidth}
\vspace{-20pt}
\begin{align}
	\label{eqn:hawkes_inhib_a}
    \lambda_k(t) &= f_k(\tilde{\lambda}_k(t))
\end{align}
\end{minipage}
\hfill
\begin{minipage}[t]{0.69\textwidth}
\vspace{-20pt}
\begin{align}
	\label{eqn:hawkes_inhib_b}
    \tilde{\lambda}_k(t) &= \mu_k + \sum_{h: t_h < t} \alpha_{k_h,k} \exp (-\delta_{k_h,k} (t-t_h) )
\end{align}
\end{minipage}
\end{subequations}
As $t$ increases between events, the intensity $\lambda_k(t)$ may both rise and fall, but eventually approaches the base rate $f(\mu_k + 0)$,
as the influence of each previous event still decays toward 0 at a rate $\delta_{j,k} > 0$.\looseness=-1

What non-linear function $f_k$ should we use? The ReLU function $f(x) = \max(x,0)$ is not strictly positive as required.
A better choice is the scaled ``softplus'' function \mbox{$f(x)=s \log(1+\exp(x/s))$}, which approaches ReLU as $s \rightarrow 0$. We learn a separate scale parameter $s_k$ for each event type $k$, which adapts to the rate of that type.
So we instantiate \eqref{eqn:hawkes_inhib_a} as
$\lambda_k(t) = f_k(\tilde{\lambda}_k(t)) = s_k \log ( 1 + \exp ( \tilde{\lambda}_k(t) / s_k ) )$.
\Cref{sec:transfer} graphs this and motivates the ``softness'' and the scale parameter.

\subsubsection{Neural Hawkes Process: A Neurally Self-Modulating MPP (N-SM-MPP)}
\label{sec:neural_hawkes_process}
\label{sec:N-SM-MPP}
\label{sec:params}
\label{sec:opengates}

Our second move removes the restriction that the past events have independent, additive influence on $\tilde{\lambda}_k(t)$.  Rather than predict $\tilde{\lambda}_k(t)$ as a simple summation \eqref{eqn:hawkes_inhib_b}, we now use a recurrent neural network.  This allows learning a complex dependence of the intensities on the number, order, and timing of past events.  We refer to our model as a \defn{neural Hawkes process}.

Just as before, each event type $k$ has an time-varying intensity $\lambda_k(t)$, which jumps discontinuously at each new event, and then drifts continuously toward a baseline intensity.  In the new process, however, these dynamics are controlled by a hidden state vector $\vec{h}(t) \in (-1,1)^D$, which in turn depends on a vector $\vec{c}(t) \in \Real^D$ of memory cells in a \defn{continuous-time LSTM}.\footnote{We use one-layer LSTMs with $D$ hidden units in our present experiments, but a natural extension is to use multi-layer (``deep'') LSTMs \citep{graves-13-hybrid}, in which case $\vec{h}(t)$ is the hidden state of the top layer.}  This novel recurrent neural network architecture  is inspired by the familiar discrete-time LSTM~\citep{hochreiter-97-lstm,graves-12}.  The difference is that in the continuous interval following an event, each memory cell $c$ {\em exponentially decays} at some rate $\delta$ toward some steady-state value $\bar{c}$.

At each time $t > 0$, we obtain the intensity $\lambda_k(t)$ by \eqref{eqn:hawkes_mod_a},
where \eqref{eqn:hawkes_mod_b} shows how the hidden states $\vec{h}(t)$ are continually obtained from the memory cells $\vec{c}(t)$ as the cells decay:
\begin{subequations} \label{eqn:hawkes_mod}
\begin{minipage}[t]{0.39\textwidth}
\vspace{-10pt}
\begin{align}
	\label{eqn:hawkes_mod_a}
    \lambda_k(t) &= f_{k}(\vec{w}_{k}^{\top} \vec{h}(t))
\end{align}
\end{minipage}
\hfill
\begin{minipage}[t]{0.59\textwidth}
\vspace{-10pt}
\begin{align}
	\label{eqn:hawkes_mod_b}
  \vec{h}(t) = \vec{o}_i \odot (2\sigma(2\vec{c}(t))-1) \text{ for } t \in (t_{i-1}, t_{i}]
\end{align}
\end{minipage}
\end{subequations}

This says that on the interval $(t_{i-1}, t_{i}]$---in other words, after event \mbox{$i\!-\!1$} up until event $i$ occurs at some time $t_i$---the $\vec{h}(t)$ defined by \cref{eqn:hawkes_mod_b} determines the intensity functions via \cref{eqn:hawkes_mod_a}.  So for $t$ in this interval, according to the model, $\vec{h}(t)$ is a sufficient statistic of the history $({\cal H}_i, t-t_{i-1})$ with respect to future events (see \cref{eqn:loglik-orig}).
$\vec{h}(t)$ is analogous to $\vec{h}_i$ in an LSTM language model~\citep{mikolov-10-rnnlm}, which summarizes the past event sequence $k_1,\ldots,k_{i-1}$.  But in our decay architecture, it will also reflect the interarrival times $t_1-0, t_2-t_1, \ldots, t_{i-1}-t_{i-2}, t-t_{i-1}$.
This interval $(t_{i-1},t_i]$ ends when the next event $k_i$ stochastically occurs at some time $t_i$.  At this point, the continuous-time LSTM reads $(k_i,t_i)$ and updates the current (decayed) hidden cells $\vec{c}(t)$ to new initial values $\vec{c}_{i+1}$, based on the current (decayed) hidden state $\vec{h}(t_i)$.

How does the continuous-time LSTM make those updates?  Other than depending on decayed values, the update formulas resemble the discrete-time case:\footnote{The upright-font subscripts $\mathrm{i}$, $\mathrm{f}$, $\mathrm{z}$ and $\mathrm{o}$ are not variables, but constant labels that distinguish different $\vec{W}$, $\vec{U}$ and $\vec{d}$ tensors. The $\bar{\vec{f}}$ and $\bar{\vec{\imath}}$ in~\cref{eqn:ct_cell_target} are defined analogously to $\vec{f}$ and $\vec{i}$ but with different weights.}

\vspace{-10pt}
\minipage[t]{0.49\textwidth}
\begin{subequations} \label{eqn:ct_lstm}
\begin{align}
	\vec{i}_{i+1}
	&\leftarrow \sigma \left( \vec{W}_{\mathrm{i}} \vec{k}_{i} + \vec{U}_{\mathrm{i}} \vec{h}(t_{i}) + \vec{d}_{\mathrm{i}}  \right) \\
	\vec{f}_{i+1}
	&\leftarrow \sigma \left( \vec{W}_{\mathrm{f}} \vec{k}_{i} + \vec{U}_{\mathrm{f}} \vec{h}(t_{i}) + \vec{d}_{\mathrm{f}}  \right) \\
	\vec{z}_{i+1}
	&\leftarrow 2 \sigma \left( \vec{W}_{\mathrm{z}} \vec{k}_{i} + \vec{U}_{\mathrm{z}} \vec{h}(t_{i}) + \vec{d}_{\mathrm{z}}  \right) - 1 \\
	\vec{o}_{i+1}
	&\leftarrow \sigma \left( \vec{W}_{\mathrm{o}} \vec{k}_{i} + \vec{U}_{\mathrm{o}} \vec{h}(t_{i}) + \vec{d}_{\mathrm{o}}  \right) \label{eqn:ot}
\end{align}
\end{subequations}
\endminipage
\hfill
\minipage[t]{0.45\textwidth}
\begin{subequations} \label{eqn:ct_cell}
\begin{align}
\vec{c}_{i+1}
&\leftarrow \vec{f}_{i+1} \odot \vec{c}(t_{i}) + \vec{i}_{i+1} \odot \vec{z}_{i+1}\label{eqn:ct_cell_instant} \\
\bar{\vec{c}}_{i+1}
&\leftarrow \bar{\vec{f}}_{i+1} \odot \bar{\vec{c}}_{i} + \bar{\vec{\imath}}_{i+1} \odot \vec{z}_{i+1} \label{eqn:ct_cell_target} \\
\vec{\delta}_{i+1}
&\leftarrow f \left( \vec{W}_{\mathrm{d}} \vec{k}_{i} + \vec{U}_{\mathrm{d}} \vec{h}(t_{i}) + \vec{d}_{\mathrm{d}} \right) \label{eqn:ct_cell_rate}
\end{align}
\end{subequations}
\endminipage
\vspace{10pt}

The vector $\vec{k}_i \in \{ 0,1 \}^K$ is the $i$\th input: a one-hot encoding of the new event $k_i$, with non-zero value only at the entry indexed by $k_i$.
The above formulas will make a discrete update to the LSTM state.  They resemble the discrete-time LSTM, but there are two differences.  First, the updates do not depend on the ``previous'' hidden state from just after time $t_{i-1}$, but rather its value $\vec{h}(t_i)$ at time $t_i$, after it has decayed for a period of $t_i - t_{i-1}$.  Second, \crefrange{eqn:ct_cell_target}{eqn:ct_cell_rate} are new.  They define how in future, as $t > t_i$ increases, the elements of $\vec{c}(t)$ will continue to deterministically decay (at different rates) from $\vec{c}_{i+1}$ toward targets $\bar{\vec{c}}_{i+1}$.  Specifically, $\vec{c}(t)$ is given by \eqref{eqn:c_decay}, which continues to control $\vec{h}(t)$ and thus $\lambda_k(t)$ (via \eqref{eqn:hawkes_mod}, except that $i$ has now increased by 1).
\begin{align}
\vec{c}(t) &\defeq \bar{\vec{c}}_{i+1} + \left( \vec{c}_{i+1} - \bar{\vec{c}}_{i+1} \right) \exp \left( -\vec{\delta}_{i+1} \left( t - t_i \right) \right) \text{ for } t \in (t_i,t_{i+1}]
\label{eqn:c_decay}
\end{align}

In short, not only does \eqref{eqn:ct_cell_instant} define the usual cell values $\vec{c}_{i+1}$, but \cref{eqn:c_decay} defines $\vec{c}(t)$ on $\Real_{>0}$.  On the interval $(t_i,t_{i+1}]$,
$\vec{c}(t)$ follows an exponential curve that begins at $\vec{c}_{i+1}$ (in the sense that $\lim_{t \rightarrow t_{i}^+} \vec{c}(t) = \vec{c}_{i+1}$) and decays toward $\bar{\vec{c}}_{i+1}$ (which it would approach as $t \rightarrow \infty$, if extrapolated).

A schematic example is shown in \cref{fig:event-example}.  As in the previous models, $\lambda_k(t)$ drifts deterministically between events toward some base rate.  But the neural version is different in three ways: \circone The base rate is not a constant $\mu_k$, but shifts upon each event.\footnote{\Cref{eqn:hawkes_mod_b,eqn:c_decay} imply that after event $i-1$, the base rate jumps to $f_k(\vec{w}^\top(\vec{o}_i \odot (2\sigma(2\bar{\vec{c}}_{i})-1)))$.} \circtwo The drift can be non-monotonic, because the excitatory and inhibitory influences on $\lambda_k(t)$ from different elements of $\vec{h}(t)$ may decay at different rates.  \circthree The sigmoidal transfer function means that the behavior of $\vec{h}(t)$ itself is a little more interesting than exponential decay. Suppose that $\vec{c}_i$ is very negative but increases toward a target $\bar{\vec{c}}_i > 0$. Then $\vec{h}(t)$ will stay close to $-1$ for a while and then will rapidly rise past 0. This usefully lets us model a delayed response (e.g. the last green segment in~\cref{fig:event-example}).

We point out two behaviors that are naturally captured by our LSTM's ``forget'' and ``input'' gates:
\begin{itemize}
\item if $\vec{f}_{i+1} \approx \vec{1}$ and $\vec{i}_{i+1} \approx \vec{0}$, then $\vec{c}_{i+1} \approx \vec{c}(t_{i})$.  So  $\vec{c}(t)$ and $\vec{h}(t)$ will be {\em continuous} at $t_{i}$. There is no jump due to event $i$, though the steady-state target may change.
\item if $\bar{\vec{f}}_{i+1} \approx \vec{1}$ and $\bar{\vec{\imath}}_{i+1} \approx \vec{0}$, then $\bar{\vec{c}}_{i+1} \approx \bar{\vec{c}}_{i}$. So although there may be a jump in activation, it is temporary.  The memory cells will decay toward the same steady states as before.
\end{itemize}
Among other benefits, this lets us fit datasets in which (as is common) some pairs of event types do {\em not} influence one another.  \Cref{sec:superposition} explains why all the models in this paper have this ability.

The drift of $\vec{c}(t)$ between events controls how the system's expectations about future events change as more time elapses with no event having yet occured.  \Cref{eqn:c_decay} chooses a moderately flexible parametric form for this drift function (see \cref{sec:future} for some alternatives).  \Cref{eqn:ct_cell_instant} was designed so that $\vec{c}$ in an LSTM could learn to count past events with discrete-time exponential discounting; and \eqref{eqn:c_decay} can be viewed as extending that to continuous-time exponential discounting.

Our memory cell vector $\vec{c}(t)$ is a {\em deterministic} function of the past history $(\mathcal{H}_{i}, t-t_i)$.\footnote{\Cref{sec:bos} explains how our LSTM handles the start and end of the sequence.}
  Thus, the event intensities at any time are also deterministic via \cref{eqn:hawkes_mod}.
The stochastic part of the model is the random choice---based on these intensities---of {\em which} event happens next and {\em when} it happens.  The events are in competition: an event with high intensity is likely to happen sooner than an event with low intensity, and whichever one happens first is fed back into the LSTM.  If no event type has high intensity, it may take a long time for the next event to occur.\looseness=-1

Training the model means learning the LSTM parameters in \cref{eqn:ct_lstm,eqn:ct_cell_rate} along with the other parameters mentioned in this section, namely $s_k \in \Real$ and $\vec{w}_k \in \Real^D$ for $k \in \{1,2,\ldots, K\}$.

\vspace{-4pt}
\section{Algorithms}
\label{sec:algo}

\vspace{-4pt}
For the proposed models,
the log-likelihood \eqref{eqn:loglik-orig} of the parameters turns out to be given by a simple formula---the sum of the log-intensities of the events that happened, at the times they happened, minus an integral of the total intensities over the observation interval $[0,T]$:
\begin{equation}\label{eqn:loglik}
    {\ell} = \sum_{i: t_i \leq T} \log \lambda_{k_i}(t_i) - \underbrace{\int_{t=0}^{T} \lambda(t) dt}_{\text{call this }\Lambda}
\end{equation}
The full derivation is given in \cref{sec:likelihood}. Intuitively, the $-\Lambda$ term (which is $\leq 0$) sums the log-probabilities of infinitely many {\em non}-events.  Why?  The probability that there was {\em not} an event of any type in the infinitesimally wide interval $[t,t+dt)$ is $1-\lambda(t)dt$, whose log is $-\lambda(t) dt$.

We can locally maximize ${\ell}$ using any stochastic gradient method. A detailed recipe is given in~\cref{sec:monte-carlo-gradient}, including the Monte Carlo trick we use to handle the integral in \cref{eqn:loglik}.

If we wish to draw random sequences from the model, we can adopt the thinning algorithm~\citep{lewis-79-sim,liniger-09-hawkes} that is commonly used for the Hawkes process.  See \cref{sec:thinning}.

Given an event stream prefix $(k_1,t_1)$, $(k_2,t_2)$, \ldots, $(k_{i-1},t_{i-1})$, we may wish
to predict the {\em time} and {\em type} of the single next event.
The next event's time $t_i$ has density
$p_i(t) = P(t_i=t \mid \mathcal{H}_i) = \lambda(t) \exp \left( -\int_{t_{i-1}}^{t} \lambda(s) ds \right)$.
To predict a single time whose expected L$_2$ loss is as low as possible,
we should choose
$    \hat{t}_{i} = \E{t_i \mid \mathcal{H}_i} = \int_{t_{i-1}}^{\infty} t p_i(t) dt$.
Given the next event time $t_i$, the most likely type would be $\argmax_k \lambda_k(t_i)/\lambda(t_i)$, but the
most likely next event type {\em without} knowledge of $t_i$ is
$    \hat{k}_{i} = \argmax_{k} \int_{t_{i-1}}^{\infty} \frac{\lambda_k(t)}{\lambda(t)} p_i(t) dt$.
The integrals in
the preceding equations can be estimated by Monte Carlo sampling much as before (\cref{sec:monte-carlo-gradient}).
For event type prediction,
we recommend a paired comparison that uses the same $t$ values for each $k$ in the $\argmax$; this also lets us share the $\lambda(t)$ and $p_i(t)$ computations  across all $k$.

\vspace{-4pt}
\section{Related Work}
\label{sec:related}

\vspace{-4pt}
The Hawkes process has been widely used to model event streams,
including for topic modeling and clustering of text document streams \citep{he-15-topic,du-15-dirichlet},
constructing and inferring network structure \citep{yang-13-mixture,choi-15-constructing,etesami-16-network},
personalized recommendations based on users' temporal behavior
\citep{du-15-time},
discovering patterns in social interaction \citep{guo-15-bayesian,lukasik-16-hawkes},
learning causality \citep{xu-16-causality},
and so on.

Recent interest has focused on expanding the expressivity of Hawkes processes.
\citet{zhou-13-kernels} describe a self-exciting process that removes the assumption of exponentially decaying influence (as we do).  They replace the scaled-exponential summands in \cref{eqn:hawkes} with learned positive functions of time (the choice of function again depends on $k_i,k$).
\citet{lee-16-hawkes} generalize the constant excitation parameters $\alpha_{j,k}$ to be stochastic, which increases expressivity.  Our model also allows non-constant interactions between event types, but arranges these via deterministic, instead of stochastic, functions of continuous-time LSTM hidden states.
\citet{wang-16-isotonic} consider non-linear effects of past history on the future, by passing the intensity functions of the Hawkes process through a non-parametric isotonic link function $g$, which is in the same place as our non-linear function $f_k$. In contrast, our $f_k$ has a fixed parametric form (learning only the scale parameter), and is approximately linear when $x$ is large. This is because we model non-linearity (and other complications) with a continuous-time LSTM, and use $f_k$ only to ensure positivity of the intensity functions.\looseness=-1

\citet{du-16-recurrent} independently combined Hawkes processes with recurrent neural networks
(and \cite{xiao-17-wgan} propose an advanced way of estimating the parameters of that model).
 However, \citeauthor{du-16-recurrent}'s architecture is different in several respects.  They use standard discrete-time LSTMs without our decay innovation, so they must encode the intervals between past events as explicit numerical inputs to the LSTM.  They have only a single intensity function $\lambda(t)$, and it simply decays exponentially toward 0 between events, whereas our more modular model creates separate (potentially transferrable) functions $\lambda_k(t)$, each of which allows complex and non-monotonic dynamics en route to a non-zero steady state intensity.  Some structural limitations of their design are that $t_i$ and $k_i$ are conditionally independent given $\vec{h}$ (they are determined by separate distributions), and that their model cannot avoid a positive probability of extinction at all times.  Finally, since they take $f=\exp$, the effect of their hidden units on intensity is effectively multiplicative, whereas we take $f=\text{softplus}$ to get an approximately additive effect inspired by the classical Hawkes process. Our rationale is that additivity is useful to capture independent (disjunctive) causes; at the same time, the hidden units that our model adds up can each capture a complex joint (conjunctive) cause.

\section[Experiments]{Experiments\footnote{Our code and data are available at \url{https://github.com/HMEIatJHU/neurawkes}.}}

We fit our various models on several simulated and real-world datasets, and evaluated them in each case by the {\em log-probability} that they assigned to held-out data.  We also compared our approach with that of \citet{du-16-recurrent} on their {\em prediction} task.
The datasets that we use in this paper range from one extreme with only $K=2$ event types but mean sequence length $> 2000$, to the other extreme with $K=5000$ event types but mean sequence length 3.
Dataset details can be found in~\cref{tab:stats_dataset} in~\cref{sec:data_stats}.
Training details (e.g., hyperparameter selection) can be found in~\cref{sec:training_details}.

\vspace{-3pt}
\subsection{Synthetic Datasets}\label{sec:simulated}
\vspace{-3pt}

In a pilot experiment with synthetic data (\cref{sec:simulated_details}), we confirmed that the neural Hawkes process generates data that is not well modeled by training an ordinary Hawkes process, but that ordinary Hawkes data can be successfully modeled by training an neural Hawkes process.

In this experiment, we were not limited to measuring the likelihood of the models on the stochastic event sequences.  We also knew the true latent intensities of the generating process, so we were able to directly measure whether the trained models predicted these intensities accurately.  The pattern of results was similar.

\vspace{-3pt}
\subsection{Real-World Media Datasets}\label{sec:num-params}\label{sec:retweet}\label{sec:memetrack}
\vspace{-3pt}

\paragraph{Retweets Dataset \textnormal{\citep{zhao-15-seismic}}.}
On Twitter, {\em novel} tweets are generated from some distribution, which we do not model here.  Each novel tweet serves as the beginning-of-stream event (see \cref{sec:bos}) for a subsequent stream of {\em retweet} events.  We model the dynamics of these streams: how retweets by various types of users ($K=3$) predict later retweets by various types of users.

Details of the dataset and its preparation are given in~\cref{sec:retweet_details}.
The dataset is interesting for its temporal pattern.  People like to retweet an interesting post soon after it is created and retweeted by others, but may gradually lose interest, so the intervals between retweets become longer over time.  In other words, the stream begins in a {\em self-exciting state}, in which previous retweets increase the intensities of future retweets, but eventually interest dies down and events are less able to excite one another.
The decomposable models are essentially incapable of modeling such a phase transition, but our neural model should have the capacity to do so.

\begin{figure}
\includegraphics[width=0.24\textwidth]{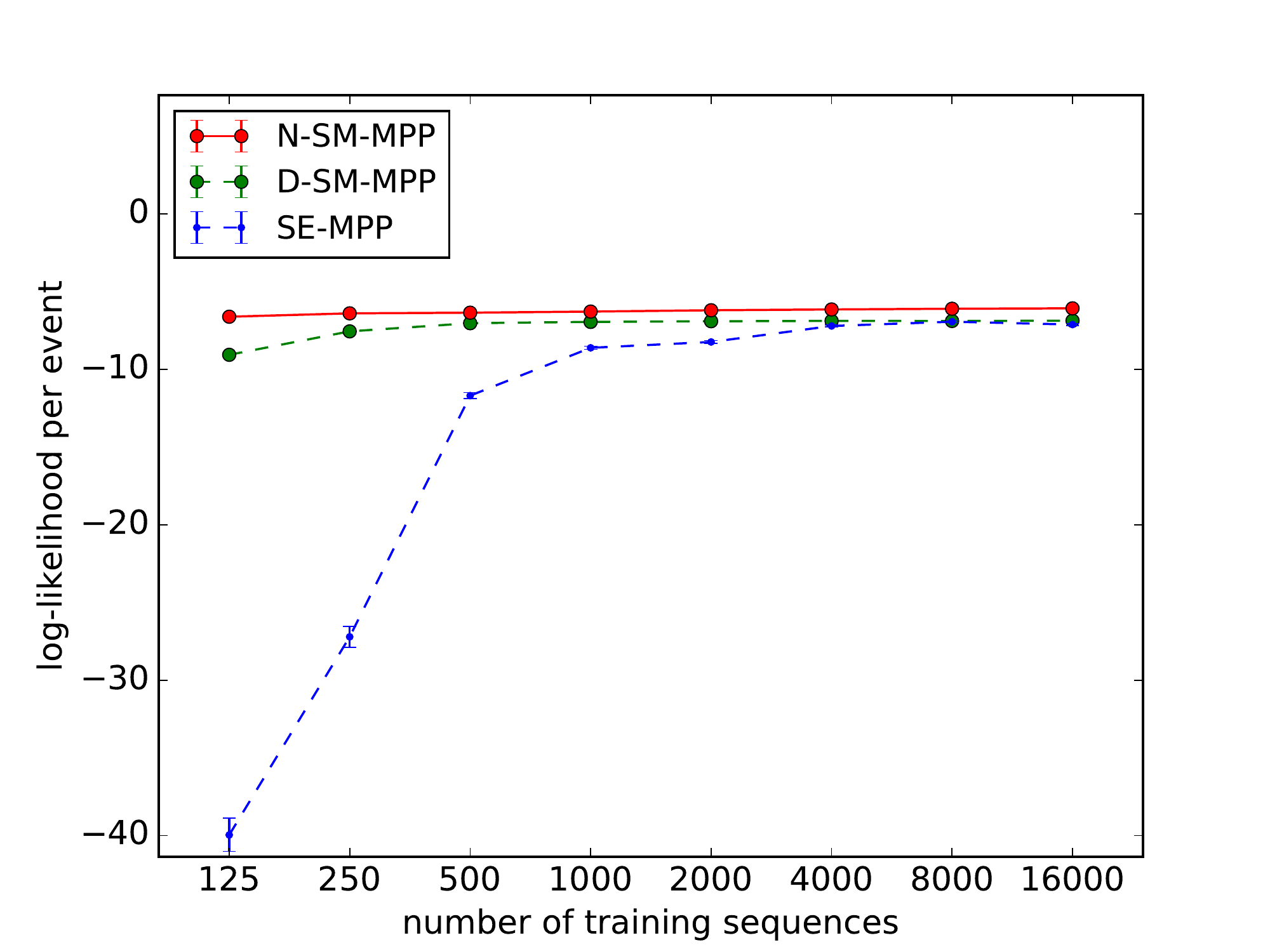}
\includegraphics[width=0.24\textwidth]{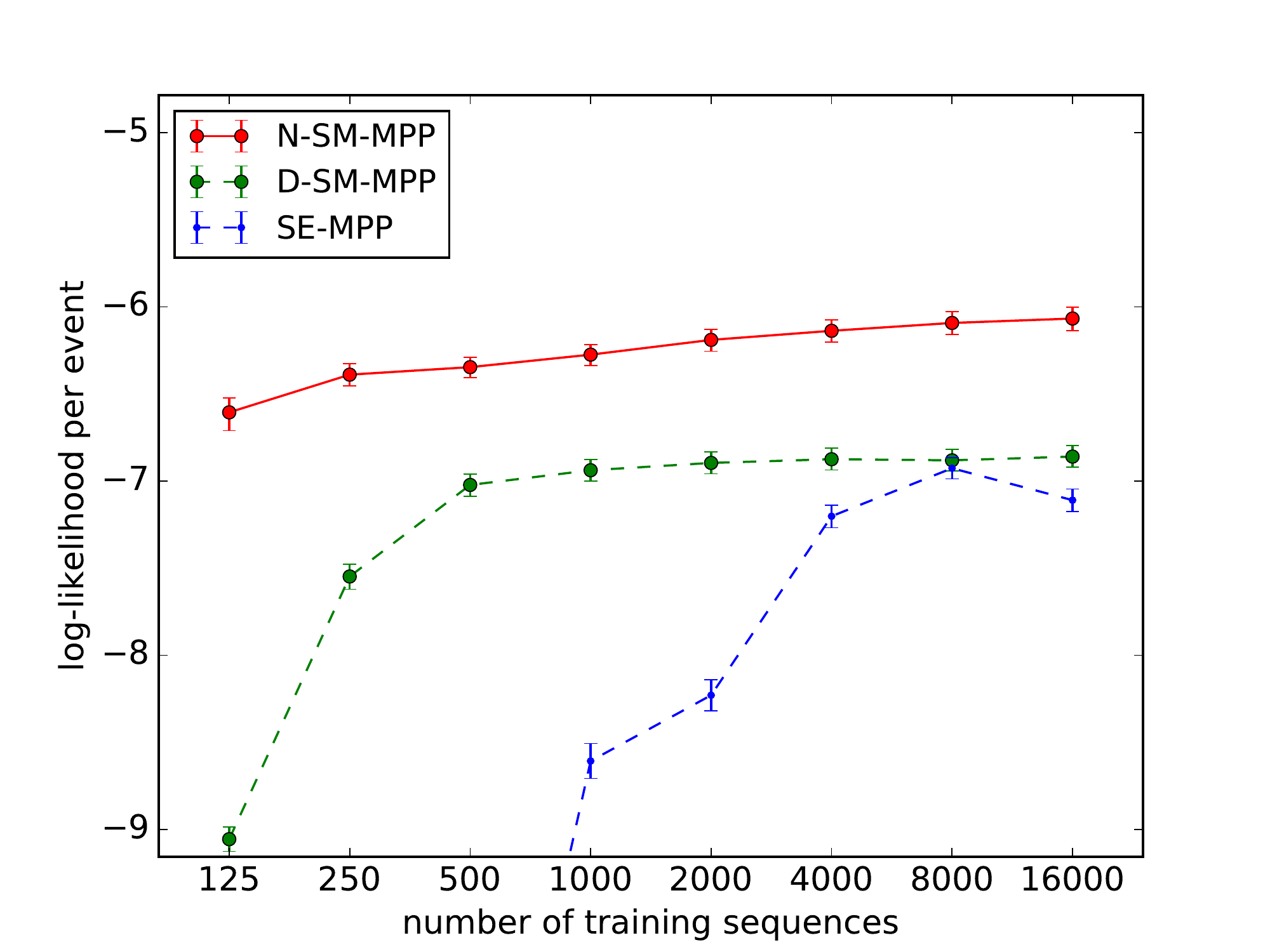}
\hfill
\includegraphics[width=0.24\textwidth]{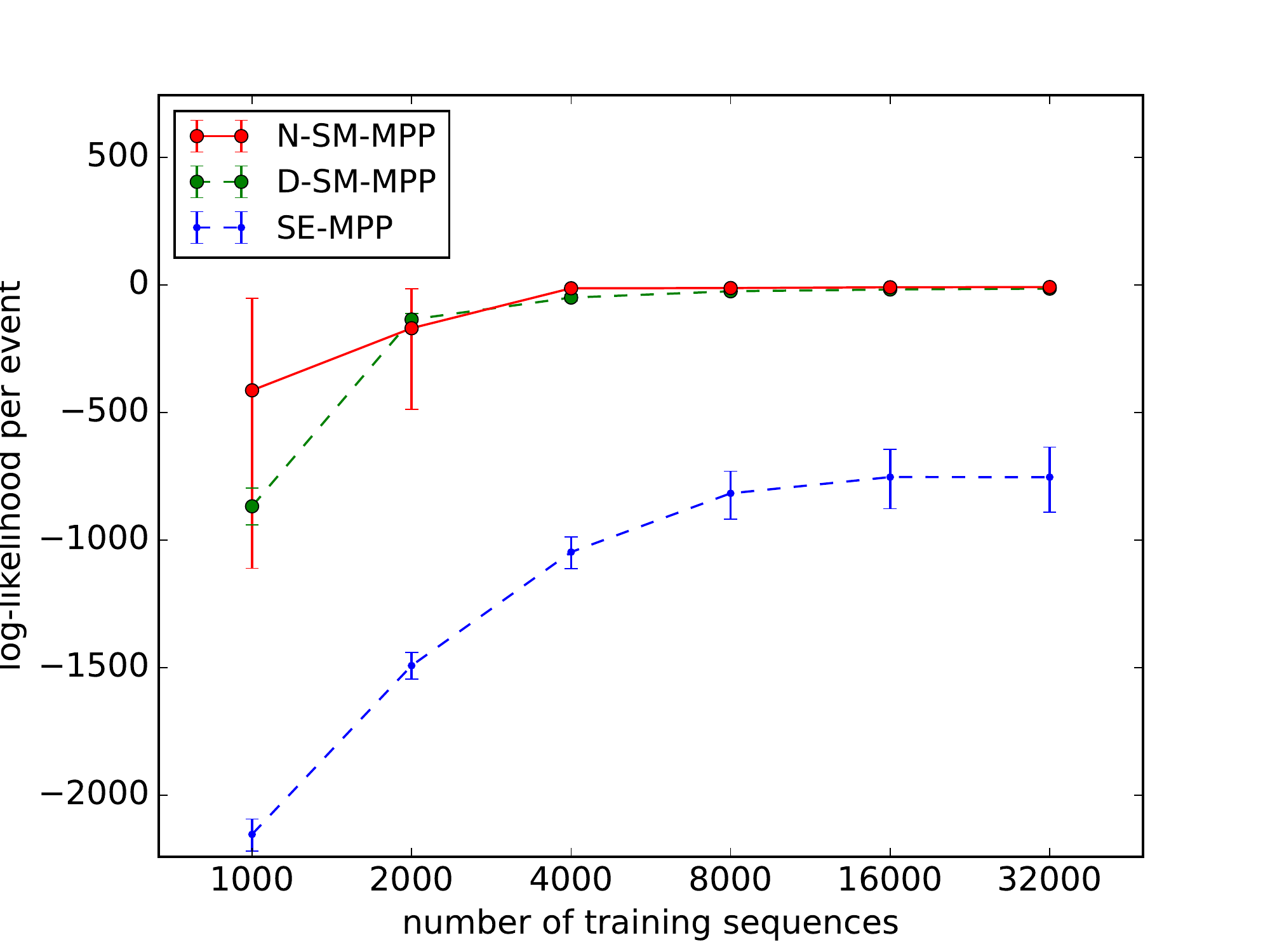}
\includegraphics[width=0.24\textwidth]{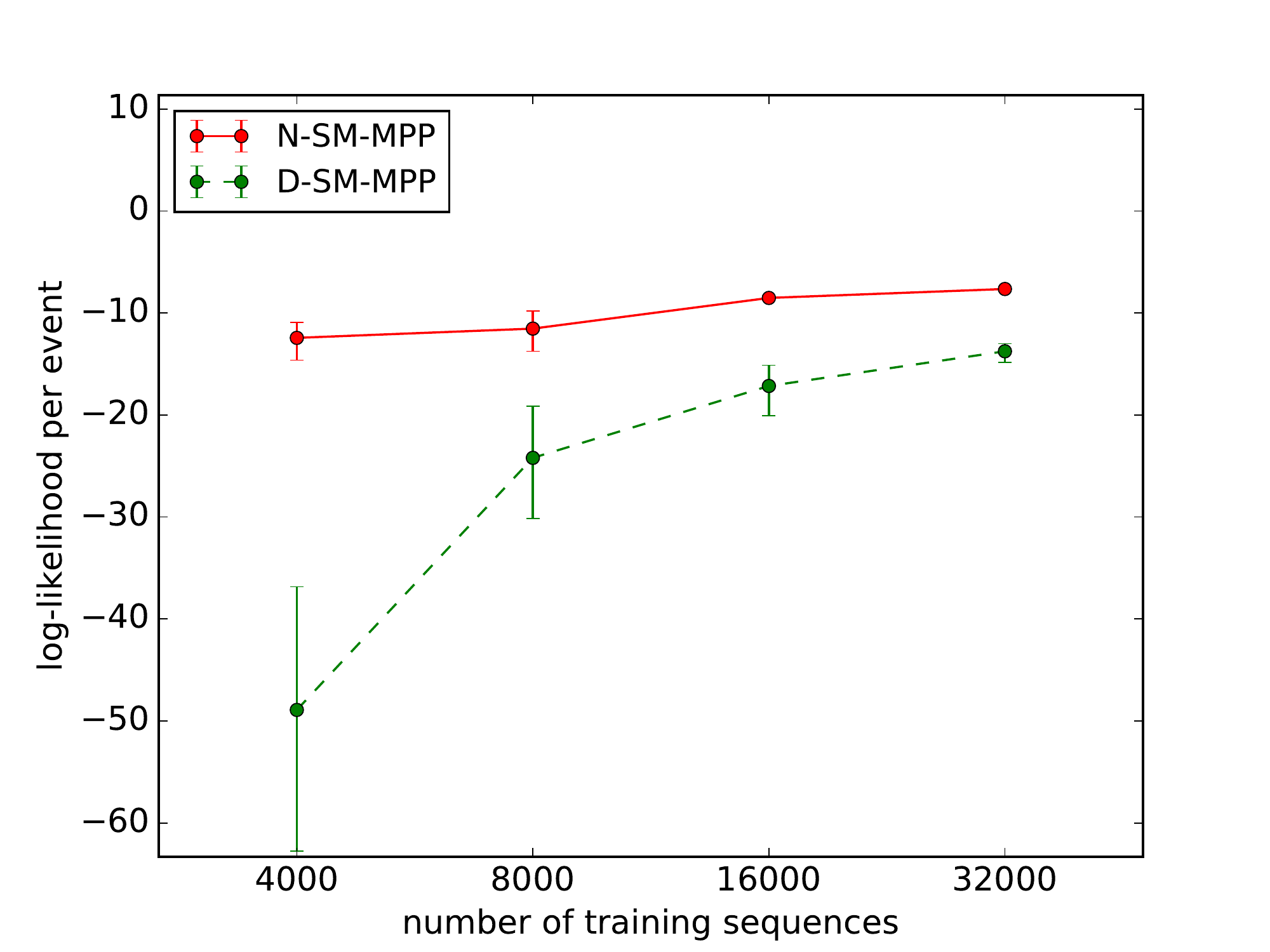}
\caption{
	Learning curve (with $95\%$ error bars) of all three models on the Retweets (left two) and MemeTrack (right two) datasets.
	Our neural model significantly outperforms our decomposable
        model (right graph of each pair), and both significantly outperform the Hawkes
        process (left of each pair---same graph zoomed out).
}
\label{fig:curve_retweet}
\label{fig:curve_meme}
\end{figure}

\begin{figure}
\minipage[t]{0.65\textwidth}
	\includegraphics[width=0.49\textwidth]{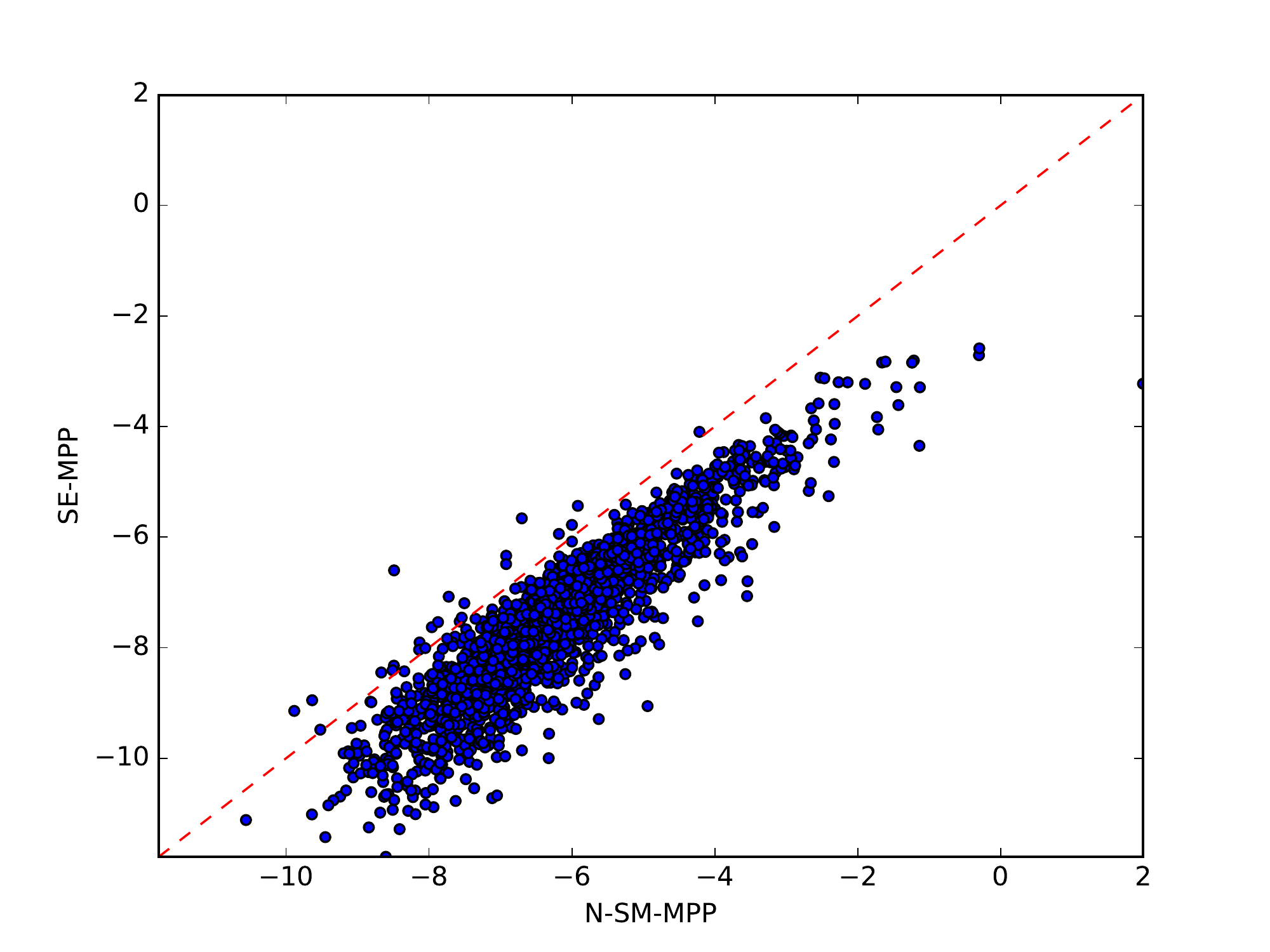}
	\includegraphics[width=0.49\textwidth]{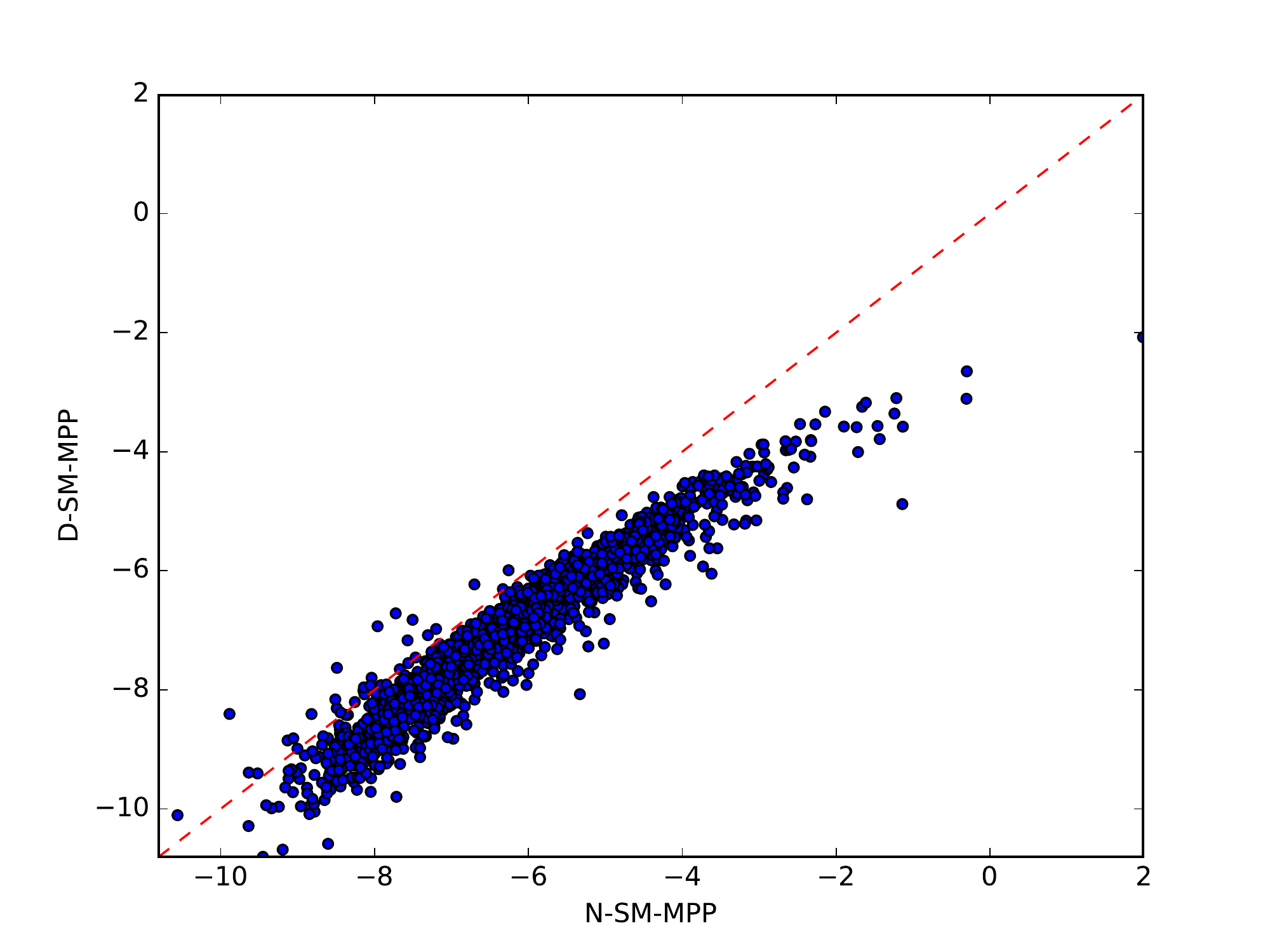}
\caption{
	Scatterplots of N-SM-MPP vs.\@ SE-MPP (left) and N-SM-MPP vs.\@ D-SM-MPP (right), comparing the held-out log-likelihood of the two models (when trained on our full Retweets training set) with respect to {\em each} of the 2000 test sequences.
	Nearly all points fall to the right of $y=x$, since N-SM-MPP (the neural Hawkes process) is consistently more predictive than our non-neural model and the Hawkes process.\looseness=-1
}
\label{fig:scatter_retweet}
\endminipage
\hfill
\minipage[t]{0.30\textwidth}
\centering
\includegraphics[width=\textwidth]{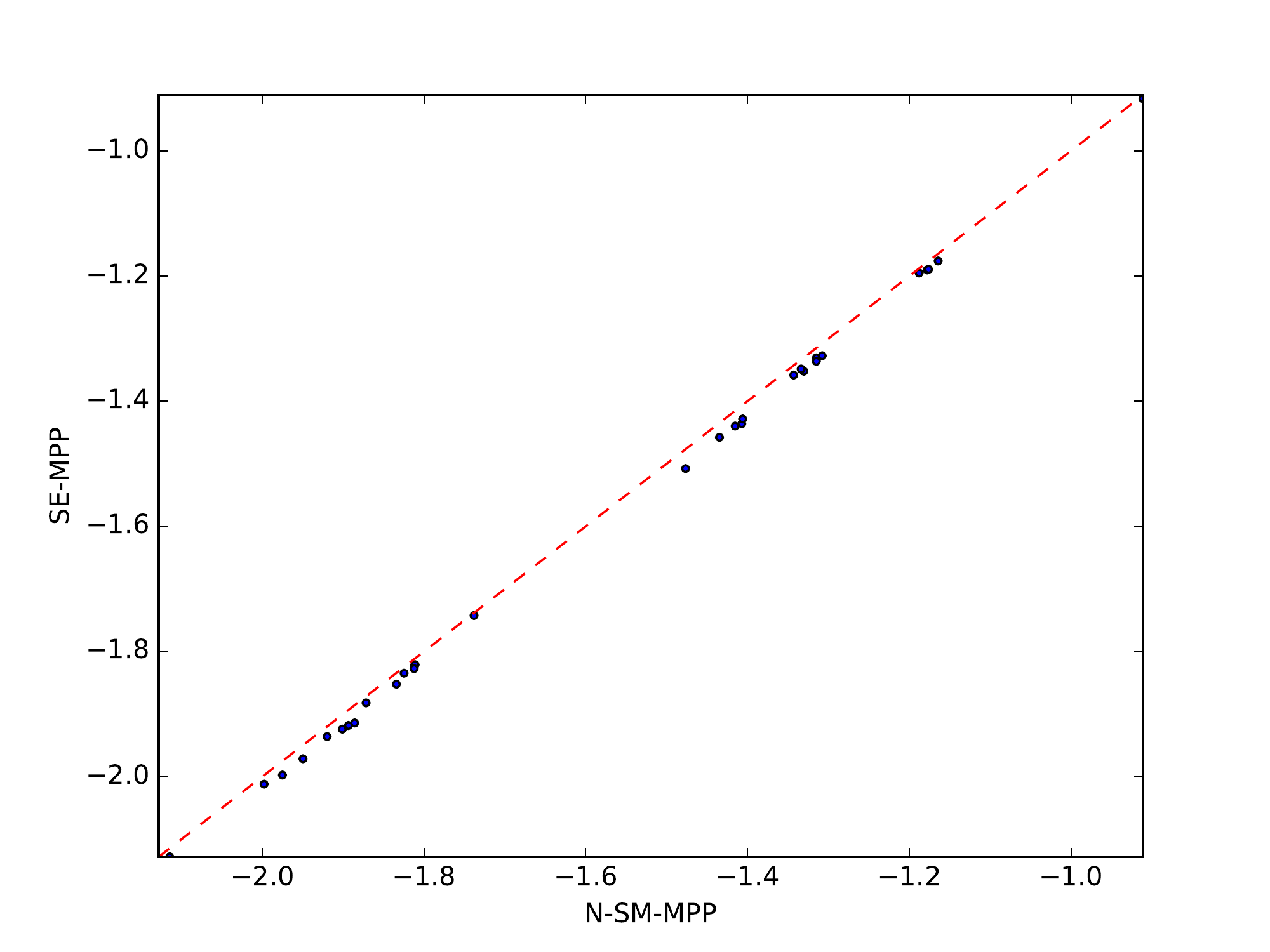}
\caption{
	Scatterplot of N-SM-MPP vs.\@ SE-MPP, comparing their log-likelihoods with respect to {\em each} of the 31 incomplete sequences' test sets.
	All 31 points fall to the right of $y=x$.
}
\label{fig:scatter_missing}
\endminipage
\end{figure}

We generated learning curves (\cref{fig:curve_retweet}) by training our models on increasingly long prefixes of the training set.
As we can see, our self-modulating processes {\em significantly} outperform the Hawkes process at {\em all} training sizes.
There is no obvious {\em a priori} reason to expect inhibition or even inertia in this application domain, which explains why the D-SM-MPP makes only a small improvement over the Hawkes process when the latter is well-trained.  But D-SM-MPP requires much less data, and also has more stable behavior (smaller error bars) on small datasets.
Our neural model is even better.  Not only does it do better on the average stream, but its {\em consistent} superiority over the other two models is shown by the per-stream scatterplots in~\cref{fig:scatter_retweet}, demonstrating the importance of our model's neural component even with large datasets.

\paragraph{MemeTrack Dataset \textnormal{\citep{snapnets}}.}
This dataset is similar in conception to Retweets, but with many more event types ($K=5000$).  It considers the reuse of fixed phrases, or ``memes,'' in online media.  It contains time-stamped instances of meme use
  in articles and posts from 1.5 million different blogs and news sites.
We model how the future occurrence of a meme is affected by its past
trajectory across different websites---that is, given one meme's past
trajectory across websites, when and where it will be mentioned again.

On this dataset,\footnote{Data preparation details are given in~\cref{sec:meme_details}.}
 the advantage of our full neural models was dramatic, yielding
 cross-entropy per event of around $-8$ relative to the $-15$ of
 D-SM-MPP---which in turn is {\em far} above the $-800$ of the Hawkes process.
\cref{fig:curve_meme}
illustrates the persistent gaps among the models.
A scatterplot similar to~\cref{fig:scatter_retweet} is given in~\cref{fig:scatter_meme} of~\cref{sec:meme_details}.
We attribute the poor performance of the Hawkes process to its failure to capture the latent properties of memes, such as their topic, political stance, or interestingness.  This is a form of missing data (\cref{sec:missing}), as we now discuss.

As the table in \cref{sec:data_stats} indicates,
most memes in MemeTrack are uninteresting and give rise to only a short sequence of mentions.  Thus the base mention probability is low.  An ideal analysis would recognize that if a specific meme has been mentioned several times already, it is {\em a posteriori} interesting and will probably be mentioned in future as well.  The Hawkes process cannot distinguish the interesting memes from the others, except insofar as they appear on more influential websites.  By contrast, our D-SM-MPP can partly capture this inferential pattern by using {\em negative} base rates $\vec{\mu}$ to create ``inertia'' (\cref{sec:D-SM-MPP}).  Indeed, all 5000 of its learned $\mu_k$ parameters were negative, with values ranging from $-10$ to $-30$, which numerically yields $0$ intensity and is hard to excite.

An ideal analysis would also recognize that if a specific meme has appeared mainly on conservative websites, it is {\em a posteriori} conservative and unlikely to appear on liberal websites in the future. The D-SM-MPP, unlike the Hawkes process, can again partly capture this, by having conservative websites {\em inhibit} liberal ones.  Indeed, 24\% of its learned $\alpha$ parameters were negative. (We re-emphasize that this inhibition is merely a predictive effect---probably not a direct causal mechanism.)

And our N-SM-MPP process is even more powerful.  The LSTM state aims to learn sufficient statistics for predicting the future, so it can learn hidden dimensions (which fall in $(-1,1)$) that encode useful posterior beliefs in boolean properties of the meme such as interestingness, conservativeness, timeliness, etc.  The LSTM's ``long short-term memory'' architecture explicitly allows these beliefs to persist indefinitely through time in the absence of new evidence, without having to be refreshed by redundant new events as in the decomposable models.  Also, the LSTM's hidden dimensions are computed by sigmoidal activation rather than softplus activation, and so can be used implicitly to perform logistic regression.  The flat left side of the sigmoid resembles softplus and can model {\em inertia} as we saw above: it takes several mentions to establish interestingness.  Symmetrically, the flat right side can model {\em saturation}: once the posterior probability of interestingness is at 80\%, it cannot climb much farther no matter how many more mentions are observed.

A final potential advantage of the LSTM is that in this large-$K$ setting, it has fewer parameters than the other models (\cref{sec:model_size}), sharing statistical strength across event types (websites) to generalize better.
The learning curves in~\cref{fig:curve_meme} suggest that on small data, the decomposable (non-neural) models
may overfit their $O(K^2)$ interaction parameters $\alpha_{j,k}$.
Our neural model only has to learn $O(D^2)$ pairwise interactions among its $D$ hidden nodes (where $D \ll K$), as well as $O(KD)$ interactions between the hidden nodes and the $K$ event types.  In this case, $K=5000$ but $D=64$.  This reduction by using latent hidden nodes is analogous to nonlinear latent factor analysis.

\vspace{-3pt}
\subsection{Modeling Streams With Missing Data}
\label{sec:missing_data}

\vspace{-3pt}

We set up an artificial experiment to more directly investigate the missing-data setting of \cref{sec:missing}, where we do not observe {\em all} events during $[0,T]$, but train and test our model just as if we had.

We sampled synthetic event sequences from a standard Hawkes process (just as in our pilot experiment from \ref{sec:simulated}),
removed all the events of selected types, and then compared the neural Hawkes process (N-SM-MPP) with the Hawkes process (SE-MPP) as models of these censored sequences. Since we took $K=5$, there were $2^5-1=31$ ways to construct a dataset of censored sequences. As shown in~\cref{fig:scatter_missing}, for {\em each} of the 31 resulting datasets, training a neural Hawkes model achieves better generalization.  \Cref{sec:missing_data_details} discusses why this kind of behavior is to be expected.

\vspace{-3pt}
\subsection{Prediction Tasks---Medical, Social and Financial}\label{sec:prediction}
\vspace{-3pt}

To compare with \citet{du-16-recurrent}, we evaluate our model on the {\em prediction} tasks and datasets that they proposed.
The Financial Transaction dataset contains long streams of high frequency stock transactions for a single stock, with the two event types ``buy'' and ``sell.'' The electrical medical records (MIMIC-II) dataset is a collection of de-identified clinical visit records of Intensive Care Unit patients for 7 years. Each patient has a sequence of hospital visit events, and each event records its time stamp and disease diagnosis.
The Stack Overflow dataset represents two years of user awards on a question-answering website: each user received a sequence of badges (of 22 different types).

We follow \citet{du-16-recurrent} and attempt to predict every held-out event $(k_i,t_i)$ from its history ${\mathcal H}_i$,
evaluating the prediction $\hat{k}_i$  with 0-1 loss (yielding an error rate, or ER) and evaluating the prediction $\hat{t}_i$ with L2 loss (yielding a root-mean-squared error, or RMSE).  We make minimum Bayes risk predictions as explained in \cref{sec:algo}.
\cref{fig:prediction} in~\cref{sec:prediction_details} shows that
our model consistently outperforms that of \citet{du-16-recurrent} on event type prediction on all the datasets, although for time prediction neither model is consistently better.

\vspace{-3pt}
\subsection{Sensitivity to Number of Parameters}
\vspace{-3pt}

Does our method do well because of its flexible nonlinearities or just because it has more parameters? The answer is both.  We experimented on the Retweets data with reducing the number of hidden units $D$.  Our N-SM-MPP substantially outperformed SE-MPP (the Hawkes process) on held-out data even with very few parameters, although more parameters does even better:

\begin{tabular}{l|rrrrrrrr} \hline
number of hidden units & Hawkes & 1 & 2 & 4 & 8 & 16 & 32 & 256 \\
number of parameters & 21 & 31 & 87 & 283 & 1011 & 3811 & 14787 & 921091  \\
log-likelihood & -7.19 & -6.51 & -6.41 & -6.36 & -6.24 & -6.18 & -6.16 & -6.10  \\ \hline
\end{tabular}

We also tried halving $D$ across several datasets, which had negligible effect, always decreasing held-out log-likelihood by $< 0.2$\% relative.

More information about model sizes is given in \cref{sec:model_size}.  Note that the neural Hawkes process does not {\em always} have more parameters.  When $K$ is large, we can greatly reduce the number of params below that of a Hawkes process, by choosing $D \ll K$, as for MemeTrack in \cref{sec:num-params}.

\vspace{-4pt}
\section{Conclusion}
\label{sec:conclusion}

\vspace{-4pt}
We presented two extensions to the multivariate Hawkes process, a popular generative model of streams of typed, timestamped events.
Past events may now either excite {\em or} inhibit future events.
They do so by {\em sequentially} updating the state of a novel {\em continuous-time} recurrent neural network (LSTM).
Whereas Hawkes sums the time-decaying influences of past events, we instead sum the time-decaying influences of the LSTM nodes.  Our extensions to Hawkes aim to address real-world phenomena,
missing data, and causal modeling.  Empirically, we have shown that
both extensions yield a significantly improved ability to predict the
course of future events.
There are several exciting avenues for further improvements (discussed
in~\cref{sec:future}), including embedding our model within a
reinforcement learner to discover causal structure and learn an
intervention policy.

\section*{Acknowledgments}

We are grateful to Facebook for enabling this work through a gift to the second author.  Nan Du kindly helped us by making his code public and answering questions, and the NVIDIA Corporation kindly donated two Titan X Pascal GPUs.  We also thank our lab group at Johns Hopkins University's Center for Language and Speech Processing for helpful comments.  The first version of this work appeared on arXiv in December 2016.

\bibliographystyle{plainnat}

\clearpage
\appendix
\appendixpage
\phantom{\begin{minipage}{\textwidth}
\textit{[Supplementary material for Hongyuan Mei \& Jason Eisner, ``The Neural Hawkes Process: A Neurally Self-Modulating Multivariate Point Process,'' NIPS 2017.]}
\vspace{12pt}
\end{minipage}}

\section{Model Details}
\label{sec:model_details}

In this appendix, we discuss some qualitative properties of our models and give details about how we handle boundary conditions.

\subsection{Discussion of the Transfer Function}
\label{sec:transfer}

As explained in~\cref{sec:self_modulate}, when we allow {\em inhibition} and {\em inertia}, we need to pass the total activation through a non-linear \defn{transfer function} $f: \Real \rightarrow \Real_+$ to obtain a positive intensity function.
This was our \cref{eqn:hawkes_inhib_a}, namely
$\lambda_k(t) = f(\tilde{\lambda}_k(t))$.

What non-linear function $f$ should we use? The ReLU function $f(x) = \max(x,0)$ seems at first a natural choice.  However, it returns 0 for negative $x$; we need to keep our intensities strictly positive at all times when an event could possibly occur, to avoid infinitely bad log-likelihood at training time or infinite log-loss at test time.

A better choice would be the ``softplus'' function  $f(x) = \log(1+\exp(x))$, which is strictly positive and approaches ReLU when $x$ is far from $0$. Unfortunately, ``far from $0$'' is defined in units of $x$, so this choice would make our model sensitive to the units used to measure time.  For example, if we switch the units of $t$ from seconds to milliseconds, then the base intensity $f(\mu_k)$ must become 1000 times lower, forcing $\mu_k$ to be very negative and thus creating a much stronger inertial effect.

\begin{figure}[h]
\begin{center}
\centerline{\includegraphics[width=0.6\textwidth]{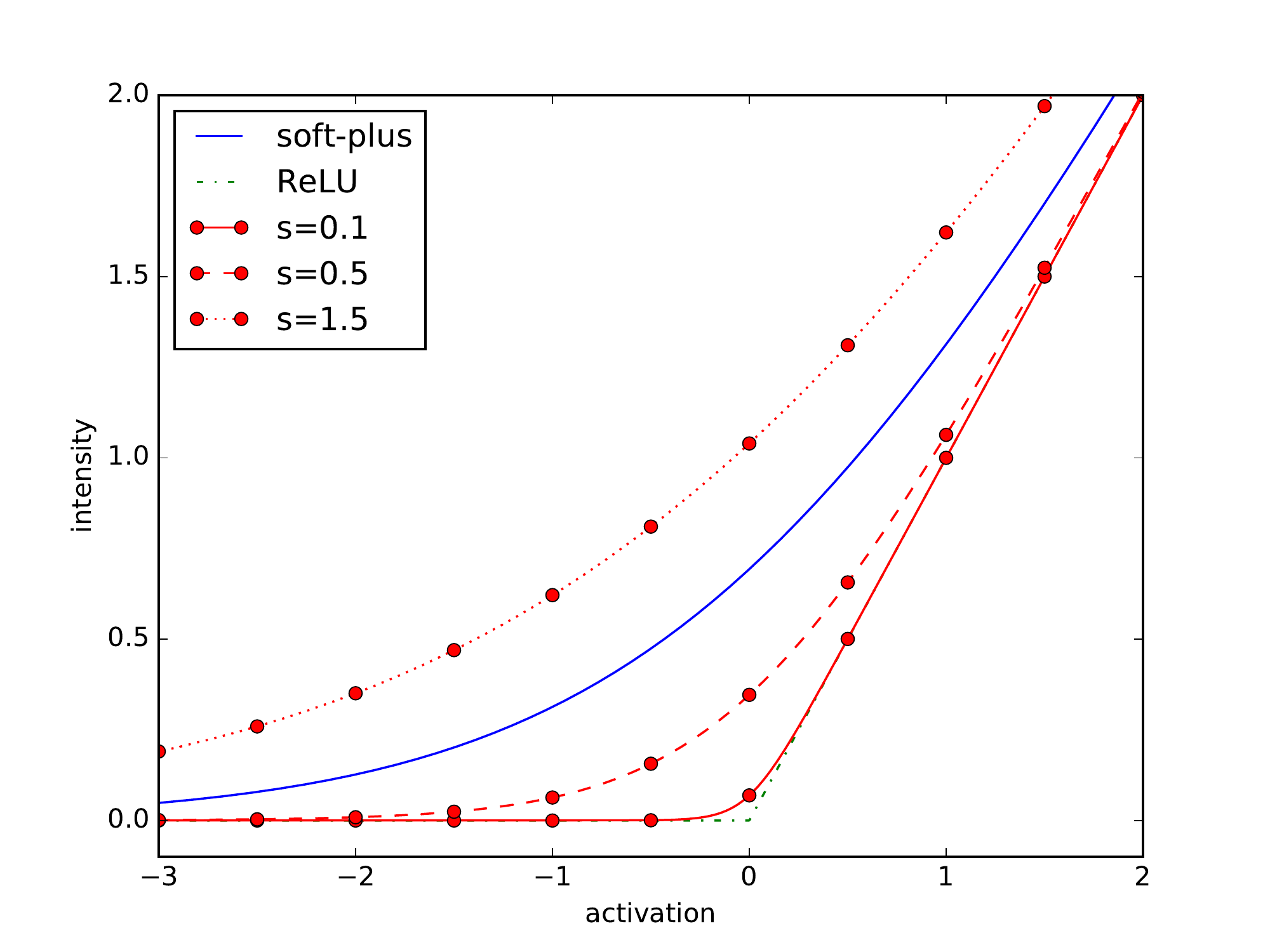}}
\caption{
	The softplus function is a soft approximation to a rectified linear unit (ReLU), approaching it as $x$ moves away from $0$.  We use it to ensure a strictly positive intensity function.   We incorporate a scale parameter $s$ that controls the curvature.
}
\label{fig:softplus}
\end{center}
\end{figure}

To avoid this problem, we introduce a scale parameter $s > 0$ and define $f(x)=s \log(1+\exp(x/s))$.
  The scale parameter $s$ controls the curvature of $f(x)$, which approaches ReLU as $s \rightarrow 0$, as shown in~\cref{fig:softplus}.  We can regard $f(x)$, $x$, and $s$ as rates, with units of inverse time, so that $f(x)/s$ and $x/s$ are unitless quantities related by softplus.  We actually learn a separate scale parameter $s_k$ for each event type $k$, which will adapt to the rate of events of that type.

\subsection{Boundary Conditions for the LSTM}
\label{sec:bos}

We initialize the continuous-time LSTM's hidden state to $\vec{h}(0) = \vec{0}$, and then have it read a special beginning-of-stream (\bos) event $(k_0, t_0)$, where $k_0$ is a special event type (i.e., expanding the LSTM's input dimensionality by one) and $t_0$ is set to be $0$.
Then \crefrange{eqn:ct_lstm}{eqn:ct_cell} define $\vec{c}_1$ (from $\vec{c}_0 \defeq \vec{0}$), $\bar{\vec{c}}_1$, $\vec{\delta}_1$, and $\vec{o}_1$.  This is the initial configuration of the system as it waits for the first event to happen: this initial configuration determines the hidden state $\vec{h}(t)$ and the intensity functions $\lambda_k(t)$ over $t \in (0,t_1]$

We do not generate the \bos\ event but only condition on it, which is why the log-likelihood formula (\cref{sec:notation}) only sums over $i=1,2,\ldots$.  This design is well-suited to various settings.  In some settings, time $0$ is special. For example, if we release children into a carnival and observe the stream of their actions there, then \bos is the release event and no other events can possibly precede it.  In other settings, data before time 0 are simply missing, e.g., the observation of a patient starts in midlife; nonetheless, \bos in this case usefully indicates the beginning of the {\em observed} sequence.  In both kinds of settings, the initial configuration just after reading $\bos$
 characterizes the model's belief about the unknown state of the true system just after time 0, as it waits for event 1.  Computing the initial configuration by explicitly transitioning on \bos ensures that the initial hidden state $\vec{h}(0^+) \defeq \lim_{t \rightarrow 0^+} \vec{h}(t)$ falls in the space of hidden states achievable by LSTM transitions.  More important, in future work, we will be able to attach metadata about the sequence as a ``mark'' to the \bos event (see \cref{fn:mark}), and the LSTM can learn how these metadata affect the initial configuration.

To allow finite streams, we could optionally choose to identify one of the observable types in $\{1,2,\ldots,K\}$ as a special end-of-stream (\eos) event after which the stream cannot possibly continue.  If the model generates \eos, all intensities are permanently forced to 0---the LSTM is no longer consulted, so it is not necessary for the model parameters to explain why no further events are observed on the interval $[0,T]$: that is, the second term of \cref{eqn:loglik-orig} can be omitted.  The integral in \cref{eqn:loglik} should therefore be taken from $t=0$ to the time of the $\eos$ event or $T$, whichever is smaller.

\subsection{Closure Under Superposition}
\label{sec:superposition}

Decomposable models have the nice property that they are closed under superposition of event streams.  Let  ${\cal E}$ and ${\cal E}'$ be random event streams, on a common time interval $[0,T]$ but over disjoint sets of event types.  If each stream is distributed according to a Hawkes process, then their superposition---that is, ${\cal E} \cup {\cal E}'$ sorted into temporally increasing order---is also distributed according to a Hawkes process.  It is easy to exhibit parameters for such a process, using a block-diagonal matrix of $\alpha_{j,k}$ so that the two sets of event types do not influence each other.  The closure property also holds for our decomposable self-modulating process, and for the same simple reason.

This is important since in various real settings, some event types tend not to interact.  For example, the activities of two people Jay and Kay rarely influence each other,\footnote{Their surnames might be Box and Cox, after the 19th-century farce about a day worker and a night worker unknowingly renting the same room.  But any pair of strangers would do.}
 although they are simultaneously monitored and thus form a single observed stream of events.  We want our model to handle such situations naturally, rather than insisting that Kay always reacts to what Jay does.

 Thus, as \cref{sec:opengates} noted, we have designed our neurally self-modulating process to preserve this ability to insulate event $k$ from event $j$.  By setting specific elements of $\vec{w}_k$ to 0, one could ensure that the intensity function $\lambda_k(t)$ depends on only a subset $S$ of the LSTM hidden nodes. Then by setting specific LSTM parameters, one would make the nodes in $S$ insensitive to events of type $j$: events of type $j$ should open these nodes' forget gates ($\vec{f} = \vec{1}$) and close their input gates ($\vec{i} = \vec{0})$---as \cref{sec:opengates} suggested---so that their cell memories $\vec{c}(t)$ and hidden states $\vec{h}(t)$ do not change at all but continue decaying toward their previous steady-state values.\footnote{To be precise, we can achieve this arbitrarily closely, but not exactly, because a standard LSTM gate cannot be fully opened or closed.  The openness is traditionally given by a sigmoid function and so falls in $(0,1)$, never achieving 1 or 0 exactly unless we are willing to set parameters to $\pm\infty$.  In practice this should not be an issue because relatively small weights can drive the sigmoid function extremely close to 1 and 0---in fact, $\sigma(37)=1$
   in 64-bit floating-point arithmetic.}  Now events of type $j$ cannot affect the intensity $\lambda_k(t)$.

 For example, the hidden states in $S$ are affected in the same way when the LSTM reads $(k,1), (j,3), (j,8), (k,12)$ as when it reads $(k,1), (k,12)$, even though the intervals $\Delta t$ between successive events are different.  In other words, the architecture ``knows'' that $2+5+4 = 11$.  The simplicity of this solution is a consequence of how our design does not encode the time intervals numerically, but only reacts to these intervals indirectly, through the interaction between the timing of events and the spontaneous decay of the hidden states.  The memory cells of $S$ decay for a total duration of 11 between the two $k$ events, even if that interval has been divided into subintervals $2+5+4$.

 With this method, we can explicitly construct a superposition process with LSTM state space $\Real^{d+d'}$---the cross product of the state spaces $\Real^d$ and $\Real^{d'}$ of the original processes---in which Kay's events are not influenced at all by Jay's.

If we know {\em a priori} that particular event types interact only weakly, we can impose an appropriate prior on the neural Hawkes parameters. And in future work with large $K$, we plan to investigate the use of sparsity-inducing regularizers during parameter estimation, to create an inductive bias toward models that have limited interactions, without specifying which particular interactions are present.

Superposition is a formally natural operation on event streams.  It barely arises for ordinary sequence models, such as language models, since the superposition of two sentences is not well-defined unless all of the words carry distinct real-valued timestamps.  However, there is an analogue from formal language theory.  The ``shuffle'' of two sentences is defined to be the set of {\em possible} interleavings of their words---i.e., the set of superpositions that could result from assigning increasing timestamps to the words of each sentence, without duplicates.  It is a standard exercise to show that regular languages are closed under shuffle.  This is akin to our remark that neural-Hawkes-distributed random variables are closed under superposition, and indeed uses a similar cross-product construction on the finite-state automata.  An important difference is that the shuffle construction does not require disjoint alphabets in the way that ours requires disjoint sets of event types.  This is because finite-state automata allow nondeterministic state transitions and our processes do not.

\subsection{Missing Data Discussion}
\label{sec:missing_data_details}

We discussed the case of missing data in \cref{sec:missing}.  Supppose the true complete-data distribution $p^*$ is itself an unknown neural Hawkes process.  As \cref{sec:missing} pointed out, a sufficient statistic for prediction from the incompletely observed past would be the posterior distribution over the true hidden neural state $\vec{t}$ of the unknown process, which was reached by reading the {\em complete} past.  We would ideally obtain our predictions by correctly modeling the missing observations and integrating over them.  However, inference would be computationally quite expensive even if $p^*$ were known, to say nothing of the case where $p^*$ is unknown and we must integrate over its parameters as well.

We instead train a neural model that attempts to bypass these problems.  The hope is that our model's hidden state, after it reads only the observed {\em incomplete} past, will be nearly as predictive as the posterior distribution above.

We can illustrate the goal with reference to the experiment in \cref{sec:missing_data}.  There, the true complete-data distribution $p^*$ happened to be a classical Hawkes process, but we censored some event types.  We then modeled the observed incomplete sequence as if it were a complete sequence.  In this setting, a Hawkes process will in general be unable to fit the data well, which is why the neural Hawkes process has an advantage in all 31 experiments.

What goes wrong with using the Hawkes model?  Suppose that in the true Hawkes model $p^*$, type 1 is rare but strongly excites type 2 and type 3, which do not excite themselves or each other.  Type 1 events are missing in the observed sequence.

What is the correct predictive distribution in this situation (with knowledge of $p^*$)?  Seeing lots of type 2 events in a row suggests that they were preceded by a (single) missing type 1 event, which predicts a higher intensity for type 3 in future.  The more type 2 events we see, the surer we are that there was a type 1 event, but we doubt that there were multiple type 1 events, so the predicted intensity of type 3 is expected to increase sublinearly as $P(\text{type}=1)$ approaches 1.

As neural networks are universal function approximators , a neural Hawkes model may be able to recognize and fit this sublinear behavior in the incomplete training data.  However, if we fit only a Hawkes model to the incomplete training data, it would have to posit that type 2 excites type 3 directly, so the predicted intensity of type 3 would incorrectly increase linearly with the number of type 2 events.

\newpage
\section{Algorithmic Details}
\label{sec:algo_details}
In this appendix, we elaborate on the details of algorithms.

\subsection{Likelihood Function}
\label{sec:likelihood}

For the proposed models, given complete observations of an event stream over the time interval $[0,T]$, the log-likelihood of the parameters turns out to be given by the simple formula shown in \cref{sec:algo}.  We start by giving the full derivation of that formula, repeated here:
\begin{equation*}
    \ell = \sum_{i: t_i\leq T} \log \lambda_{k_i}(t_i) - \underbrace{\int_{t=0}^{T} \lambda(t) dt}_{\text{call this }\Lambda} \tag{\ref{eqn:loglik}}
\end{equation*}

First, we define $N(t) = |\{h: t_h \leq t\}|$ to be the count of events (of any type) preceding time $t$. So given the past history $\mathcal{H}_i$, the number of events in $(t_{i-1},t]$ is denoted as $\Delta N(t_{i-1},t) \defeq N(t)-N(t_{i-1})$.  Let $T_i > t_{i-1}$ be the random variable of the next event time and let $K_{i+1}$ be the random variable of the next event type.  The cumulative distribution function and probability density function of $T_i$ (conditioned on $\mathcal{H}_i$) are given by:
\begin{subequations} \label{eqn:nonhomo_F}
\begin{align}
	F(t)
	&= P(T_i \leq t) = 1-P(T_i > t) \\
	&= 1- P(\Delta N(t_{i-1},t)=0) \\
	&= 1- \exp \left( -\int_{t_{i-1}}^{t} \lambda(s) ds \right) \\
	&= 1- \exp \left( \Lambda(t_{i-1}) - \Lambda(t) \right) \\
	f(t)
	&= \exp \left( \Lambda(t_{i-1}) - \Lambda(t) \right) \lambda(t)
\end{align}
\end{subequations}
where \mbox{$\Lambda(t)=\int_{0}^{t} \lambda(s)ds$} and \mbox{$\lambda(t)=\sum_{k=1}^K \lambda_k(t)$}.

Moreover, given the past history $\mathcal{H}_i$ and the next event time $t_i$, the distribution of $k_i$ is given by:
\begin{equation}
    P(K_i=k_i \mid t_i) = \frac{\lambda_{k_i} (t_i)}{\lambda(t_i)}
\end{equation}

Therefore, we can derive the likelihood function as follows:
\begin{subequations} \label{eqn:likelihood_nonhomo}
\begin{align}
    \mathcal{L}
    &= \prod_{i: t_i\leq T}{\mathcal{L}_i} = \prod_{t_i\leq T}\{{f(t_i ) P(K_i=k_i \mid t_i)} \} \\
    &= \prod_{i: t_i\leq T} \{ \exp \left( \Lambda(t_{i-1}) - \Lambda(t_i) \right) \lambda_{k_i}(t_i) \}
\end{align}
\end{subequations}
and
\begin{subequations} \label{eqn:log-likelihood_nohomo}
\begin{align}
    \ell
    &\defeq \log \mathcal{L} \\
    &= \sum_{i: t_i\leq T} \log \lambda_{k_i}(t_i) - \sum_{i:t_i\leq T} \left( \Lambda(t_i)-\Lambda(t_{i-1}) \right) \\
    &= \sum_{i: t_i\leq T} \log \lambda_{k_i}(t_i) - \Lambda(T) \\
    &= \sum_{i: t_i\leq T} \log \lambda_{k_i}(t_i) - \int_{t=0}^{T} \lambda(t) dt
\end{align}
\end{subequations}

\subsection{Monte Carlo Gradient and Training Speed}
\label{sec:monte-carlo-gradient}

We can locally maximize the log-likelihood ${\ell}$ from \cref{eqn:loglik} using any stochastic gradient method.  For this, we need to be able to get an unbiased estimate of the gradient $\nabla {\ell}$ with respect to the model parameters.  This is straightforward to obtain by back-propagation.  The trick for handling the integral in \cref{eqn:loglik} is that the single function evaluation $T \lambda(t)$ at a random $t \sim \Uniform(0,T)$ gives an unbiased estimate of the entire integral---that is, its expected value is $\Lambda$.  Its gradient via back-propagation is therefore a unbiased estimate of $\nabla\Lambda$ (since gradient commutes with expectation).  The Monte Carlo algorithm in \cref{alg:sample_integral} averages over several samples to reduce the variance of this noisy estimator.
\begin{algorithm}[t]
	\caption{Integral Estimation (Monte Carlo)}
	\label{alg:sample_integral}
	\begin{algorithmic}
          \State {\bfseries Input:} \parbox[t]{2.5in}{interval $[0,T]$; model parameters and \\
                                                    events $(k_1,t_1),\ldots$ for determining $\lambda_j(t)$}
	\State ${\Lambda}\gets 0$; $\nabla\Lambda\gets\vec{0}$
	\For{$N$ samples} \Comment{e.g., take $N > 0$ proportional to $T$}
	   \State draw $t \sim \Uniform(0, T)$
	   \For{$j\gets 1$ {\bfseries to} $K$}
              \State $\Lambda \pluseq \lambda_j(t)$ \Comment{via current model parameters}
              \State $\nabla\Lambda \pluseq \nabla \lambda_j(t)$ \Comment{via back-propagation}
           \EndFor
	\EndFor
	\State $\Lambda \leftarrow T \Lambda / N$; $\nabla\Lambda \leftarrow T\nabla \Lambda / N$ \Comment{weight the samples}
        \State \textbf{return} $(\Lambda,\nabla\Lambda)$
\end{algorithmic}
\end{algorithm}

Each step of Adam training computes the gradient on a training sequence.  With $P$ params, this takes time $O(IP)$ for Hawkes and $O((I+M)P)$ for neural Hawkes, if $I$ is the number of observed events and $M$ is the number of samples used to estimate the integral. We take $M=O(I)$ in practice (see \cref{sec:training_details}), so we have runtime $O(IP)$ like Hawkes.

Note that our stochastic gradient is unbiased for any $M$; large $M$ merely reduces its variance.  The gradient for the Hawkes process has 0 variance, since it has analytical form and does not require sampling at all.

\subsection{Thinning Algorithm for Sampling Sequences}
\label{sec:thinning}

\begin{figure}
\includegraphics[width=0.32\linewidth]{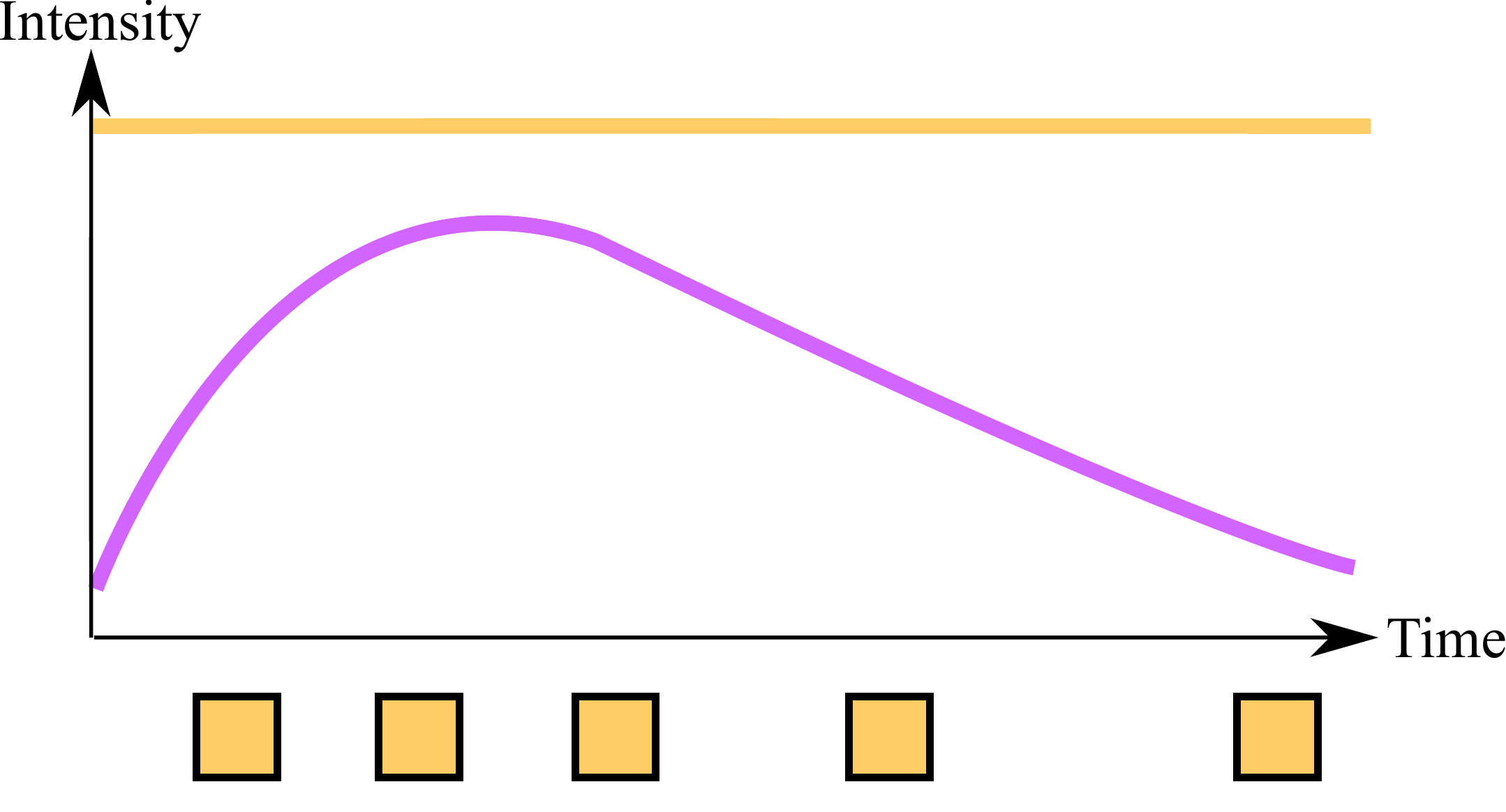}
\includegraphics[width=0.32\linewidth]{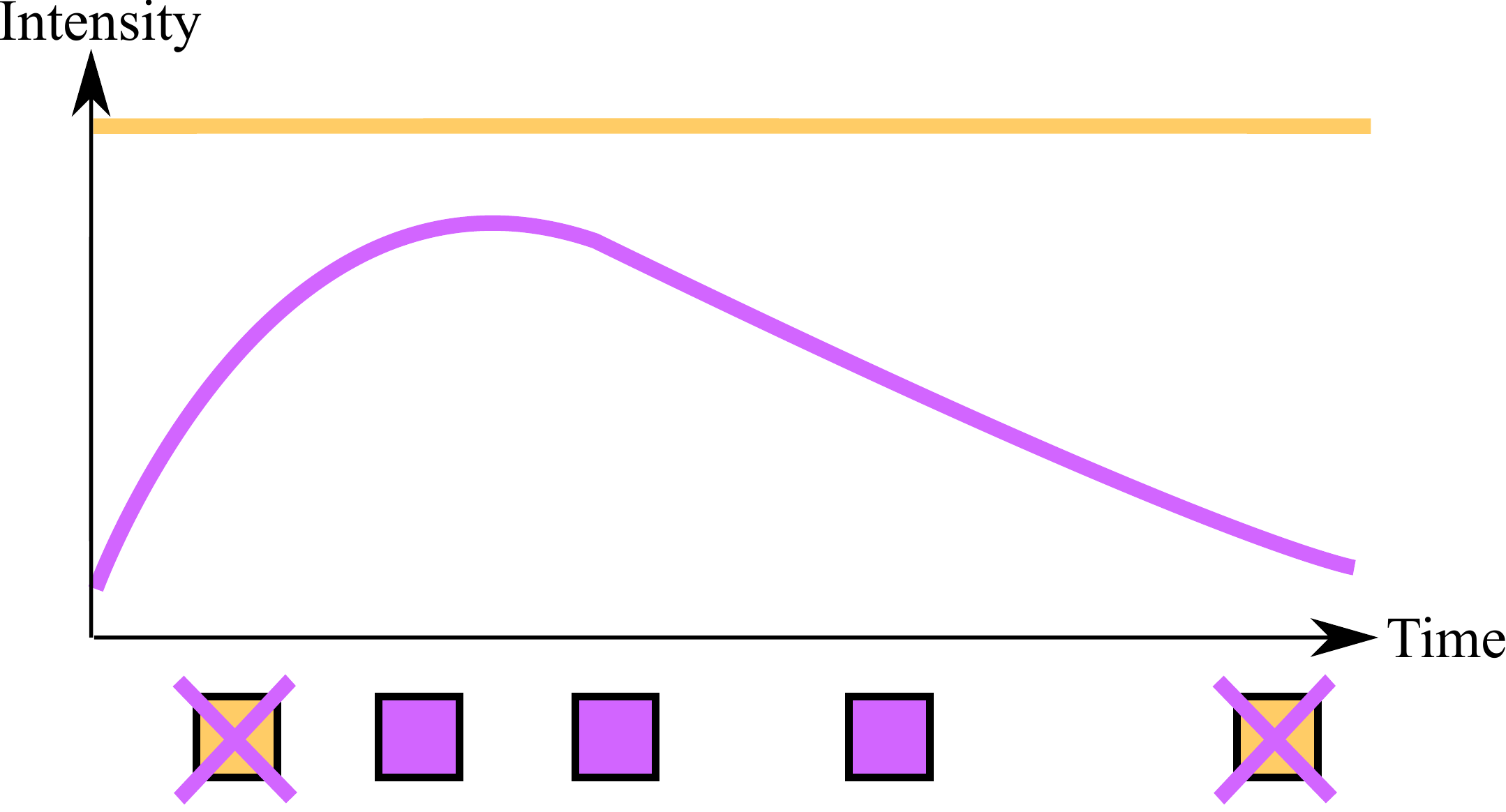}
\includegraphics[width=0.32\linewidth]{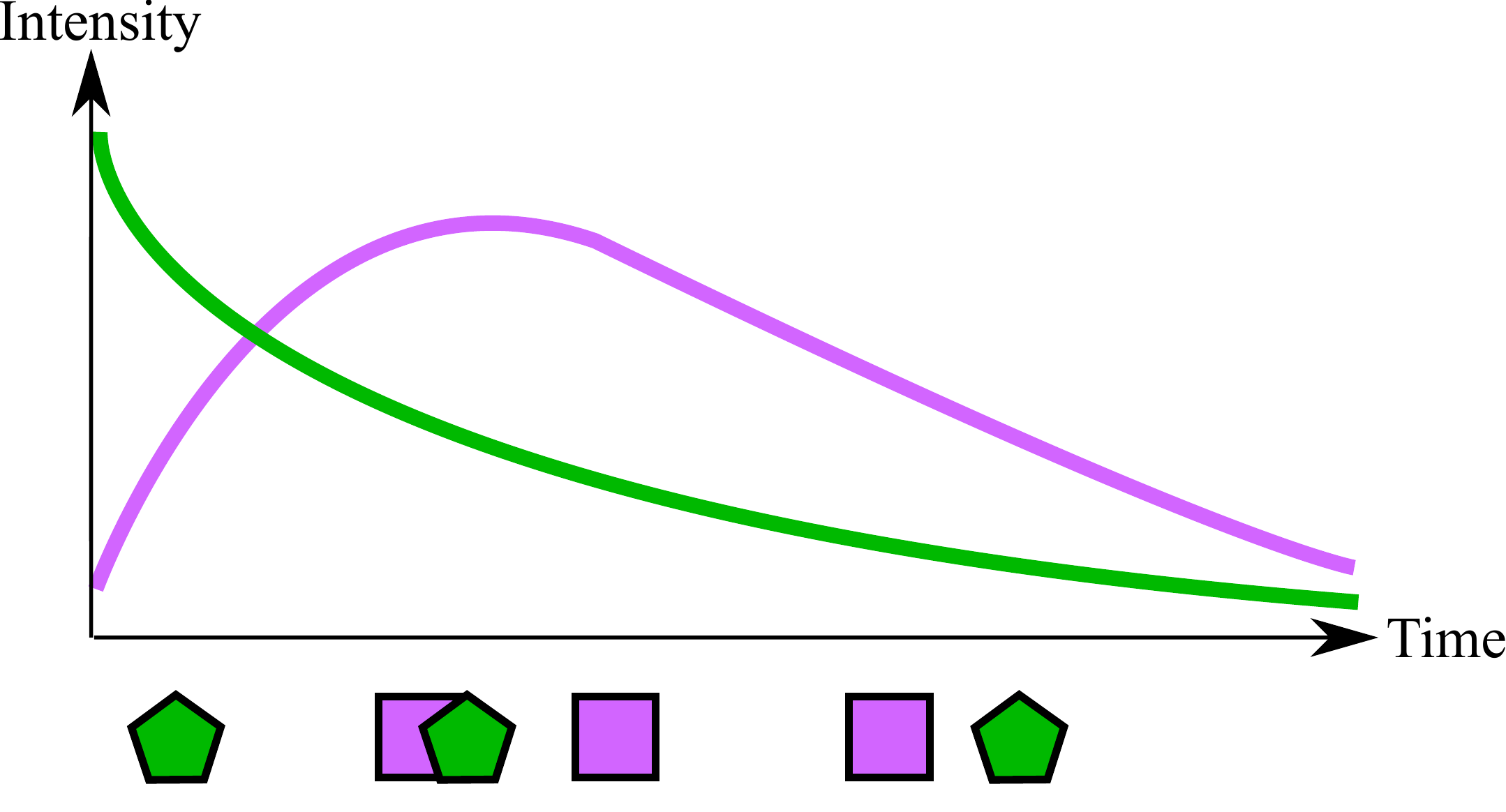}
\caption{
Sampling the next event, using the same visual notation as in~\cref{fig:event-example}. The $x$ axis shows a prefix of the infinite interval $(t_{i-1,\infty})$.
In the first graph, {\em gold} events are proposed from a homogeneous Poisson process with intensity $\lambda^*$ (gold straight line). In the second graph, the purple curve $\lambda_1^i$ randomly accepts some of these gold events, with probability $\lambda_1^i(t) / \lambda^*$ for the event at time $t$; here it accepts three of the ones shown and rejects the others. In the third graph, the surviving type-1 events (purple squares) are interleaved with the surviving type-2 events (green pentagons). The next event is the earliest one among these surviving candidates.  In practice, these sequences are constructed lazily so that we find only the earliest surviving event of each type.  This is possible because the inter-arrival times between gold proposed events are distributed as $\text{Exp}(\lambda^*)$, making it straightforward to enumerate any finite prefix of a random infinite gold sequence.
}\label{fig:thinning}
\end{figure}

If we wish to draw sequences from the self-modulating models of \ref{sec:self_modulate}, we can adopt the thinning algorithm~\citep{lewis-79-sim,liniger-09-hawkes} that is commonly used for the multivariate Hawkes process, as shown in~\cref{alg:thinning}.  We explain the algorithm here and illustrate its conception in \cref{fig:thinning}.

Suppose we have already sampled the first $i-1$ events.
The $K$ event types are now in a race to see who generates the next event.  (Typically, the winning type
will have relatively high intensity.)  In our model, that next event will join the multivariate event stream as $(k_i,t_i)$, whereupon it updates the LSTM state and thus modulates the subsequent intensities that will be used to sample event $i+1$.

How do we conduct the race?  For each event type $k$, let the function $\lambda_k^i: (t_{i-1},\infty) \rightarrow \Real_{\geq 0}$ map each time $t$ to the intensity $\lambda_k^i(t)$ that our model will define at time $t$
provided that event $i$ has not yet happened in the interval $(t_{i-1},t)$.  For each $k$ independently, we draw the time $t_{i,k}$ of the next event from the non-homogeneous Poisson process over $(t_{i-1},\infty)$ whose intensity function is $\lambda_k^i$.  We then take $t_i = \min_k t_{i,k}$ and $k_i = \argmin_k t_{i,k}$.  That is, we keep just the earliest of the $K$ events.  We cannot keep the rest because they are not correctly distributed according to the new intensities as updated by the earliest event.

But how do we draw the next event time $t_{i,k}$ from the non-homogeneous Poisson process given by $\lambda_k^i$?  Recall from \ref{sec:poisson_process} that a draw from such a point process is actually a whole {\em set} of times in $(t_{i-1},\infty)$: we will take $t_{i,k}$ to be the earliest of these.  In theory, this set is drawn by {\em independently} choosing at each time $t \in (t_{i-1},\infty)$, with infinitesimal probability proportional to $\lambda_k^i(t)$, whether an event occurs.  One could do this by {\em independently} applying rejection sampling at each time $t$: choose with larger probability $\lambda^*$ whether a ``proposed event'' occurs at time $t$, and if it does, accept the proposed event with probability only $\lambda_k^i(t)/{\lambda^*} \leq 1$. This is equivalent to simultanously drawing a set of proposed times from a {\em homogenous} Poisson process with constant rate $\lambda^*$, and then ``thinning'' that proposed set, as illustrated in \cref{fig:thinning}.   This approach helps because it is easy to draw from the homogenous process: the intervals between successive proposed events are IID $\Exp(\lambda^*)$, so it is easy to sample the events in sequence.  The inner {\bf repeat} loop in \cref{alg:thinning} lazily carries out just enough of this infinite homogenous draw from $\lambda^*$ to determine the time $t_{i,k}$ of the earliest {\em accepted} event, which is the earliest event in the non-homogeneous draw from $\lambda_k^i$, as desired.

Finally, how do we construct the upper bound $\lambda^*$ on $\lambda_k^i$?
Recall that both of our self-modulating models (\cref{eqn:hawkes_inhib_a,eqn:hawkes_mod_a})
define $\lambda_k^i = f_k(\tilde{\lambda}_k^i)$, where $f_k$ is monotonically non-decreasing.  In both cases, $\tilde{\lambda}_k^i$ is a sum of {\em bounded} functions on $(t_{i-1},\infty)$
(\cref{eqn:hawkes_inhib_b,eqn:hawkes_mod}).  In other words, we can express $\tilde{\lambda}_k^i(t)$ as $\mu + g_1(t) + \cdots + g_n(t)$.  We can therefore replace each $g$ function by its upper bound to obtain $\lambda^* = f_k(\mu + \max_t g_1(t) + \cdots + \max_t g_n(t))$, in which the argument to $f_k$ is a finite constant.

Specifically, in \cref{eqn:hawkes_inhib_b}, each summand $\alpha_{k_h,k} \exp (-\delta_{k_h,k} (t-t_i) )$
is upper-bounded by $\max(\alpha_{k_h,k}, 0)$.  In \cref{eqn:hawkes_mod}, each summand $w_{kd} h_d(t) = w_{kd} \cdot o_{id} \cdot (2\sigma(2c_d(t))-1)$ is upper-bounded by $\max_{c \in \{c_{id},\bar{c}_{id}\}} w_{kd} \cdot o_{id} \cdot (2\sigma(2c)-1)$.  Note that the coefficients $\alpha_{k_i,k}$ and $w_{kd}$ may be either positive or negative.

\begin{algorithm}[tb]
	\caption{Data Simulation (thinning algorithm)}
	\label{alg:thinning}
	\begin{algorithmic}
	\State {\bfseries Input:} interval $[0,T]$; model parameters
        \State $t_0\gets 0$; $i\gets 1$
		\While{$t_{i-1} < T$} \Comment{draw event $i$, as it might fall in [0,T]}
			\For{$k=1$ {\bfseries to} $K$} \Comment{draw ``next'' event of each type}
				\State find upper bound $\lambda^* \geq \lambda_k^i(t)$ for all $t \in (t_{i-1},\infty)$
				\State $t \gets t_{i-1}$
					\Repeat
						\State draw $\Delta \sim \Exp(\lambda^*)$, $u \sim \Uniform(0,1)$
						\State $t \pluseq \Delta$ \Comment{time of next proposed event}
					\Until{$u\lambda^* \leq \lambda_k^i(t)$} \Comment{accept proposal with prob $\frac{\lambda_k^i(t)}{\lambda^*}$}
                                \State $t_{i,k} \gets t$
			\EndFor
			\State $t_{i} \gets \min_{k} t_{i,k}$; $k_{i} \gets \argmin_{k} t_{i,k}$ \Comment{earliest event wins}
                        \State $i \gets i+1$
		\EndWhile
        \State \textbf{return} $(k_1,t_1),\ldots(k_{i-1},t_{i-1})$
\end{algorithmic}
\end{algorithm}

While \cref{alg:thinning} is classical and intuitive, we also implemented a more efficient variant.  Instead of drawing the next event from each of $K$ {\em different} non-homogeneous Poisson processes and keeping the earliest, we can construct a {\em single} non-homogenous Poisson process with aggregate intensity function $\lambda^i(t)=\sum_{k=1}^{K} \lambda^i_{k}(t)$ over $(t_{i-1},\infty)$.  An upper bound $\lambda^*$ on this aggregate function can be obtained by summing the upper bounds on the individual $\lambda^i_k$ functions.  We then use the thinning algorithm only to sample the next event time $t_i$ from this aggregate process $\lambda^i$.  Finally, we ``disaggregate'' by choosing $k_i$ from the distribution $p(k \mid t_i) = \lambda_{k}^i(t_i)/\lambda^i(t_i)$.\footnote{In practice, acceptance and disaggregation can be combined into a single step.  That is, each successive event $t$ proposed from the homogeneous $\text{Poisson}(\lambda^*)$ process is either kept as type $k$, with probability
  $\lambda_{k}^i(t)/\lambda^*$, or rejected, with probability $1 - \lambda^i(t)/\lambda^*$.
  If it is accepted, we have found our next event $(k_i,t_i)$.  If it is rejected, we increment $t$ by $\Delta \sim \text{Exp}(\lambda^*)$ to get the next proposed event.}
  This is equivalent to \cref{alg:thinning}.  In terms of \cref{fig:thinning}, this more efficient version enumerates a gold sequence that is the union of the $K$ gold sequences, and stops with the first accepted gold event.  Thus, whereas \cref{fig:thinning} had to propose two type-1 events in order to get the first accepted type-1 event (the leftmost purple event), the more efficient version would not have had to spend time proposing either of those, because an earlier proposed event (the leftmost green event) had already been accepted and determined to be of type 2.

\section{Experimental Details}
\label{sec:experimental_details}

In this appendix, we elaborate on the details of data generation, processing, and experimental results.

\subsection{Dataset Statistics}
\label{sec:data_stats}

\Cref{tab:stats_dataset} shows statistics about each dataset that we use in this paper.

\begin{table*}[t]
\begin{center}
\begin{small}
\begin{sc}
\begin{tabularx}{1.00\textwidth}{l *{1}{S}*{3}{R}*{3}{S}}
\toprule
Dataset & \multicolumn{1}{r}{$K$} & \multicolumn{3}{c}{\# of Event Tokens} & \multicolumn{3}{c}{Sequence Length} \\
\cmidrule(lr){3-8}
  &  & Train & Dev & Test & Min & Mean & Max \\
\midrule
Synthetic & $5$ & $\approx480449$ & $\approx 60217$ & $\approx 60139$ & $20$ & $\approx 60$ & $100$ \\
Retweets & $3$ & $1739547$ & $215521$ & $218465$ & $50$ & $109$ & $264$ \\
MemeTrack & $5000$ & $93267$ & $14932$ & $15440$ & $1$ & $3$ & $31$ \\
MIMIC-II & $75$ & $\approx 1946$ & $\approx 228$ & $\approx 245$ & $2$ & $4$ & $33$ \\
StackOverflow & $22$ & $\approx 343998$ & $\approx 39247$ & $\approx 97168$ & $41$ & $72$ & $736$ \\
Financial & $2$ & $\approx 298710$ & $\approx 33190$ & $\approx 82900$ & $829$ & $2074$ & $3319$ \\
\bottomrule
\end{tabularx}
\end{sc}
\end{small}
\end{center}
 \caption{
 	Statistics of each dataset.  We write ``$\approx N$'' to indicate that $N$ is the average value over multiple splits of one dataset (MIMIC-II, Stack Overflow, Financial Transaction); the variance is small in each such case.
}
\label{tab:stats_dataset}
\end{table*}

\begin{table*}[t]
\begin{center}
\begin{small}
\begin{sc}
\begin{tabularx}{0.95\linewidth}{l*{2}C*{3}{R}}
\toprule
Dataset & \multicolumn{1}{c}{$K$} & \multicolumn{1}{c}{$D$} & \multicolumn{3}{c}{\# of Model Parameters}\\
\cmidrule(lr){4-6}
  & & & SE-MPP & D-SM-MPP & N-SM-MPP \\
\midrule
Synthetic & $5$ & $256$ & $55$ & $60$ & $922117$ \\
Retweets & $3$ & $256$ & $21$ & $24$ & $921091$ \\
MemeTrack & $5000$ & $64$ & $50005000$ & $50010000$ & $702856$ \\
\bottomrule
\end{tabularx}
\end{sc}
\end{small}
\end{center}
\caption{
	Size of each trained model on each dataset. The number of parameters of neural Hawkes process is followed by the number of hidden nodes $D$ in its LSTM (chosen automatically on dev data).
}
\label{tab:model_size}
\end{table*}

\subsection{Training Details}
\label{sec:training_details}
We used a single-layer LSTM~\citep{graves-12} in \cref{sec:neural_hawkes_process}, selecting the number of hidden nodes from a small set $\{64, 128, 256, 512, 1024\}$ based on the performance on the dev set of each dataset.
We empirically found that the model performance is robust to these hyperparameters.

When estimating integrals with Monte Carlo sampling, $N$ is the number of sampled negative observations in \cref{alg:sample_integral}, while $I$ is the number of positive observations.  In practice, setting $N=I$ was large enough for stable behavior, and we used this setting during training.  For evaluation on dev and test data, we took $N=10\,I$ for extra accuracy, or $N=I$ when $I$ was very large.

For learning, we used the Adam algorithm with its default settings \citep{kingma-15}.  Adam is a stochastic gradient optimization algorithm that continually adjusts the learning rate in each dimension based on adaptive estimates of low-order moments.  Our training objective was unregularized log-likelihood.\footnote{L$_2$ regularization did not appear helpful in pilot experiments, at least for our dataset size and when sharing a single regularization coefficient among all parameters.}  We initialized the Hawkes process parameters and $s_k$ scale factors to 1,
and all other non-LSTM parameters (\cref{sec:params}) to small random values from ${\cal N}(0,0.01)$.
We performed early stopping based on log-likelihood on the held-out dev set.

\subsection{Model Sizes}
\label{sec:model_size}

The size of each trained model on each dataset is shown in~\cref{tab:model_size}. Our neural model has many parameters for expressivity, but it actually has considerably fewer parameters than the other models in the large-$K$ setting (MemeTrack).

\subsection{Pilot Experiments on Simulated Data}
\label{sec:simulated_details}

Our hope is that the neural Hawkes process is a flexible tool that can be used to fit naturally occurring data.  As mentioned in \cref{sec:simulated}, we first checked that we could successfully fit data generated from {\em known} distributions.  That is, when the generating distribution actually fell within our model family, could our training procedure recover the distribution in practice?  When the data came from a decomposable process, could we nonetheless train our neural process to fit the distribution well?

We used the thinning algorithm (\cref{sec:thinning}) to sample event streams from different processes with randomly generated parameters: (a) a standard Hawkes process (SE-MPP, \cref{sec:SE-MPP}), (b) our decomposable self-modulating process (D-SM-MPP, \cref{sec:D-SM-MPP}), (c) our neural self-modulating processes (N-SM-MPP, \cref{sec:N-SM-MPP}).  We then tried to fit each dataset with all these models.\footnote{Details of data generation can be found in~\cref{sec:simulated_details}.}

\begin{figure}
\includegraphics[width=0.32\linewidth]{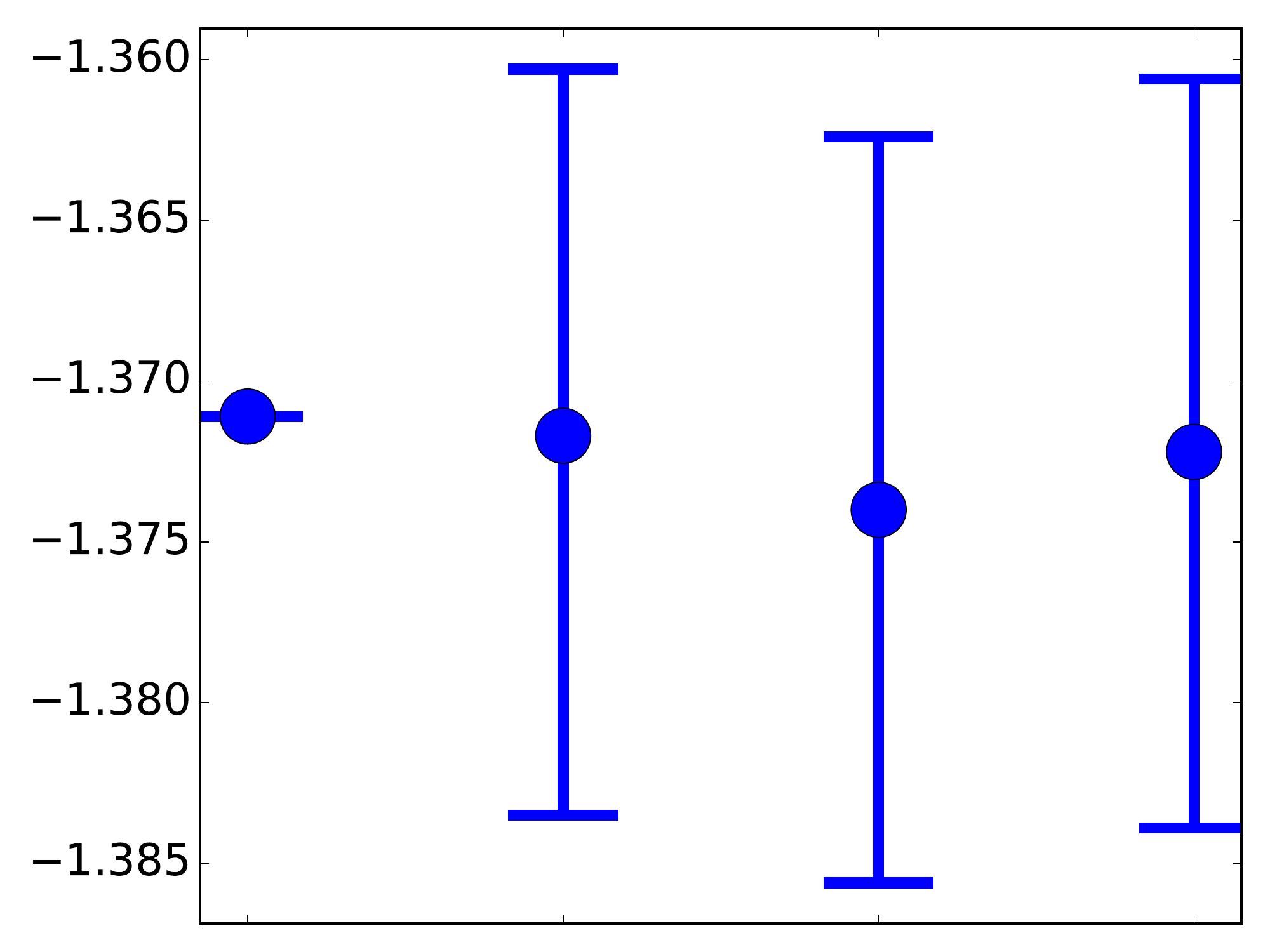}
\includegraphics[width=0.32\linewidth]{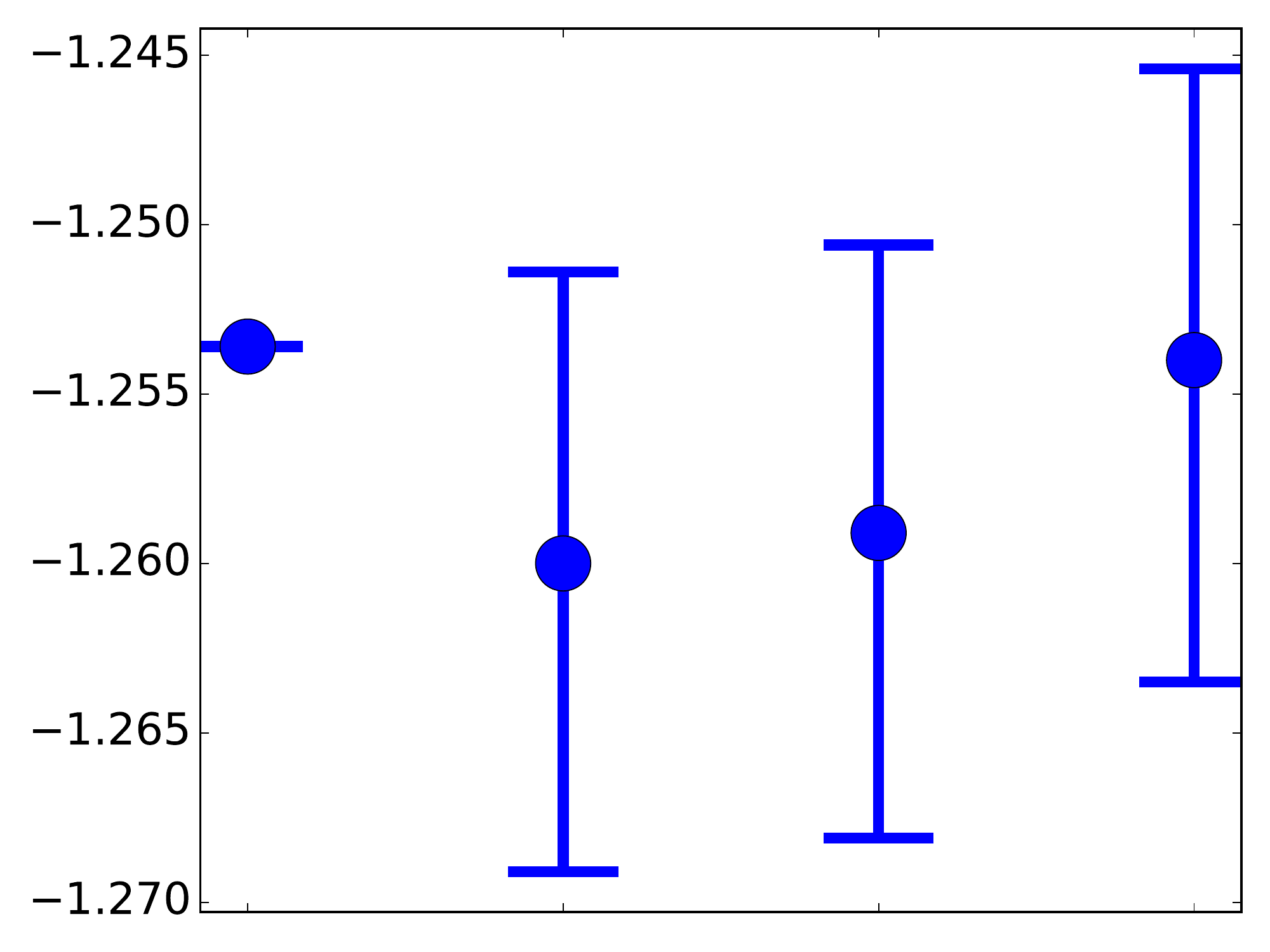}
\includegraphics[width=0.32\linewidth]{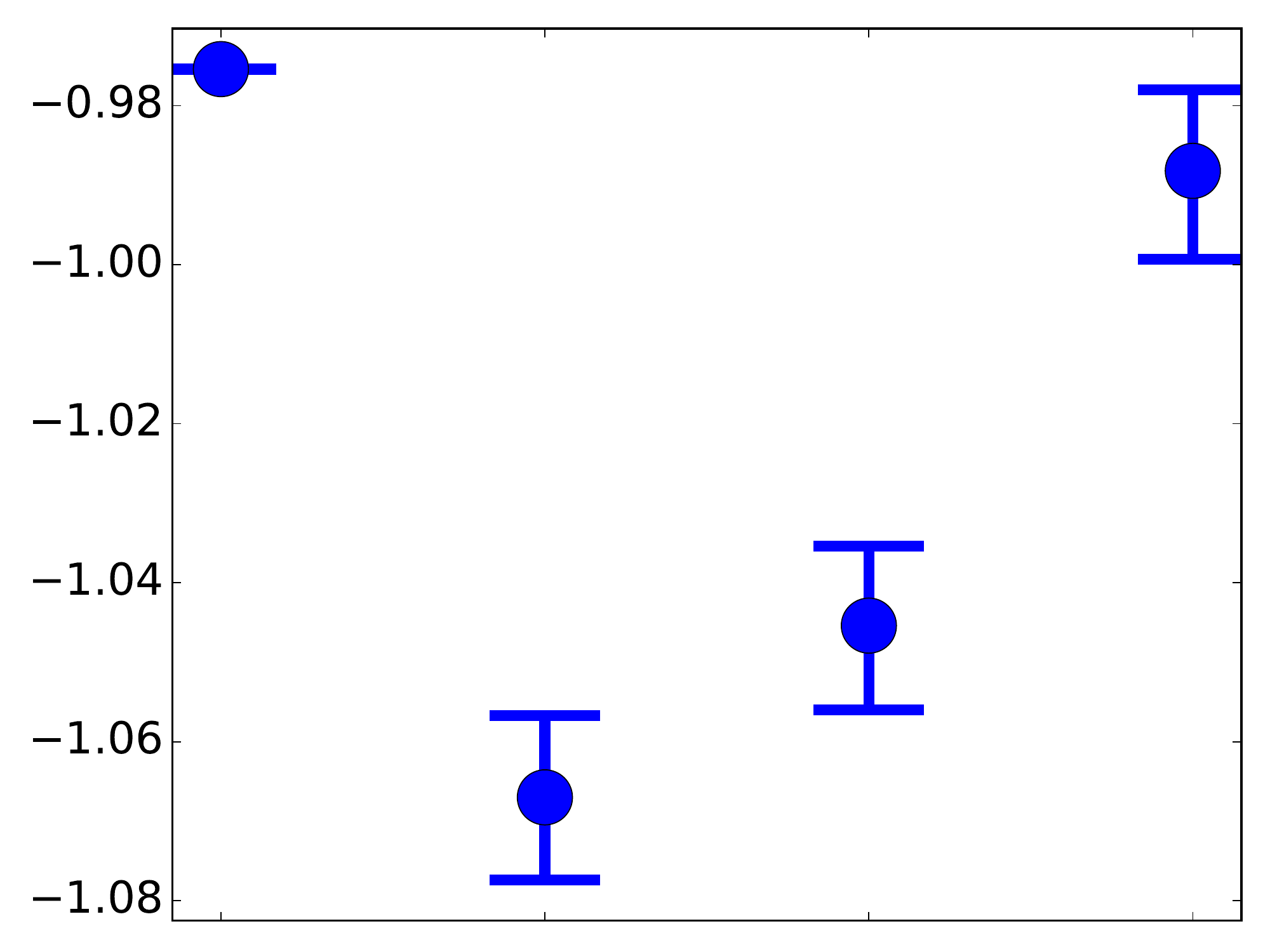}

\includegraphics[width=0.32\linewidth]{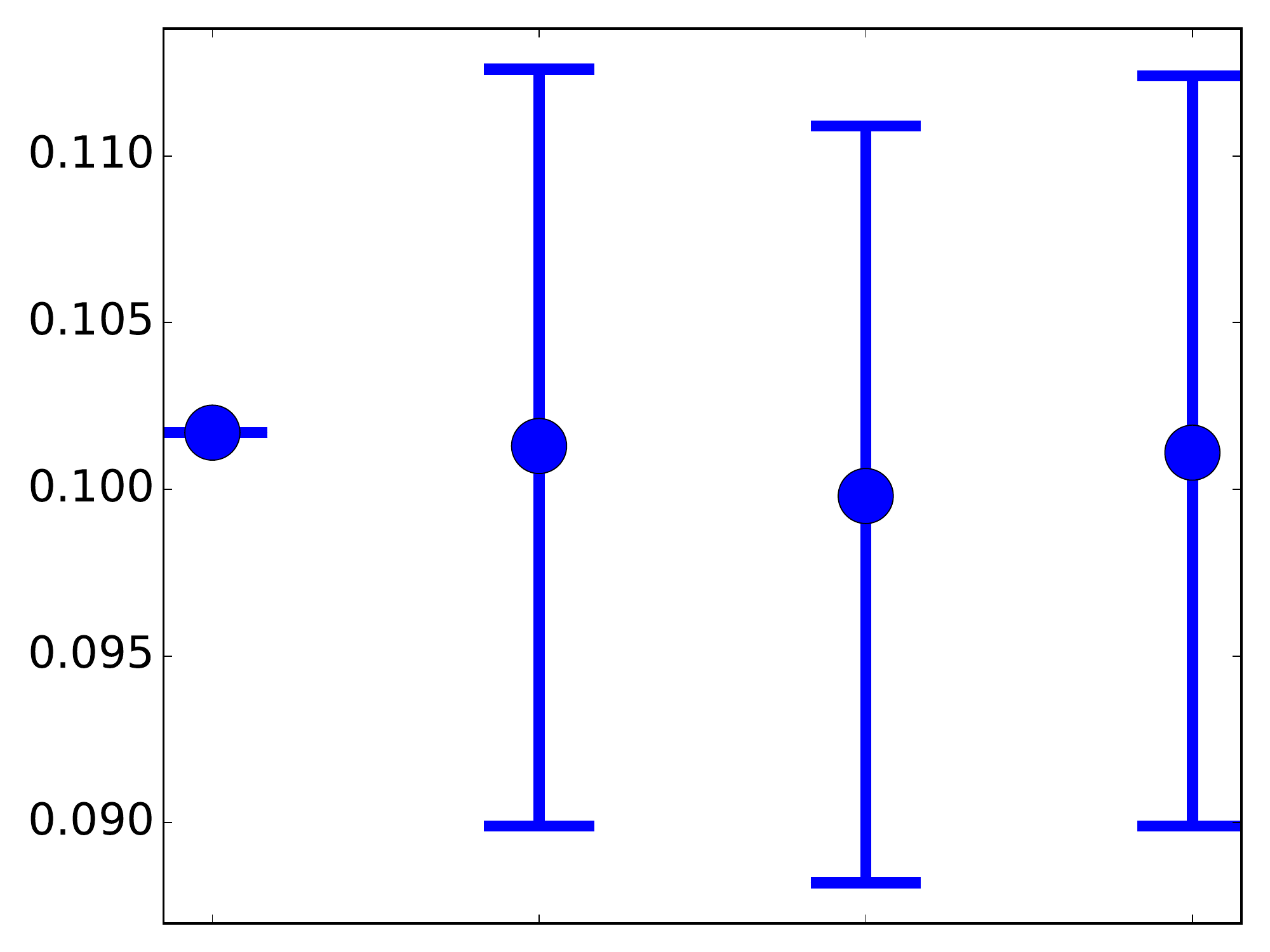}
\includegraphics[width=0.32\linewidth]{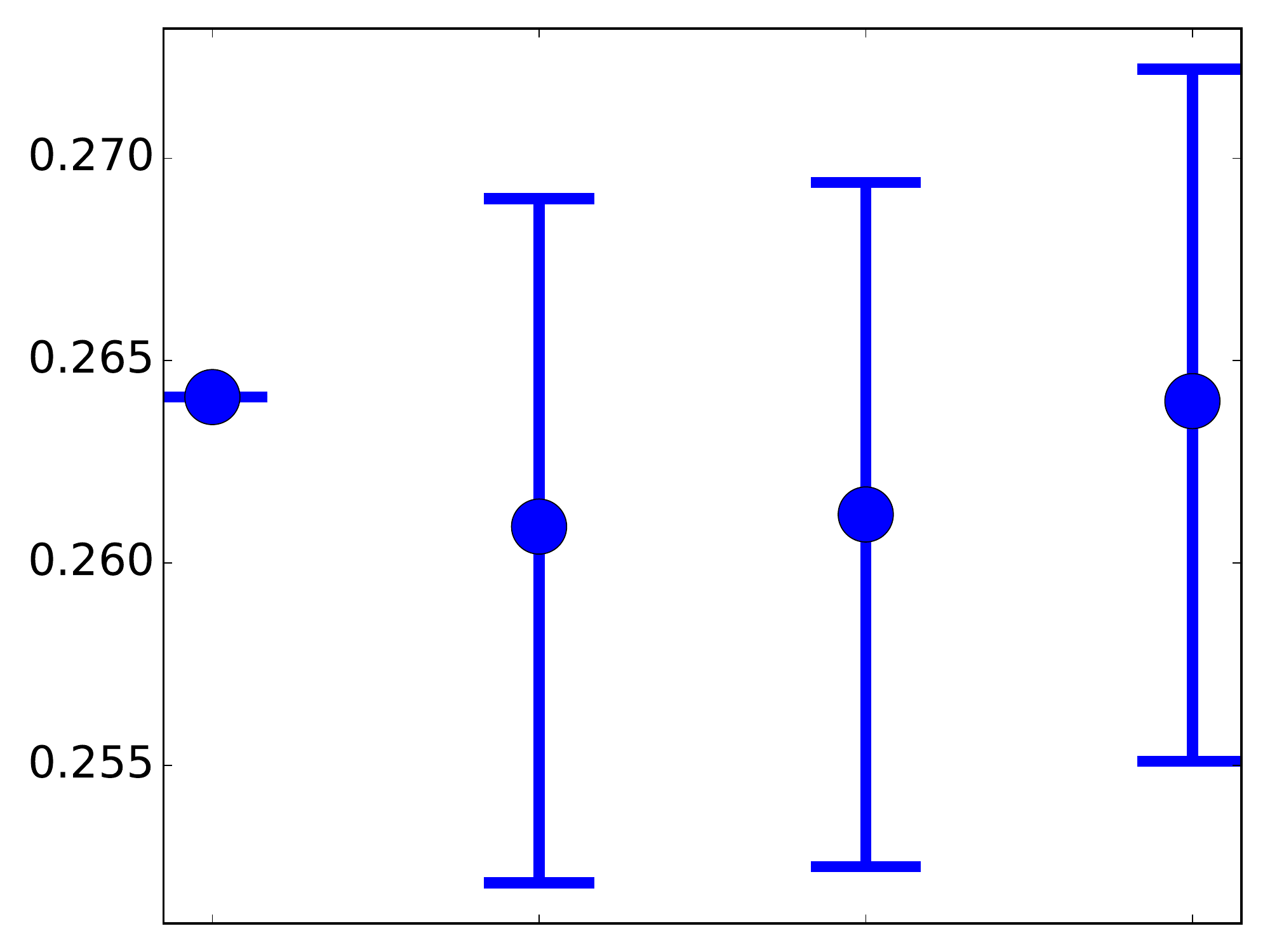}
\includegraphics[width=0.32\linewidth]{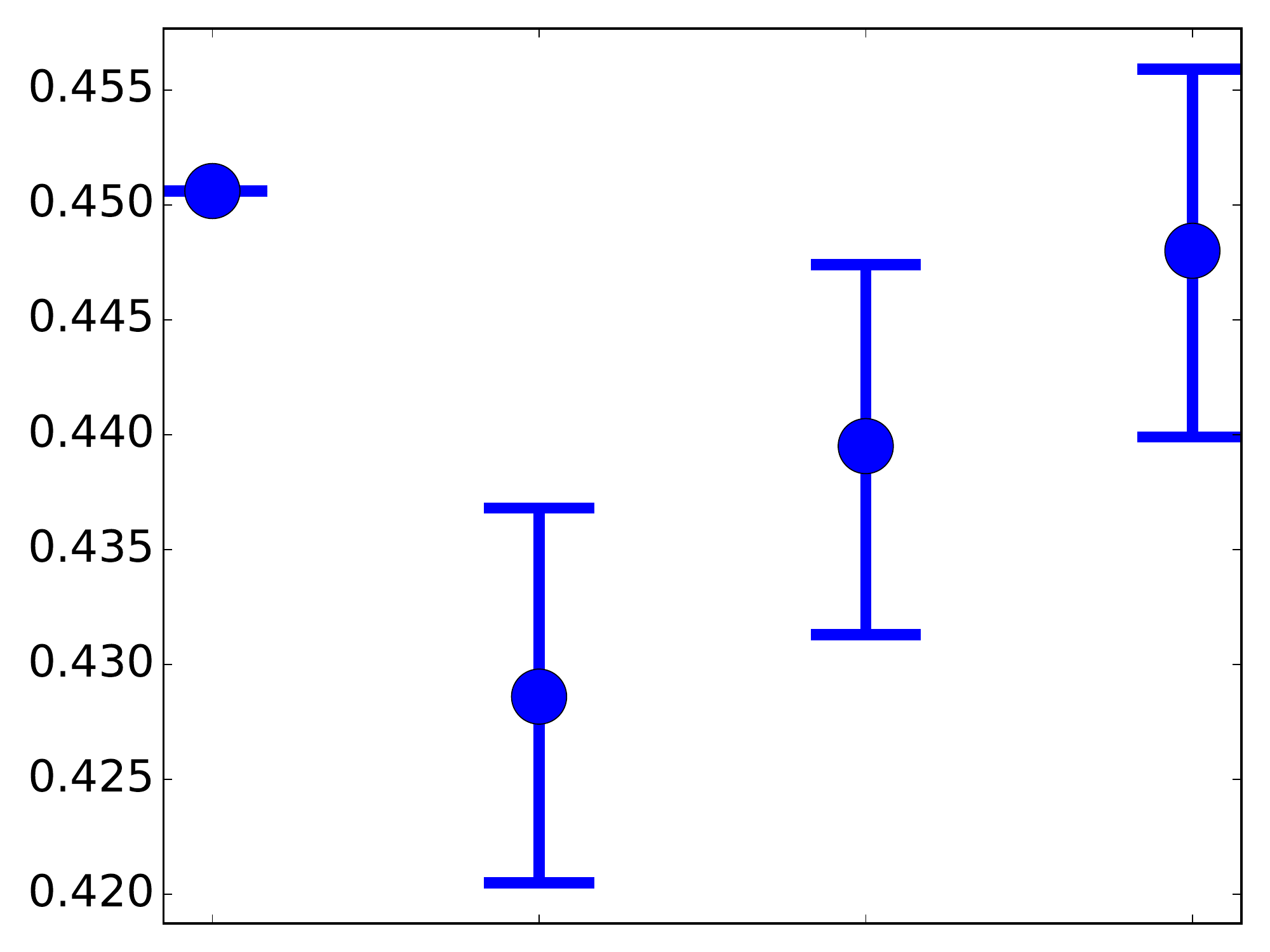}

\includegraphics[width=0.32\linewidth]{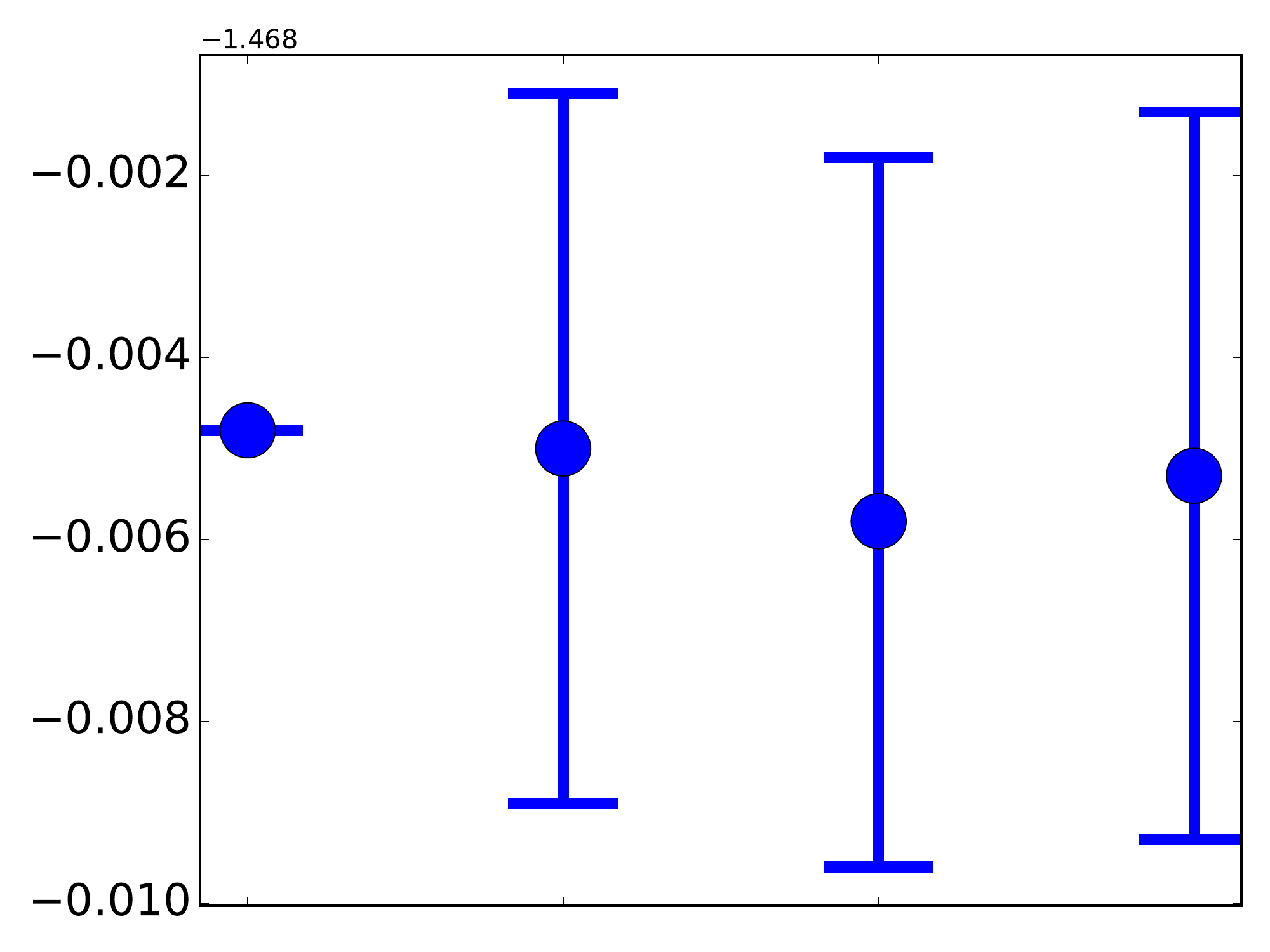}
\includegraphics[width=0.32\linewidth]{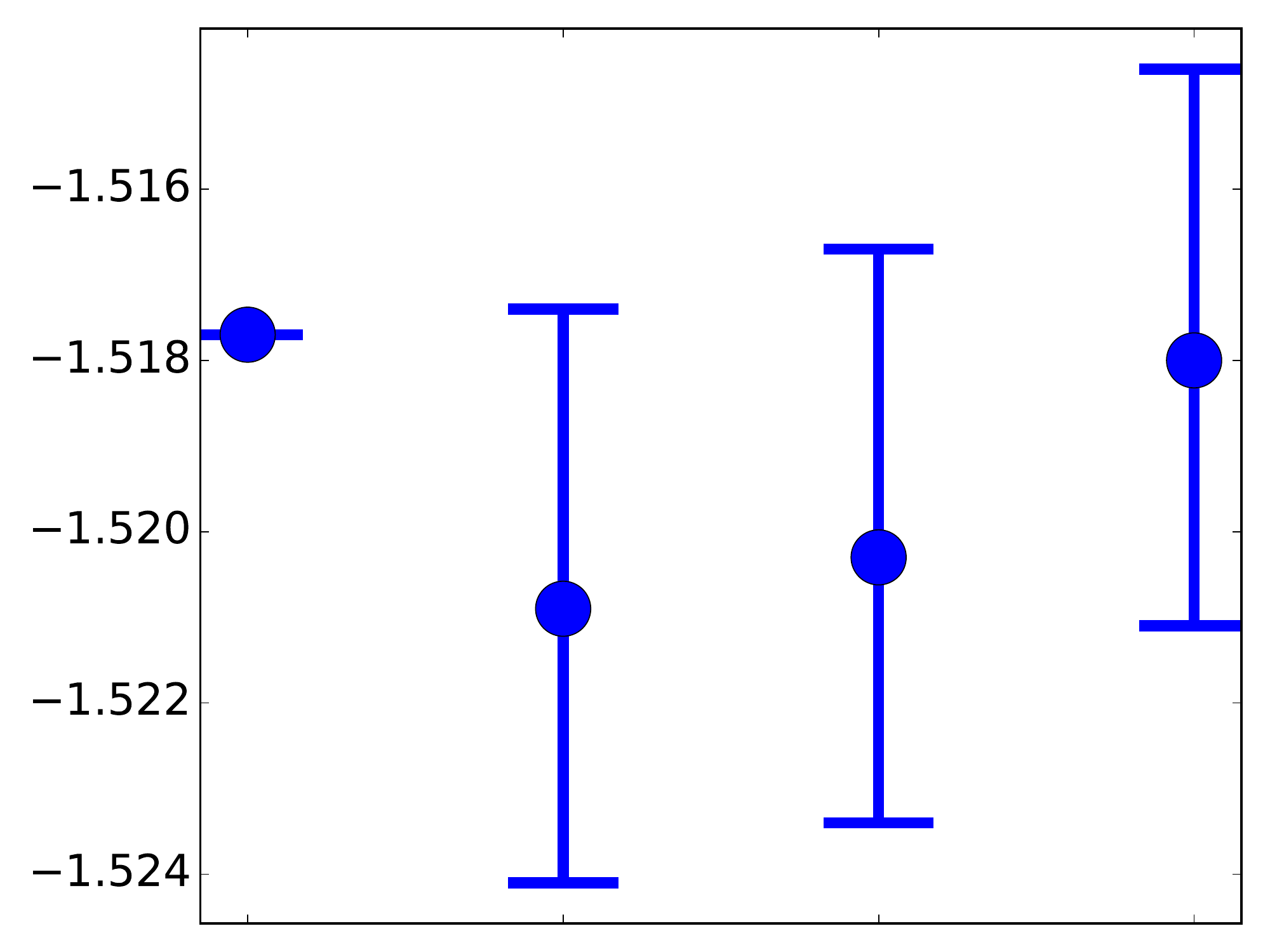}
\includegraphics[width=0.32\linewidth]{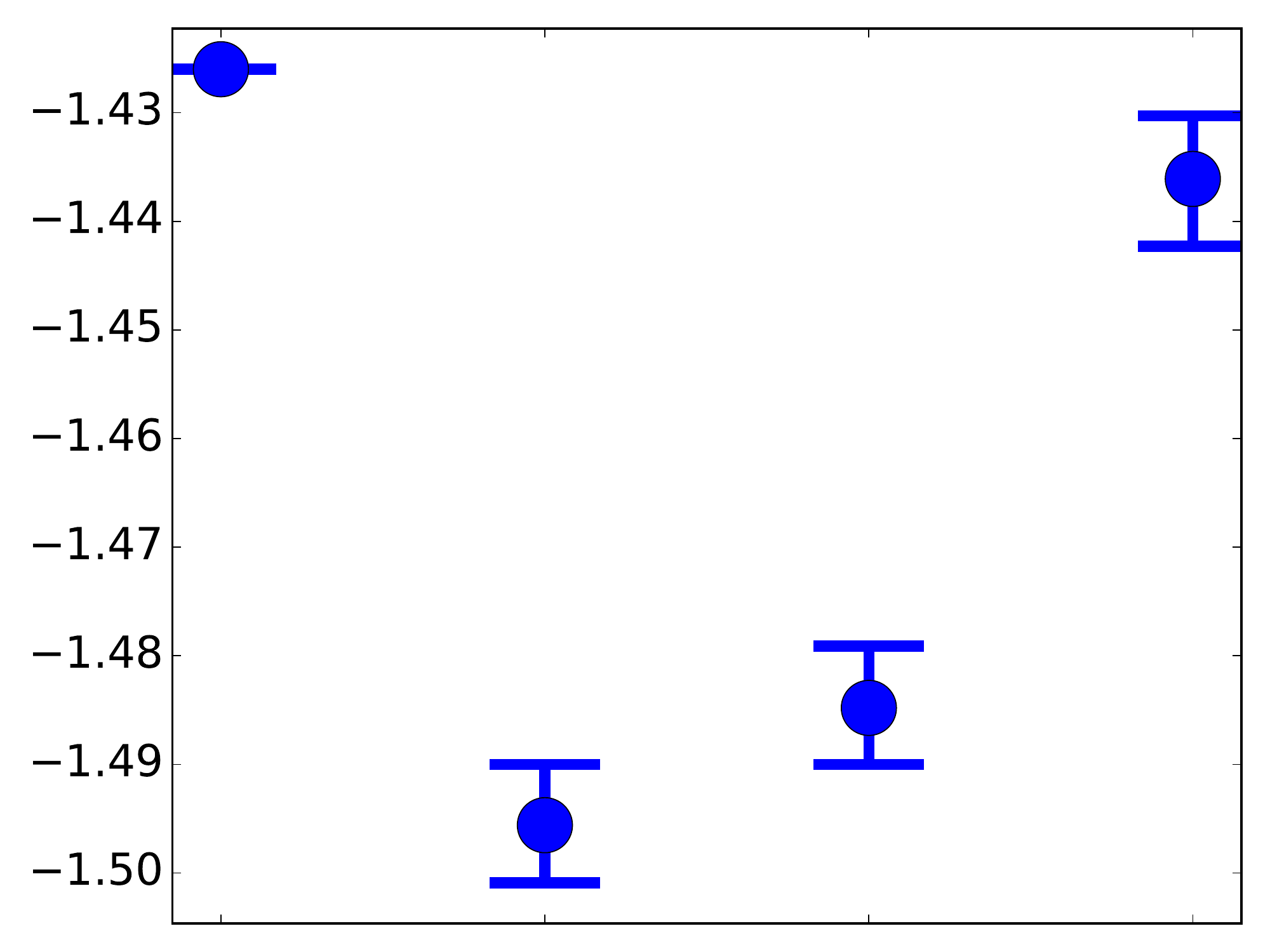}
\caption{
Log-likelihood (reported in nats per event) of each model on held-out synthetic data. Rows (top-down) are log-likelihood on the entire sequence, time interval, and event type. On each row, the figures (from left to right) are datasets generated by SE-MPP, D-SM-MPP and N-SM-MPP. In each figure, the models (from left to right) are Oracle, SE-MPP, D-SM-MPP and N-SM-MPP.  Larger values are better.  Note that log-likelihood for continuous variables can be positive, since it uses the log of a probability density that may be $> 1$.
}\label{fig:sim_results_details}
\end{figure}

The results are shown in~\cref{fig:sim_results_details}. We found that all models were able to fit the (a) and (b) datasets well with no statistically significant difference among them, but that the (c) models were substantially and significantly better at fitting the (c) datasets.  In all cases, the (c) models were able to obtain a low KL divergence from the true generating model (the difference from the oracle column).
This result suggests that the neural Hawkes process may be a wise choice: it introduces extra expressive power that is sometimes necessary and does not appear (at least in these experiments) to be harmful when it is not necessary.

We used \cref{alg:thinning} to sample event streams from three different processes with randomly generated parameters: (a) a standard Hawkes process (SE-MPP), (b) our decomposable self-modulating process (D-SM-MPP), (c) our neural self-modulating processes (N-SM-MPP).  We then tried to fit each dataset with all these models.

For each dataset, we took $K=5$ as the number of event types.
To generate each event sequence, we first chose the sequence length $I$ (number of event tokens) uniformly from $\{ 20, 21, 22, \ldots, 100 \}$ and then used the thinning algorithm to sample the first $I$ events over the interval $[0,\infty)$.  For subsequent training or testing, we treated this sequence (appropriately) as the complete set of events observed on the interval $[0,T]$ where $T = t_I$, the time of the last generated event.
For each dataset, we generate $8000$, $1000$ and $1000$ sequences for the training, dev, and test sets respectively.

For SE-MPP, we sampled the parameters as $\mu_k \sim \Uniform[0.0, 1.0]$, $\alpha_{j,k} \sim \Uniform[0.0, 1.0]$, and $\delta_{j,k} \sim \Uniform[10.0, 20.0]$.
The large decay rates $\delta_{j,k}$
were needed to prevent the intensities from blowing up as the sequence accumulated more events.
For D-SM-MPP, we sampled the parameters as $\mu_k \sim \Uniform[-1.0, 1.0]$, $\alpha_{j,k} \sim \Uniform[-1.0, 1.0]$, and $\delta_{j,k} \sim \Uniform[10.0, 20.0]$.
For N-SM-MPP, we sampled parameters from $\Uniform[-1.0, 1.0]$.

The results are shown in \cref{fig:sim_results_details}, including log-likelihood (reported in nats per event) on the sequences and the breakdown of time interval and event types.

Another interesting question is whether the trained neural Hawkes model accurately predicts the real-valued {\em intensities}, since for the synthetic data we actually know the intensities.  This is a more direct evaluation of whether the model is accurately recovering the dynamics of the underlying generative process.  Here we compared only SE-MPP and N-SM-MPP.

All types behaved similarly, so we report only averages over the $K$ types.  For both processes (a) and (c), the true intensity's variance was about 30\% of the squared mean intensity.  Thus, the intensity changes enough over time that predicting it at particular times is not a trivial challenge.  To determine how well a model predicted the true intensity function, we measured the mean squared error (MSE) of predicted intensity at a large sample of times in the held-out test seqs, and report the MSE here as a percentage of the {\em variance} of the true intensity.  By this construction, a simple baseline of predicting each event type's mean intensity at all times would get 100\% MSE.

Both the Hawkes and neural-Hawkes models predict the Hawkes intensities (a) accurately, at 1\% MSE.  This is similar to the leftmost column of \cref{fig:sim_results_details}, where both models essentially achieved oracle performance.  By contrast, for the complex neural Hawkes intensities (c), the neural Hawkes model achieves 9\% MSE (still quite good) whereas Hawkes does far worse at 70\% MSE.  This is similar to the rightmost column of \cref{fig:sim_results_details}, where the neural Hawkes model approached oracle performance but the Hawkes model did much worse.

\subsection{Retweet Dataset Details}
\label{sec:retweet_details}
The Retweets dataset (\cref{sec:retweet}) includes $166076$ retweet sequences, each corresponding to some original tweet.  Each retweet event is labeled with the retweet time relative to the original tweet creation, so that the time of the original tweet is 0.  (The original tweet serves as the beginning-of-stream (\bos) marker as explained in \cref{sec:bos}.)  Each retweet event is also marked with the number of followers of the retweeter.  As usual, we assume that these $166076$ streams are drawn independently from the same process, so that retweets in different streams do not affect one another.

Unfortunately, the dataset does not specify the identity of each retweeter, only his or her popularity.  To distinguish different kinds of events that might have different rates and different influences on the future, we divide the events into $K=3$ types: retweets by ``small,'' ``medium'' and ``large'' users.  Small users have fewer than $120$ followers ($50\%$ of events), medium users have fewer than $1363$ ($45\%$ of events), and the rest are large users ($5\%$ events).  Given the past retweet history, our model must learn to predict how soon it will be retweeted again and how popular the retweeter is (i.e., which of the three categories).

We randomly sampled disjoint train, dev and test sets with $16000$, $2000$ and $2000$ sequences respectively. We truncated sequences to a maximum length of 264, which affected 20\% of them.  For computing training and test likelihoods, we treated each sequence as the complete set of events observed on the interval $[0,T]$, where $0$ denotes the time of the original tweet (which is not included in the sequence) and $T$ denotes the time of the last tweet in the (truncated) sequence.

\cref{fig:curve_retweet_breakdown} shows the learning curves of all the models, broken down by the log-probabilities of the event types and the time intervals separately.
The scatterplot \cref{fig:scatter_retweet_copy} is a copy of \cref{fig:scatter_retweet}, and \cref{fig:scatter_retweet_with_hawkes_breakdown} breaks down the log-likelihood by event type and time interval.

\begin{figure}
\begin{center}
\minipage[c]{0.74\textwidth}
	\includegraphics[width=0.32\textwidth]{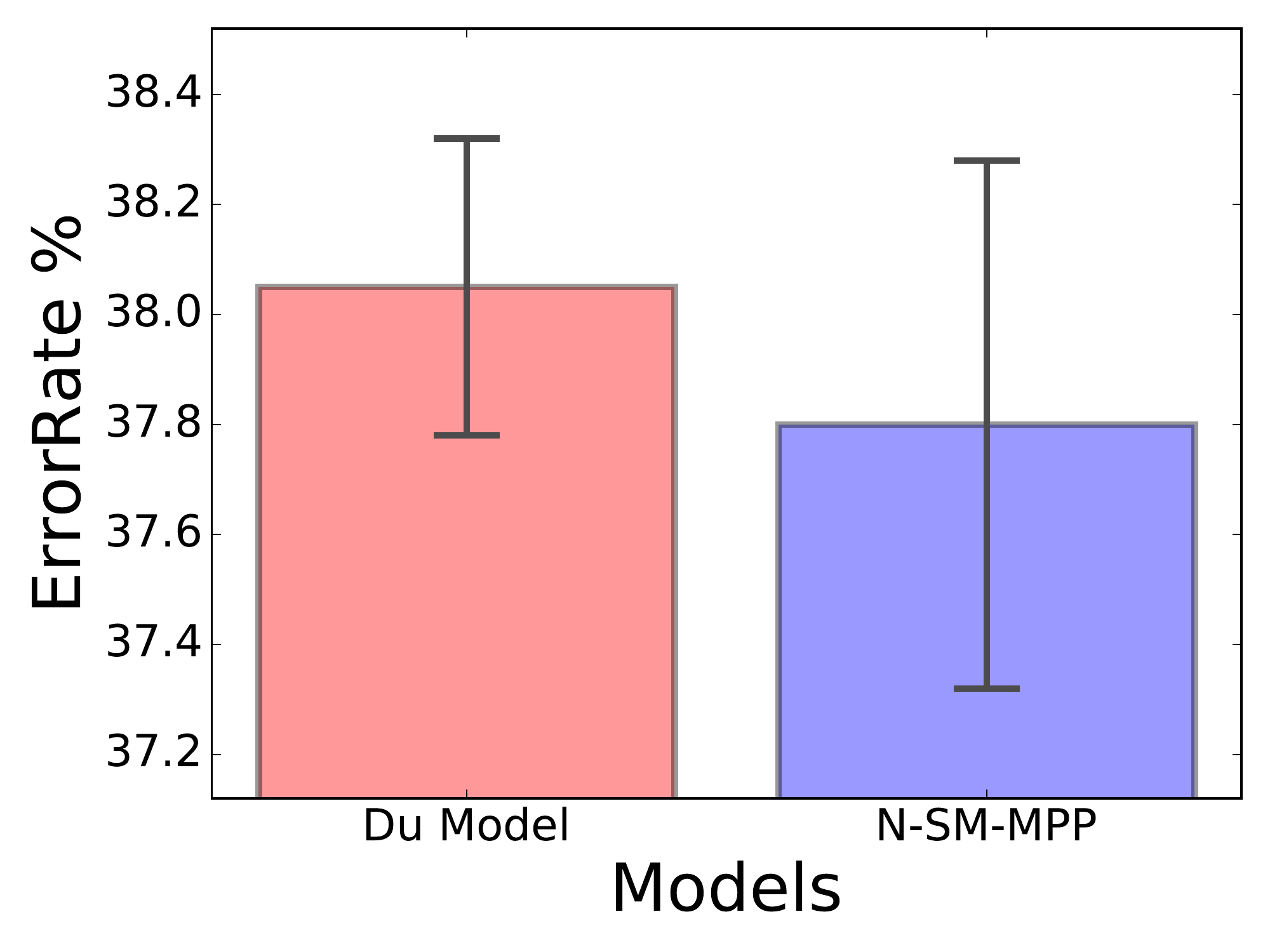}
	\includegraphics[width=0.32\textwidth]{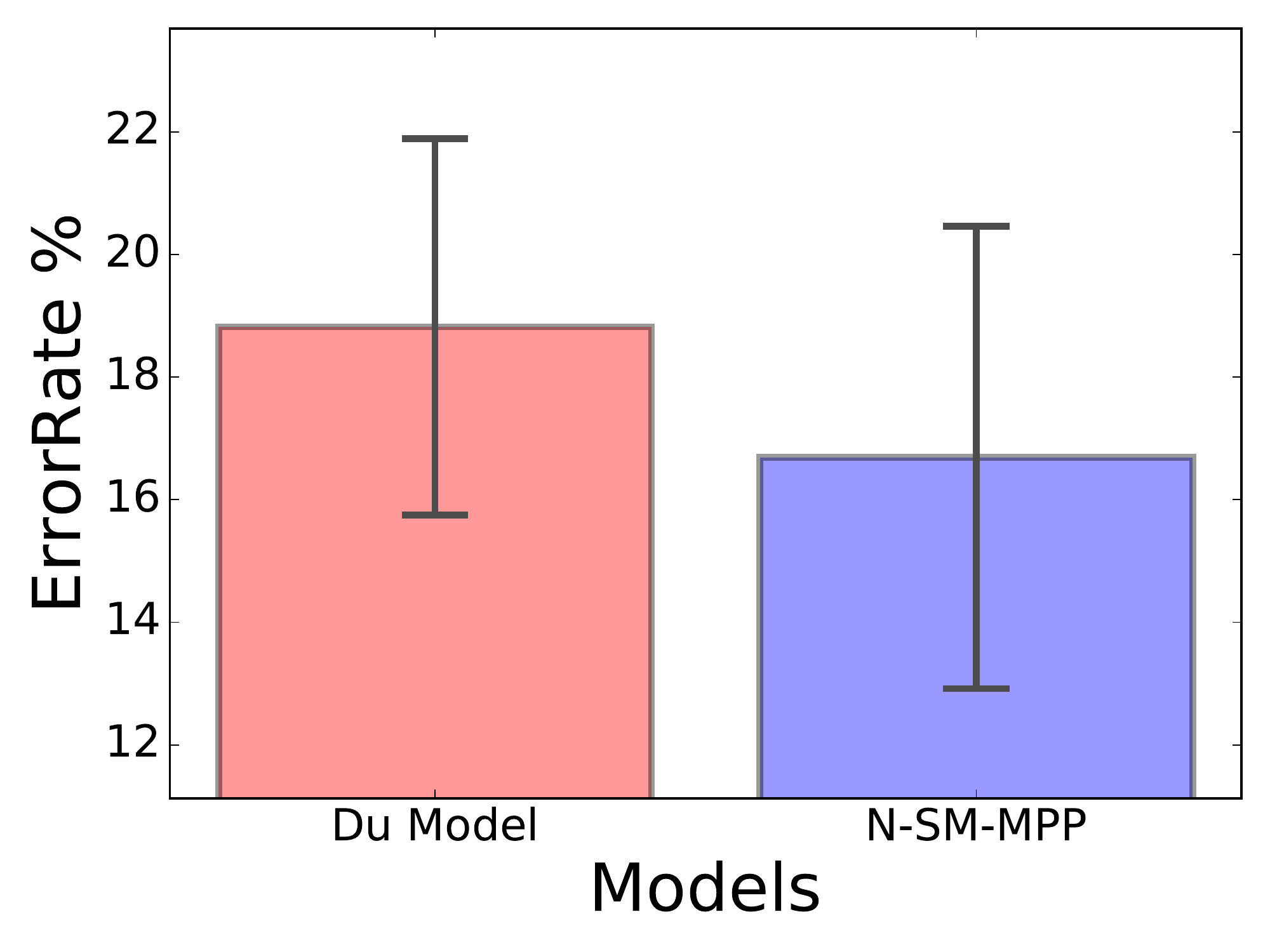}
	\includegraphics[width=0.32\textwidth]{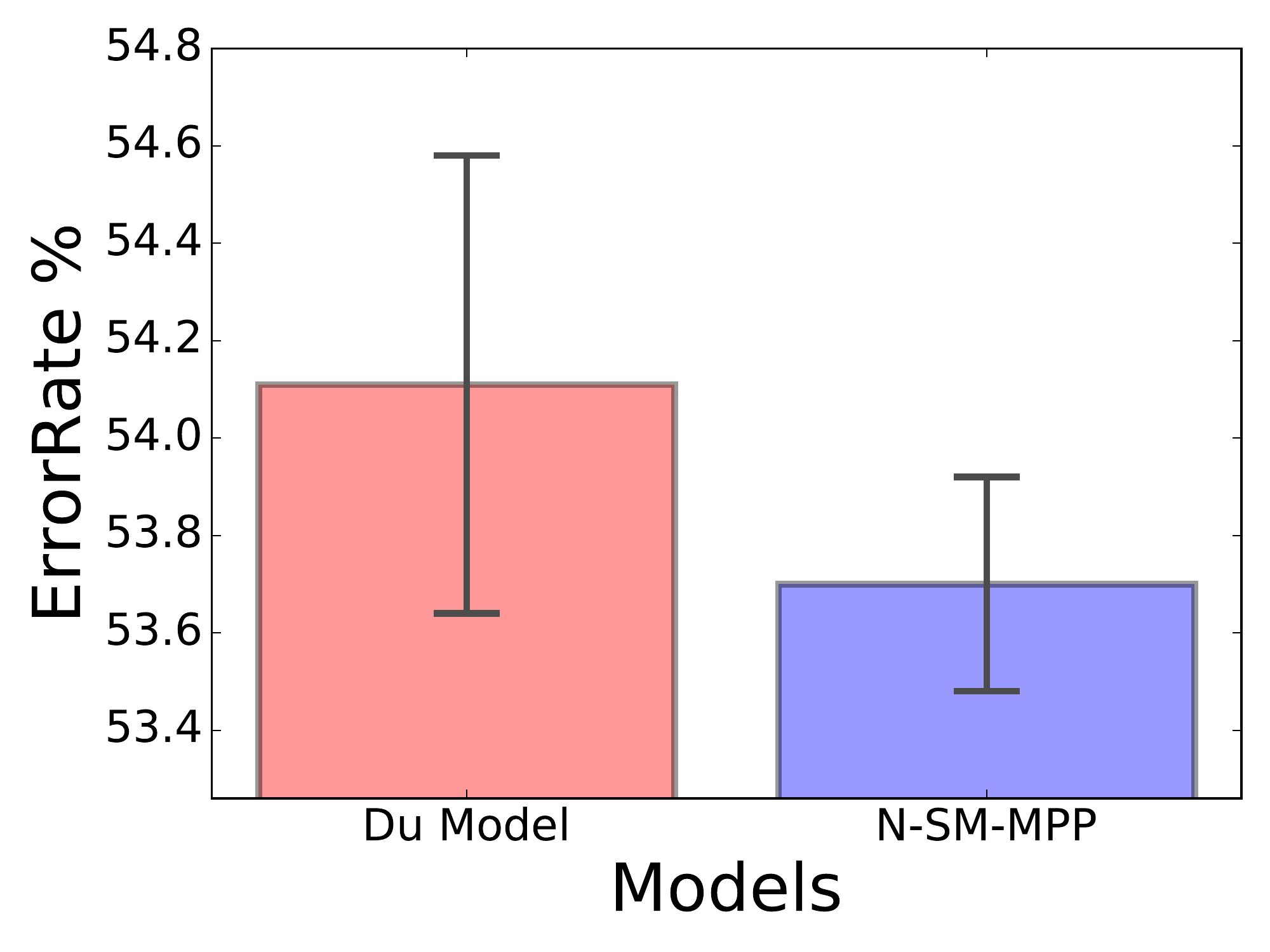}

	\includegraphics[width=0.32\textwidth]{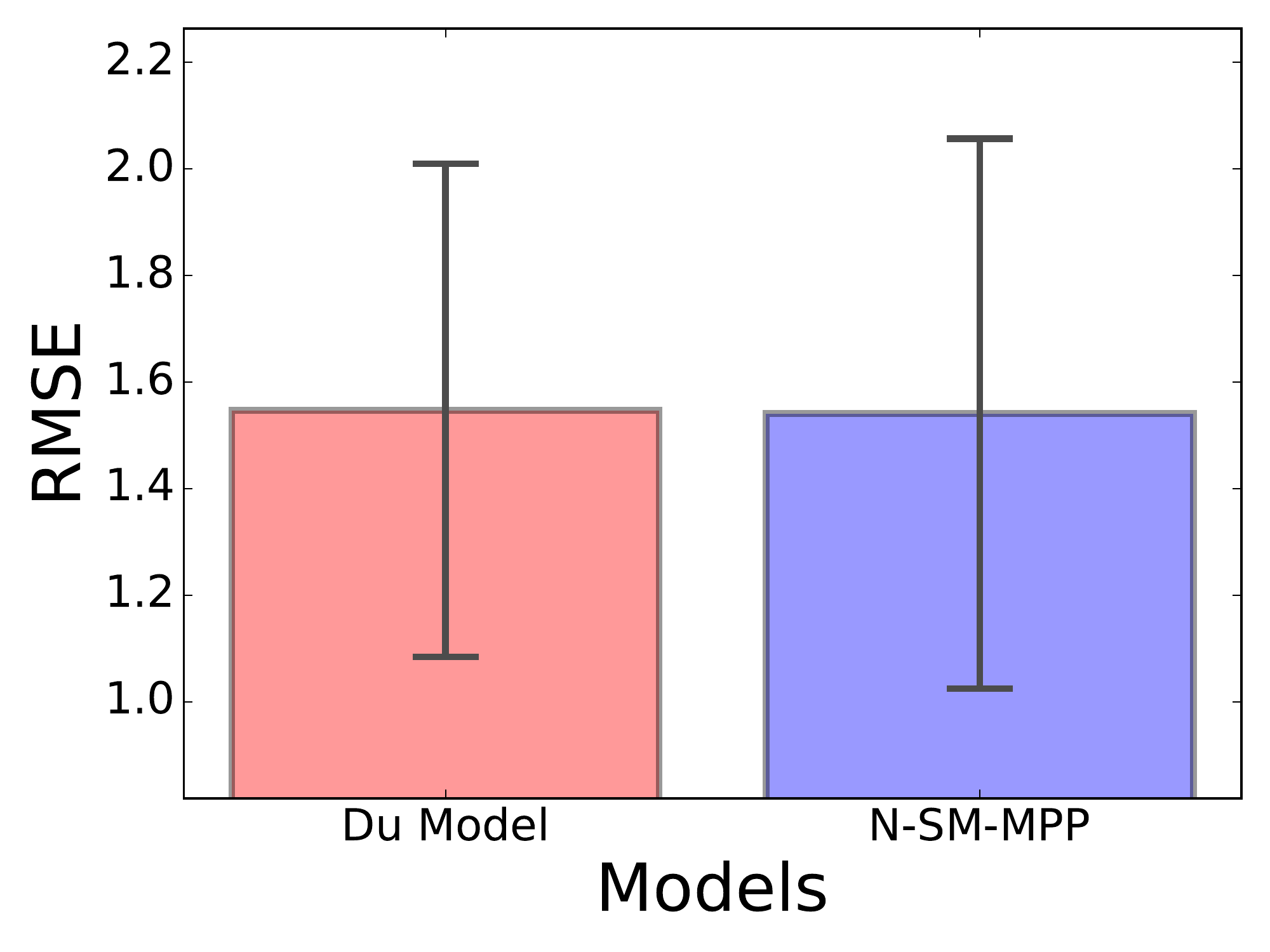}
	\includegraphics[width=0.32\textwidth]{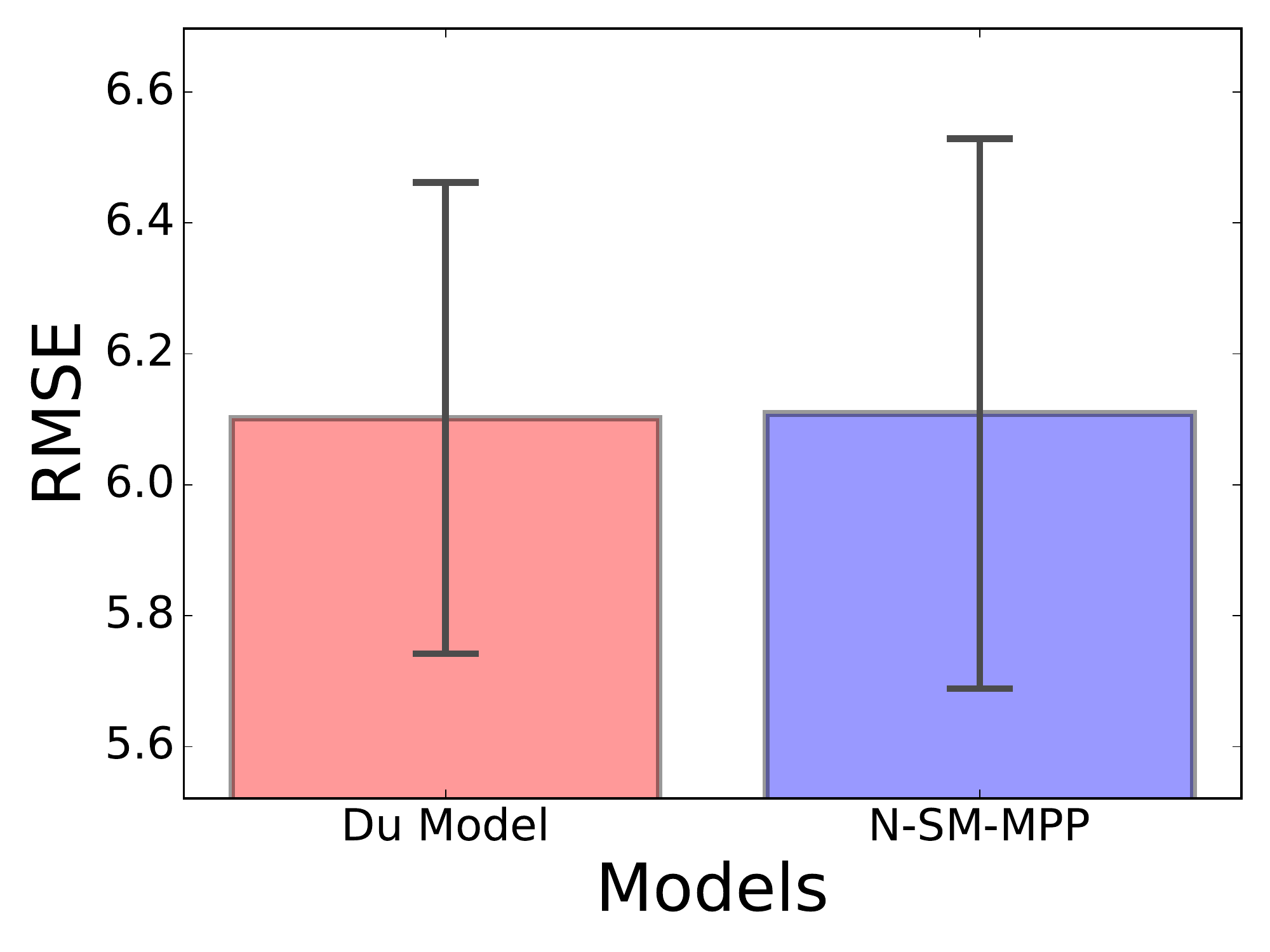}
	\includegraphics[width=0.32\textwidth]{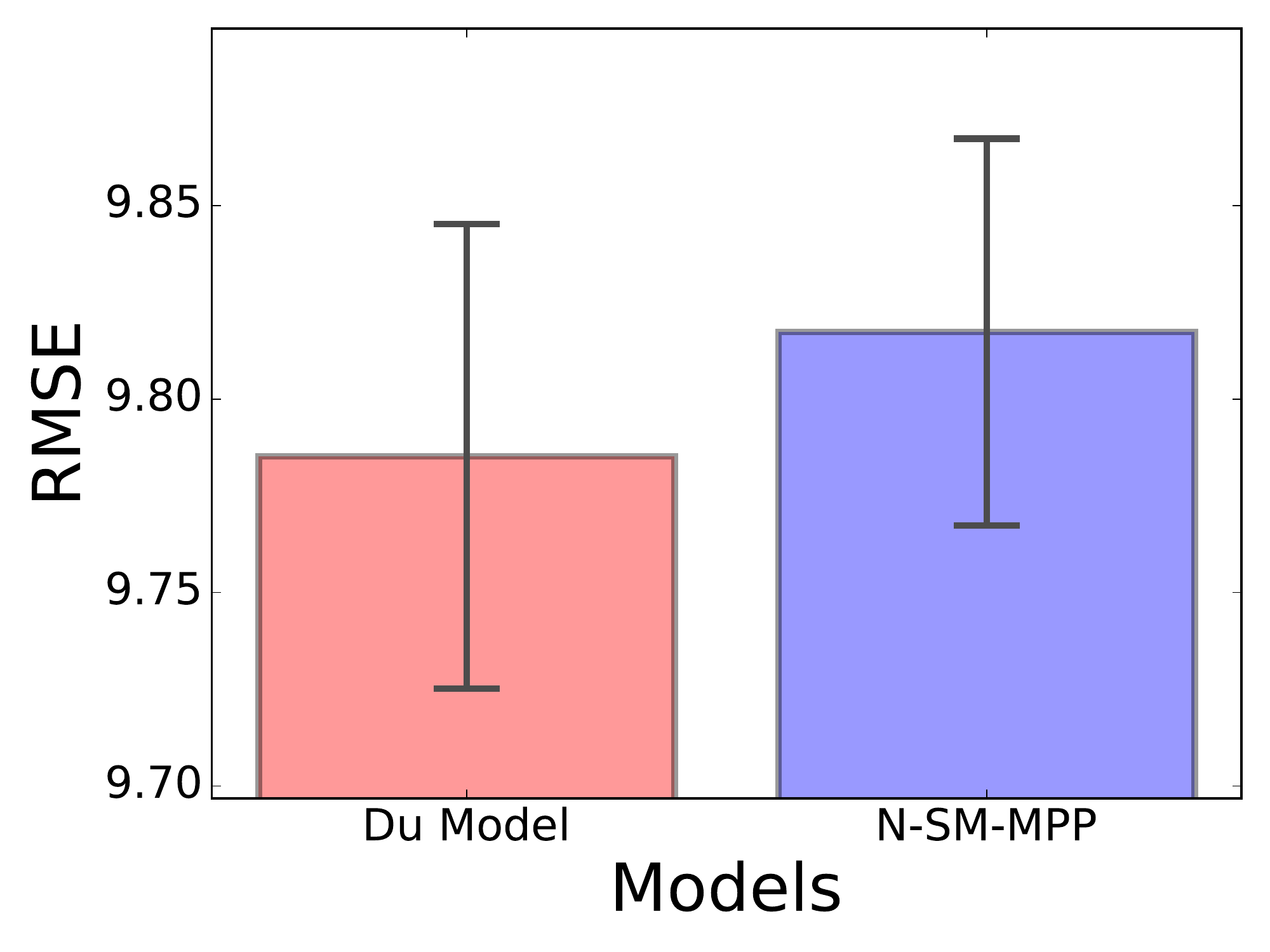}
\endminipage
\minipage[c]{0.26\textwidth}
\caption{
	Prediction results on Financial Transactions, MIMIC-II, and Stack Overflow datasets (from left to right). Error bars show standard deviation over 5 experiments with different train-dev-test splits.  For prediction of the types $k_i$ (top row), our method achieved lower error in 4/5, 5/5, and 5/5 of the experiments.  For prediction of the times $t_i$ (bottom row), our method achieved lower error in 5/5, 2/5, and 0/5 of the experiments.
}
\label{fig:prediction}
\endminipage
\end{center}
\end{figure}

\subsection{MemeTrack Dataset Details}
\label{sec:meme_details}
The MemeTrack dataset (\cref{sec:memetrack}) contains time-stamped instances of meme use in articles and posts from 1.5 million different blogs and news sites, spanning 10 months from August 2008 till May 2009, with several hundred million documents.

As in Retweets, we decline to model the appearance of novel memes.  Each novel meme serves as the \bos event for a stream of mentions on other websites, which we do model.  The $K$ event types correspond to the different websites. Given one meme's past trajectory across websites, our model must learn to predict how soon it will be mentioned again and where.

We used the version of the dataset processed by~\citet{gomez-13-structure}, which selected the top $5000$ websites in terms of the number of memes they mentioned.  We truncated sequences to a maximum length of 32, which affected only 1\% of them.  We randomly sampled disjoint train, dev and test sets with $32000$, $5000$ and $5000$ sequences respectively, treating them as before.

Because our current implementation does not allow for a marked \bos event (see \cref{sec:bos}), we currently ignore where the novel meme was originally posted, making the unfortunate assumption that the stream of websites is independent of the originating website.  Even worse, we must assume that the stream of websites is independent of the actual text of the meme.  However, as we see, our novel models have some ability to recover from these forms of missing data.

\begin{figure}
\begin{center}
\includegraphics[width=0.48\textwidth]{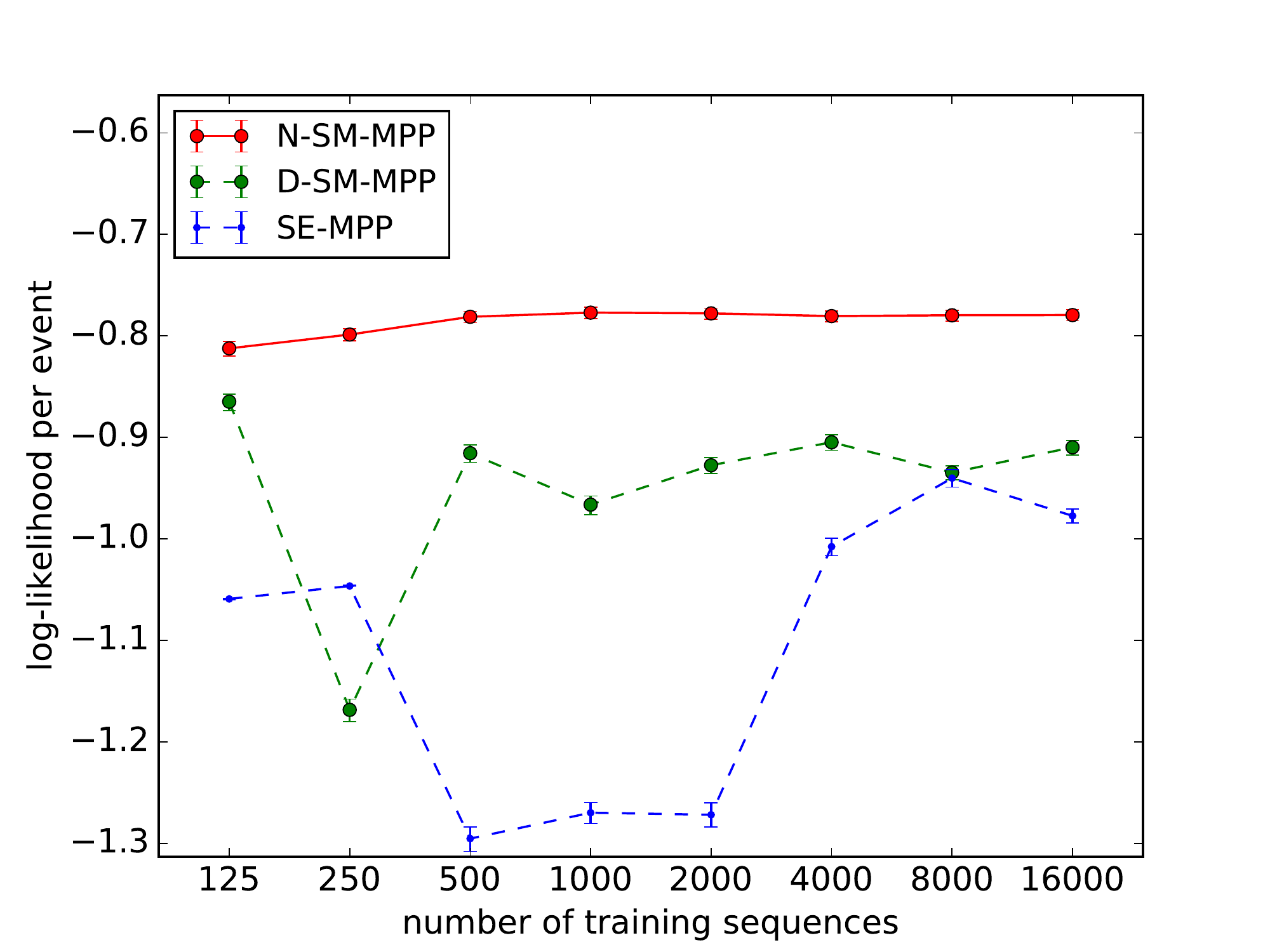}
\includegraphics[width=0.48\textwidth]{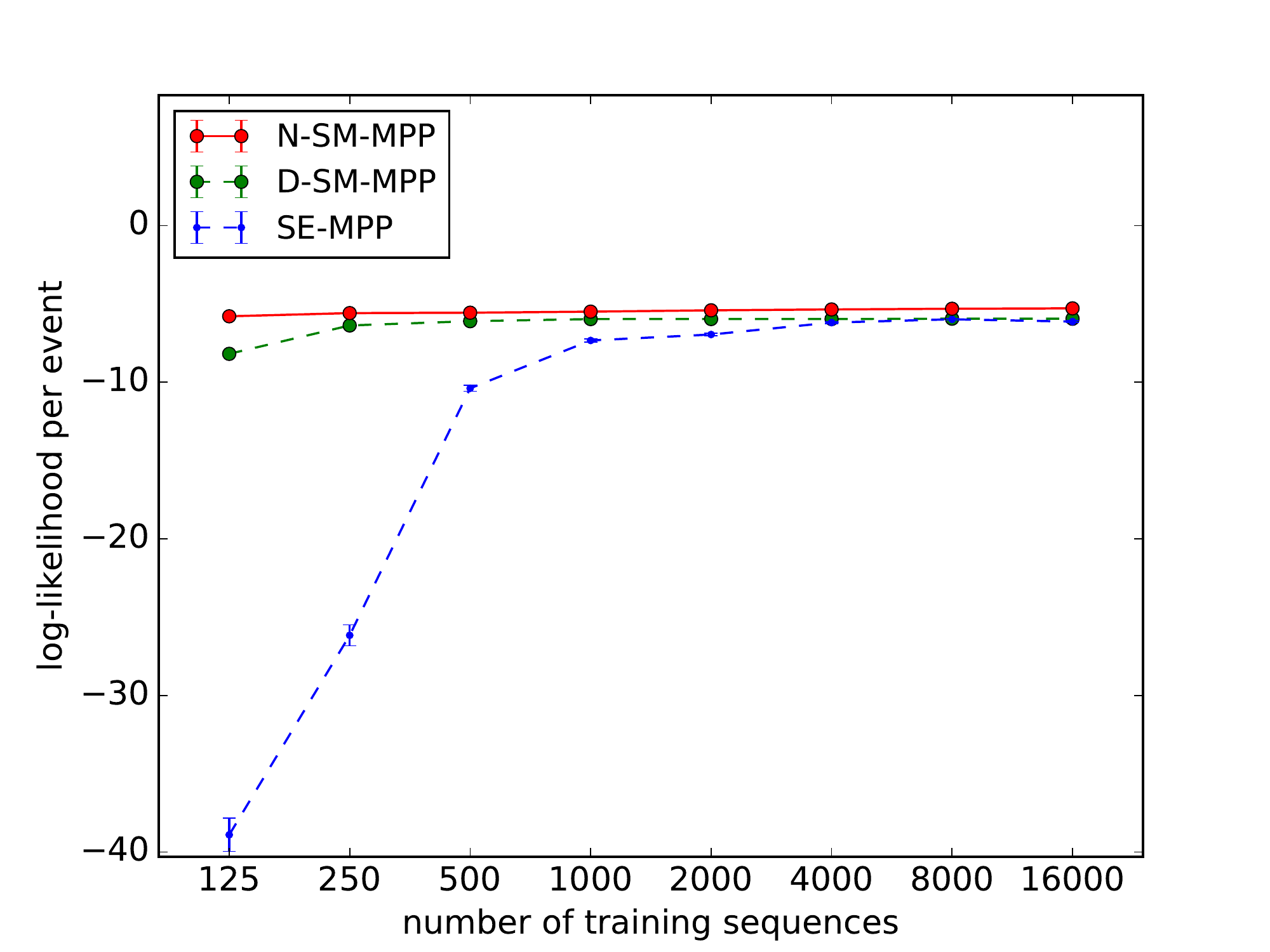}
\caption{
	Learning curves (with $95\%$ error bars) of all these models on the Retweets dataset, broken down by the log-probabilities of just the event types (left graph) and just the time intervals (right graph).
}
\label{fig:curve_retweet_breakdown}
\end{center}
\end{figure}

\begin{figure}
\begin{center}
\includegraphics[width=0.48\textwidth]{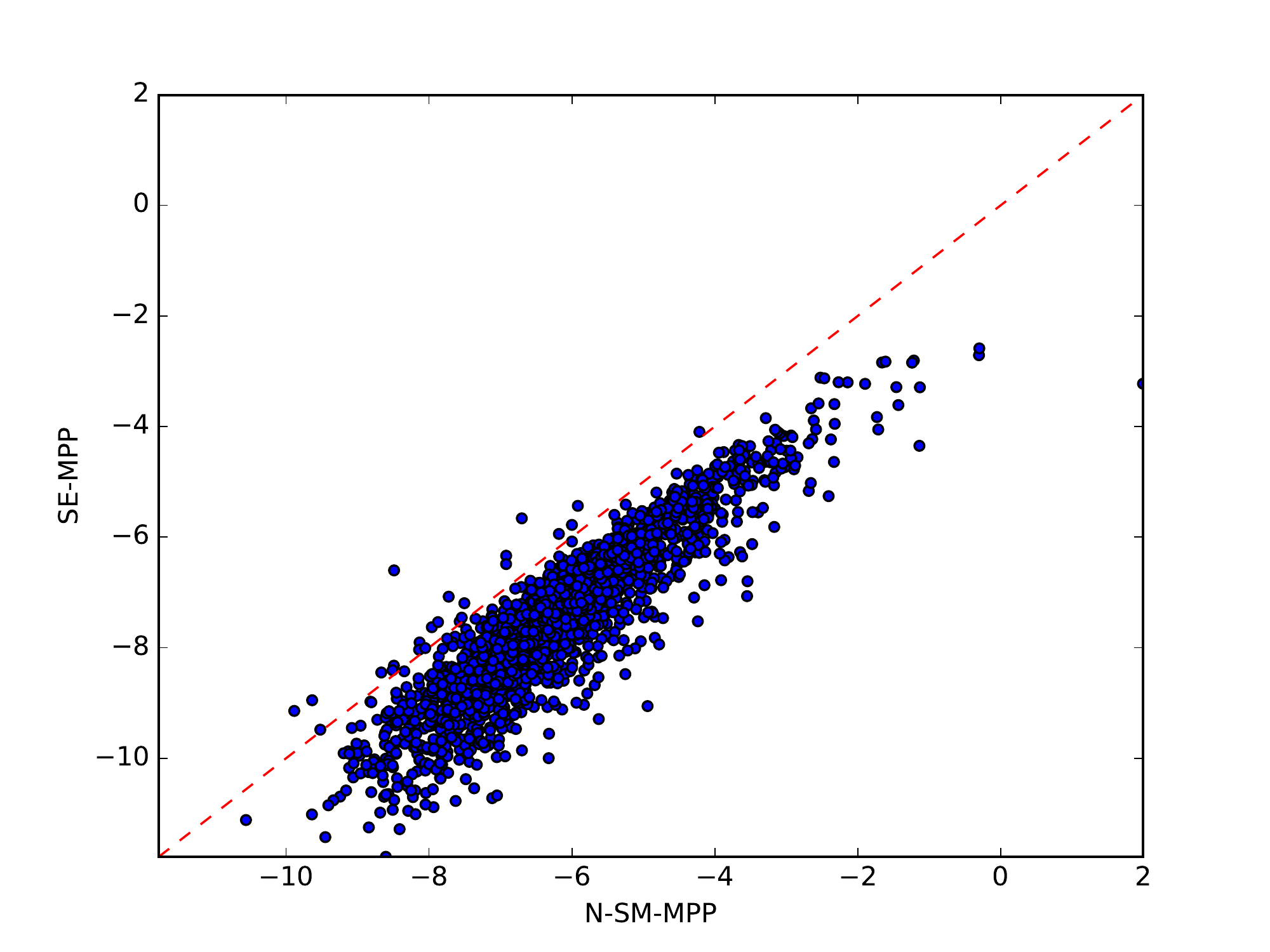}
\includegraphics[width=0.48\textwidth]{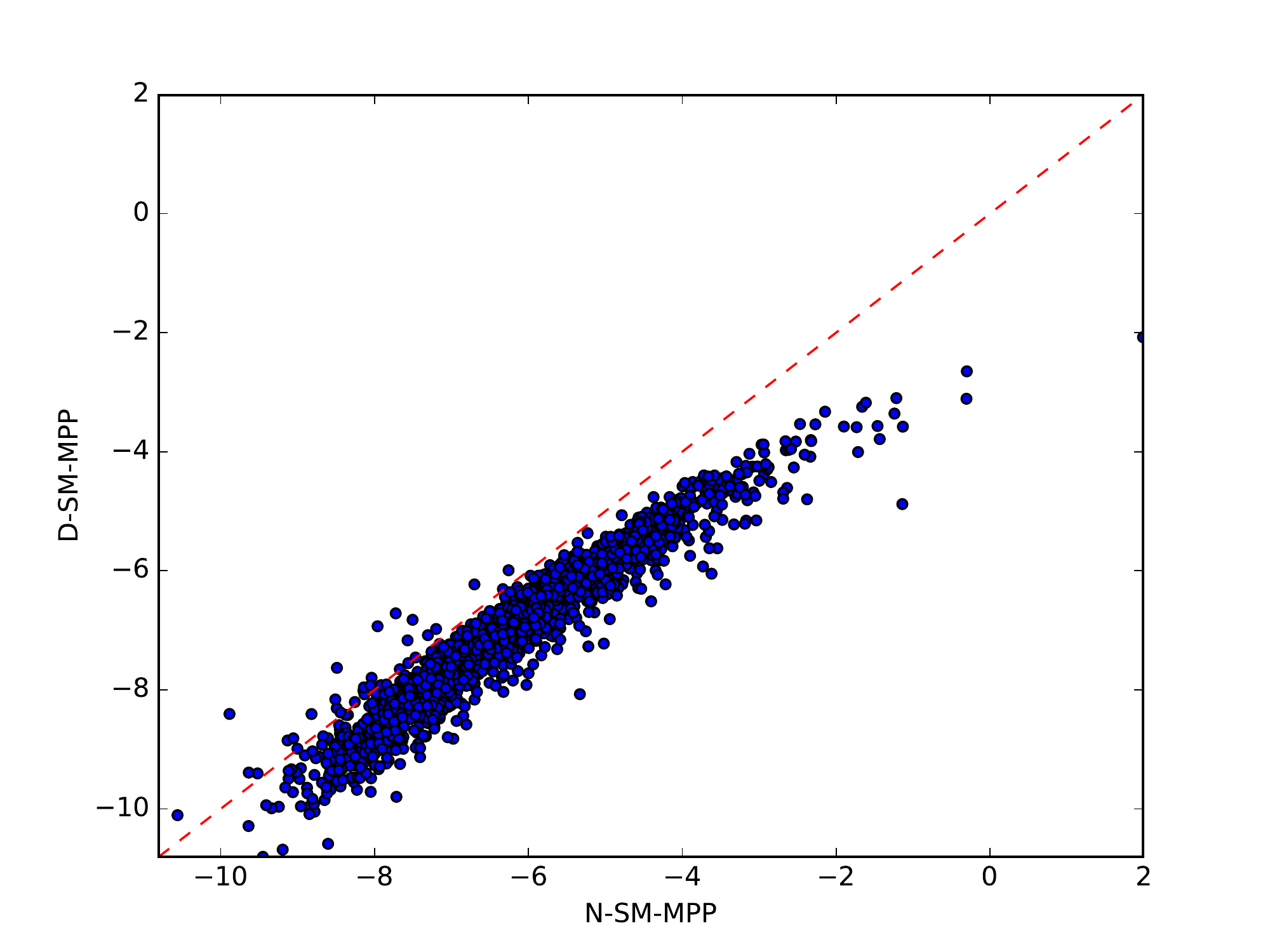}
\caption{A larger copy of \cref{fig:scatter_retweet}, repeated here for convenience.
}
\label{fig:scatter_retweet_copy}
\end{center}
\end{figure}

\begin{figure}
\begin{center}
\includegraphics[width=0.48\textwidth]{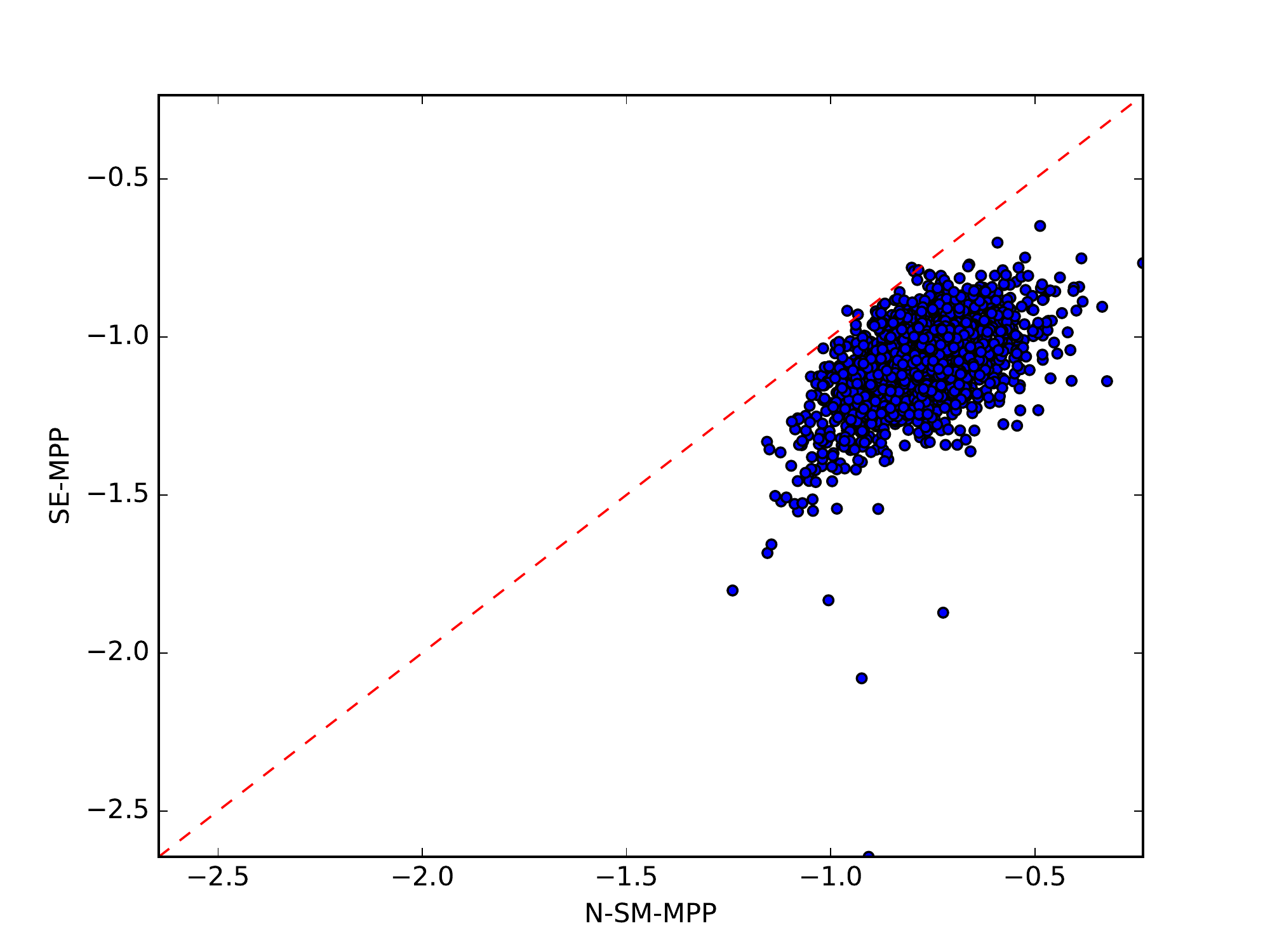}
\includegraphics[width=0.48\textwidth]{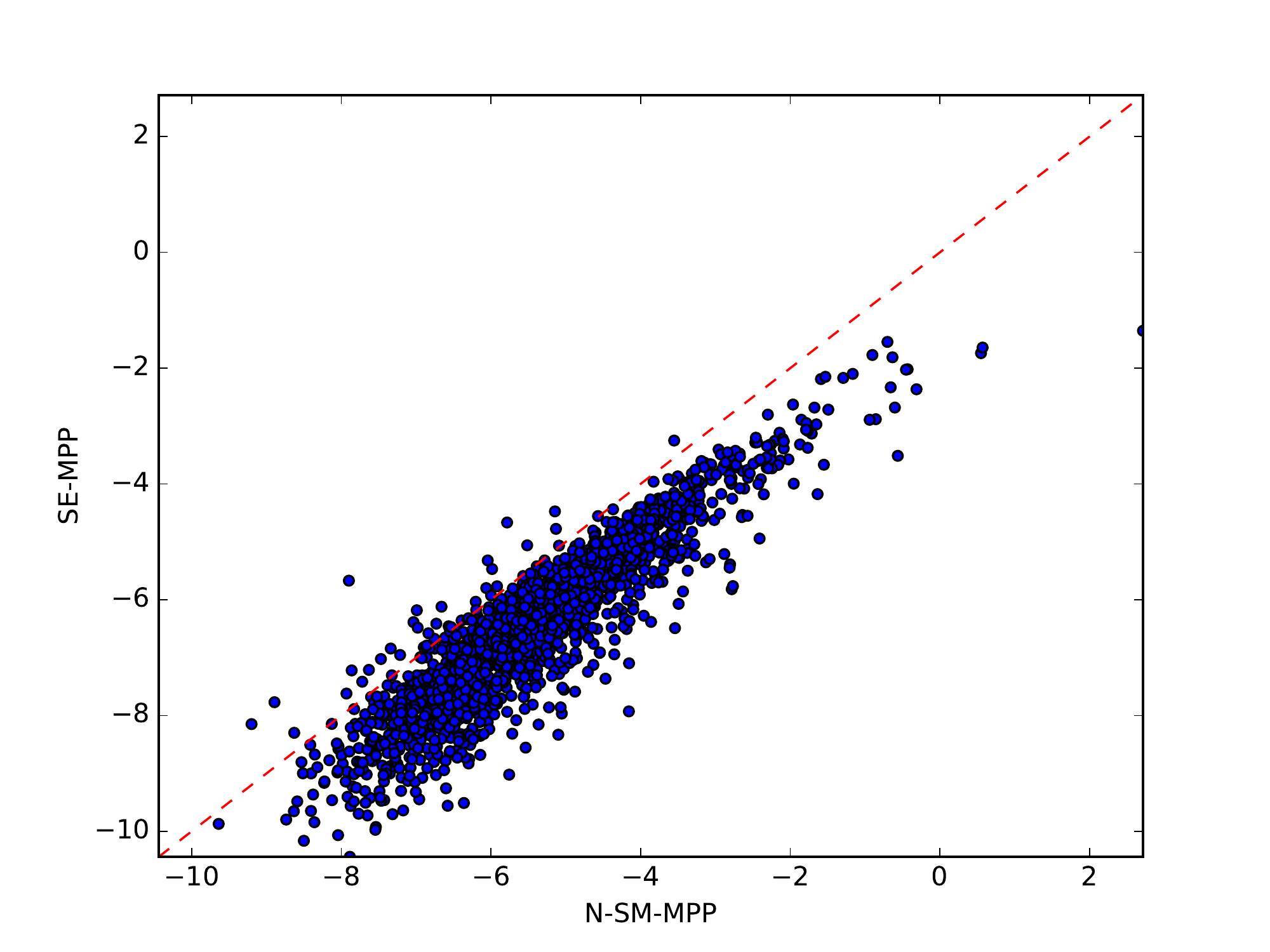}
\caption{
	Scatterplots of N-SM-MPP vs.\@ SE-MPP on Retweets. Same comparison
        as the left graph in \cref{fig:scatter_retweet_copy}, but broken down by the log-probabilities of the event types (left graph) and the time intervals (right graph).
}
\label{fig:scatter_retweet_with_hawkes_breakdown}
\end{center}
\end{figure}

\begin{figure}
\begin{center}
\includegraphics[width=0.48\textwidth]{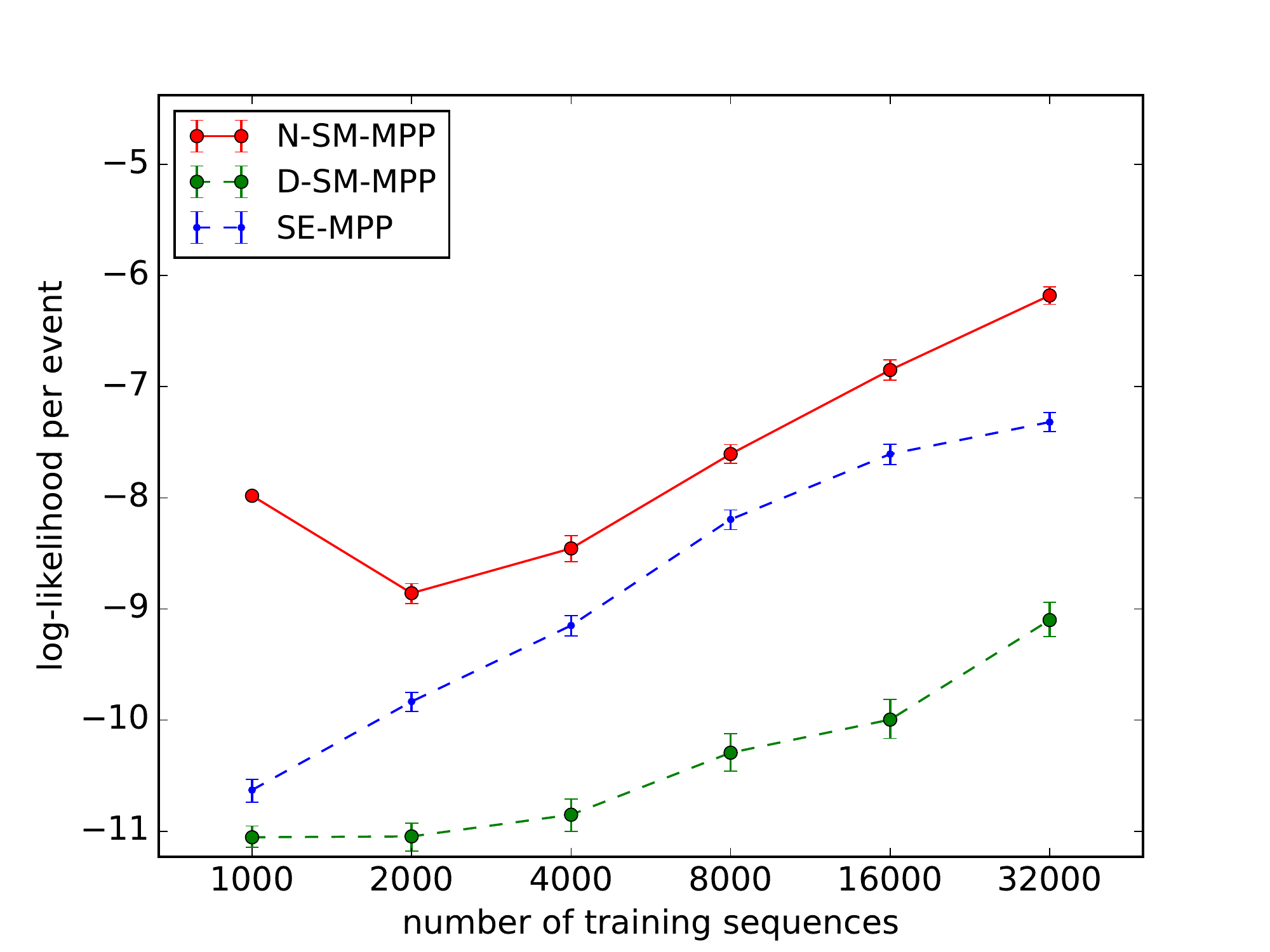}
\includegraphics[width=0.48\textwidth]{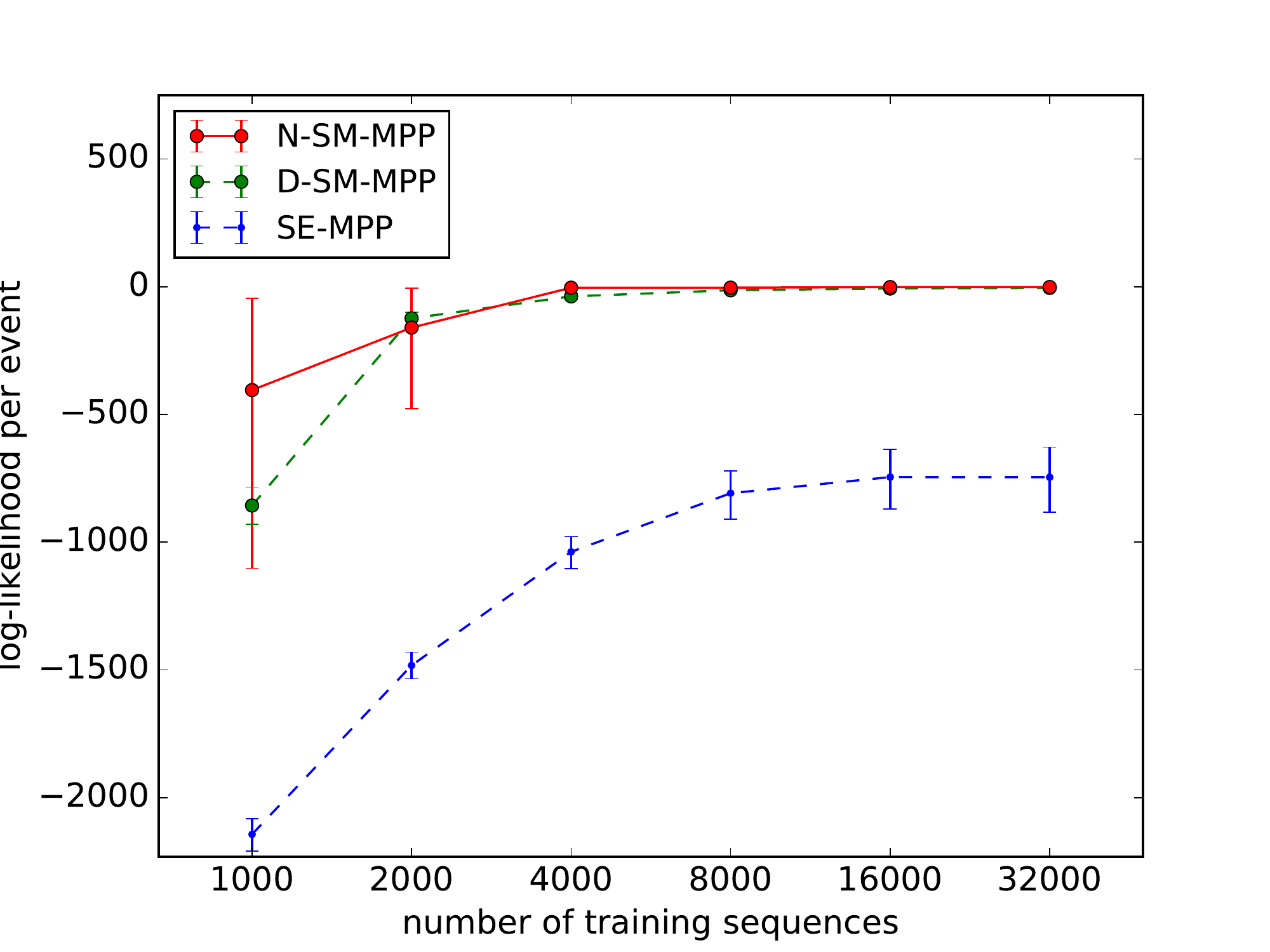}
\caption{
	Learning curve (with $95\%$ error bars) of all three models on the MemeTrack dataset, broken down by the log-probabilities of the event types (left graph) and the time intervals (right graph).
}
\label{fig:curve_meme_breakdown}
\end{center}
\end{figure}

\begin{figure}
\begin{center}
\includegraphics[width=0.48\textwidth]{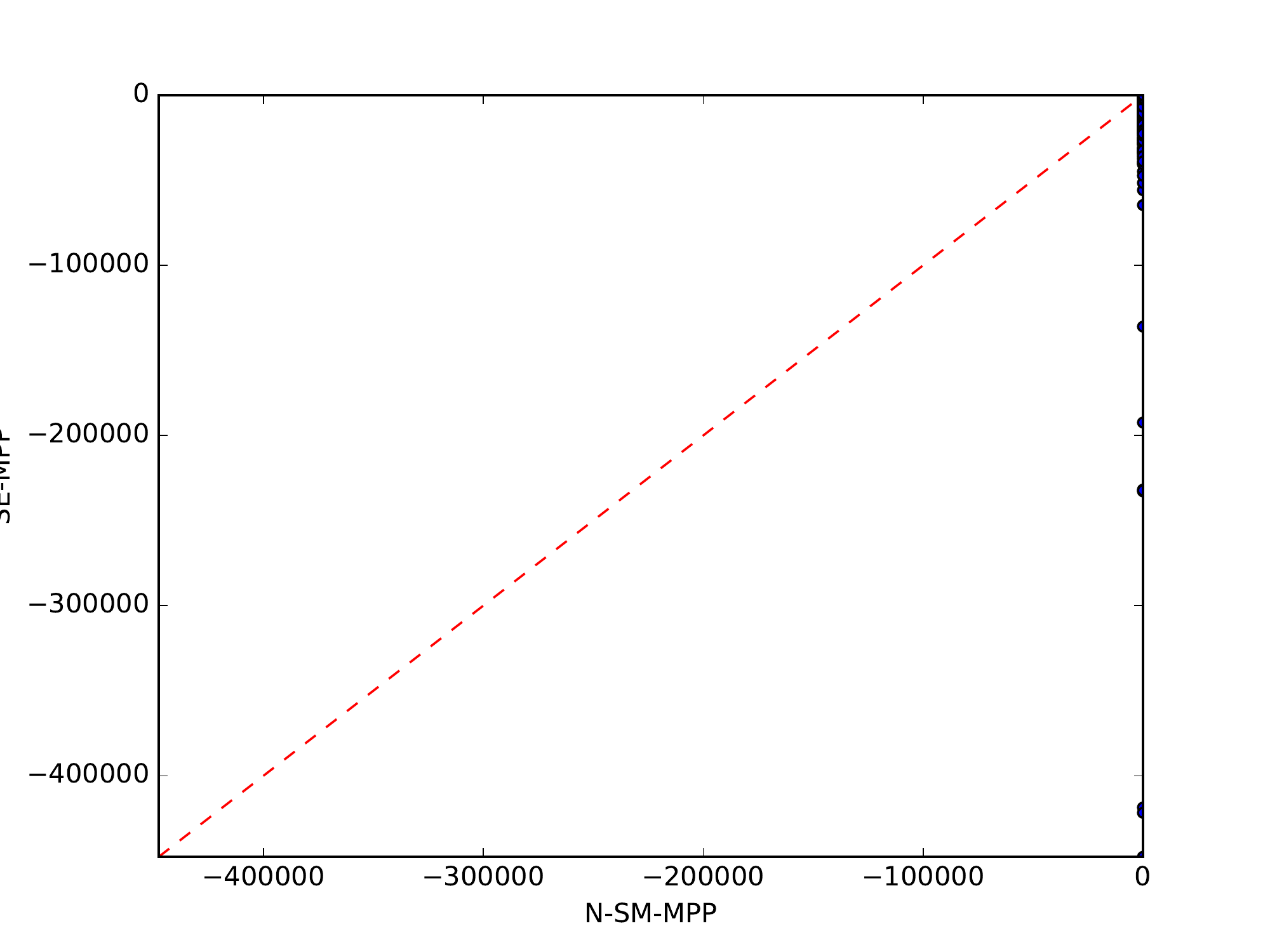}
\includegraphics[width=0.48\textwidth]{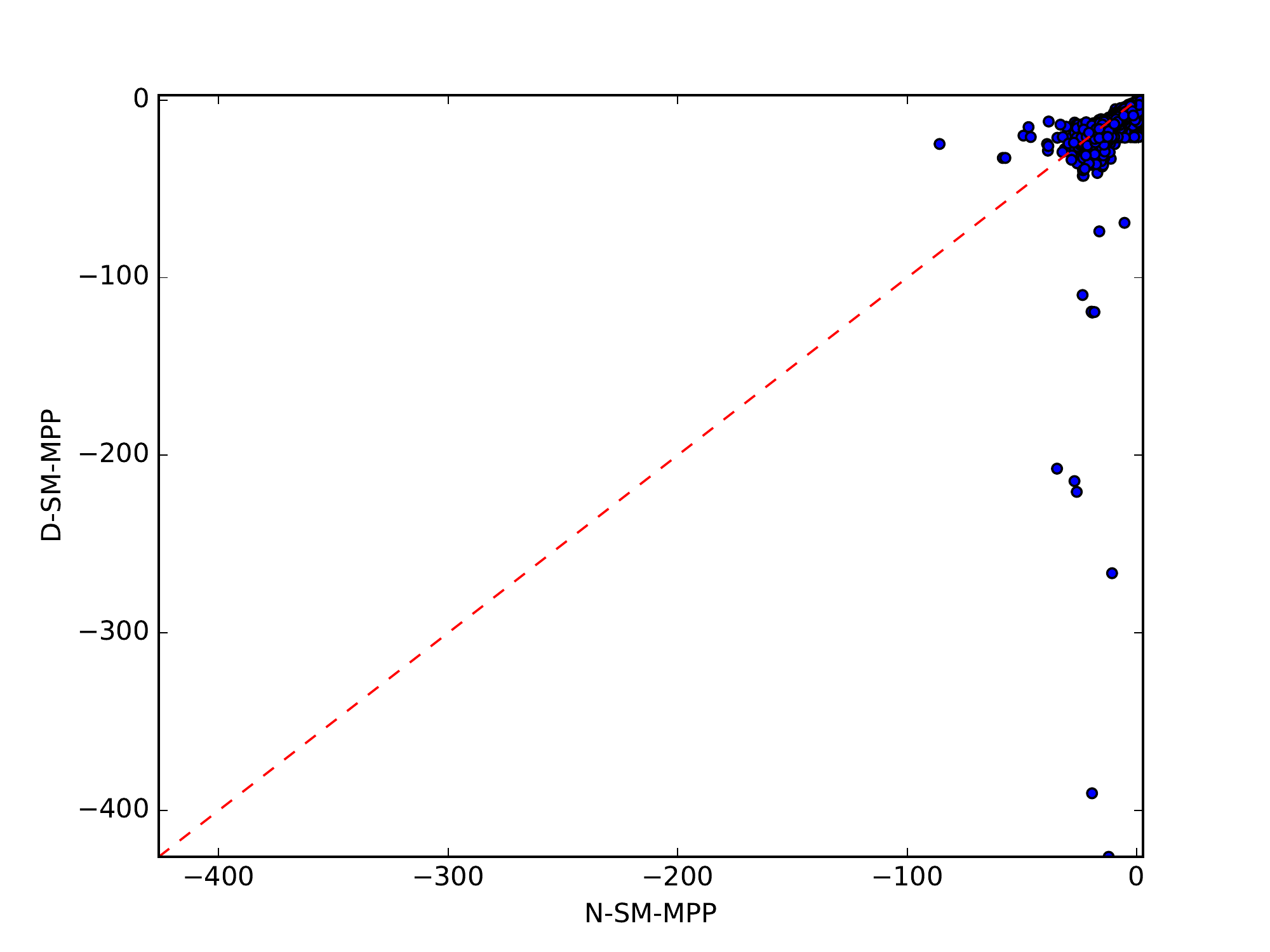}
\caption{
	Scatterplot of N-SM-MPP vs.\@ SE-MPP (left graph) and vs.\@ D-SM-MPP (right graph) on MemeTrack.
	N-SM-MPP outperforms D-SM-MPP on 93.02\% of the test sequences.  This is not obvious from the plot, because almost all of the 5000 points are crowded near the upper right corner.  Most of the visible points are outliers where N-SM-MPP performs unusually badly---and D-SM-MPP typically does even worse.
}
\label{fig:scatter_meme}
\end{center}
\end{figure}

\begin{figure}
\begin{center}
\includegraphics[width=0.48\textwidth]{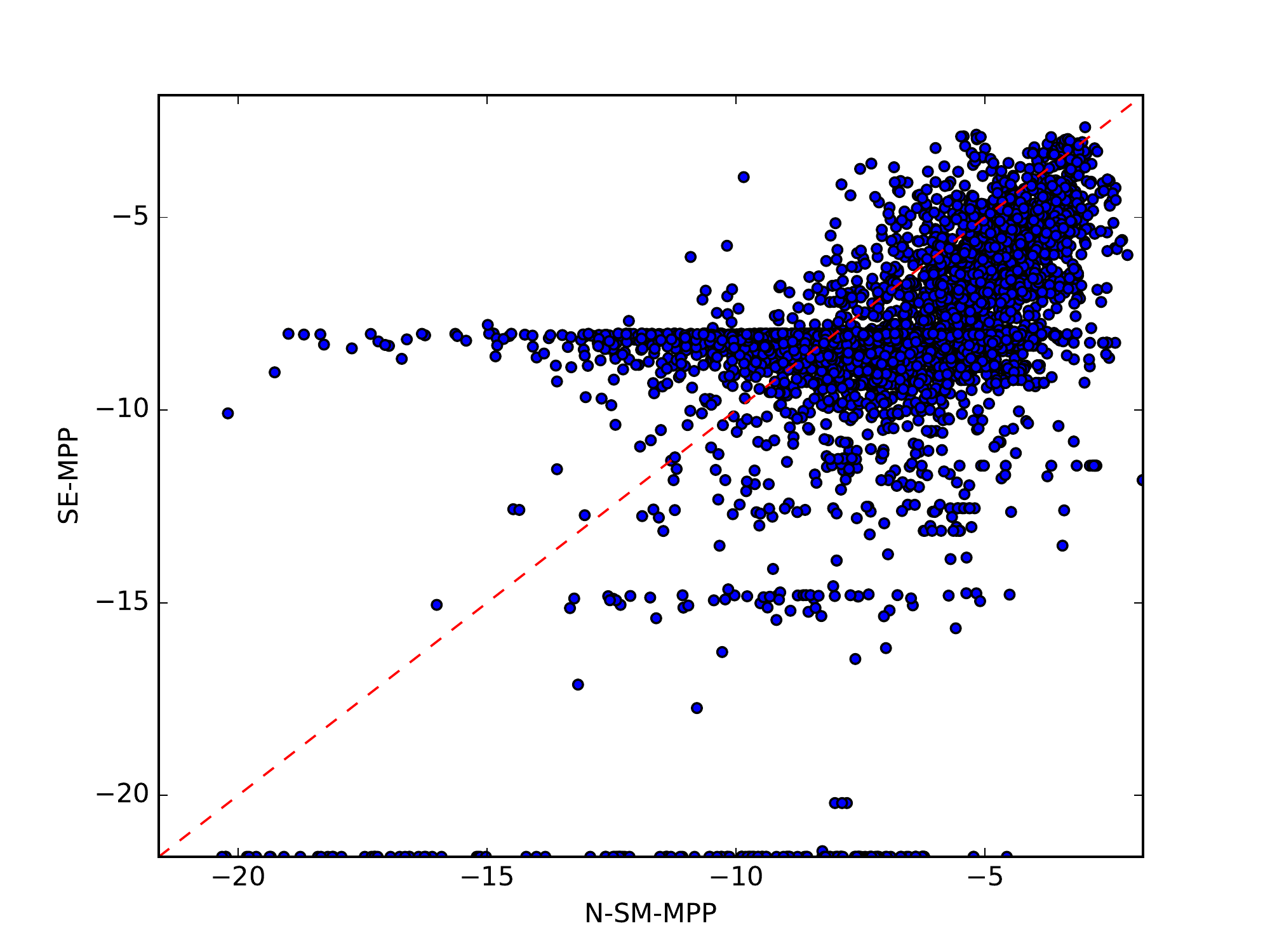}
\includegraphics[width=0.48\textwidth]{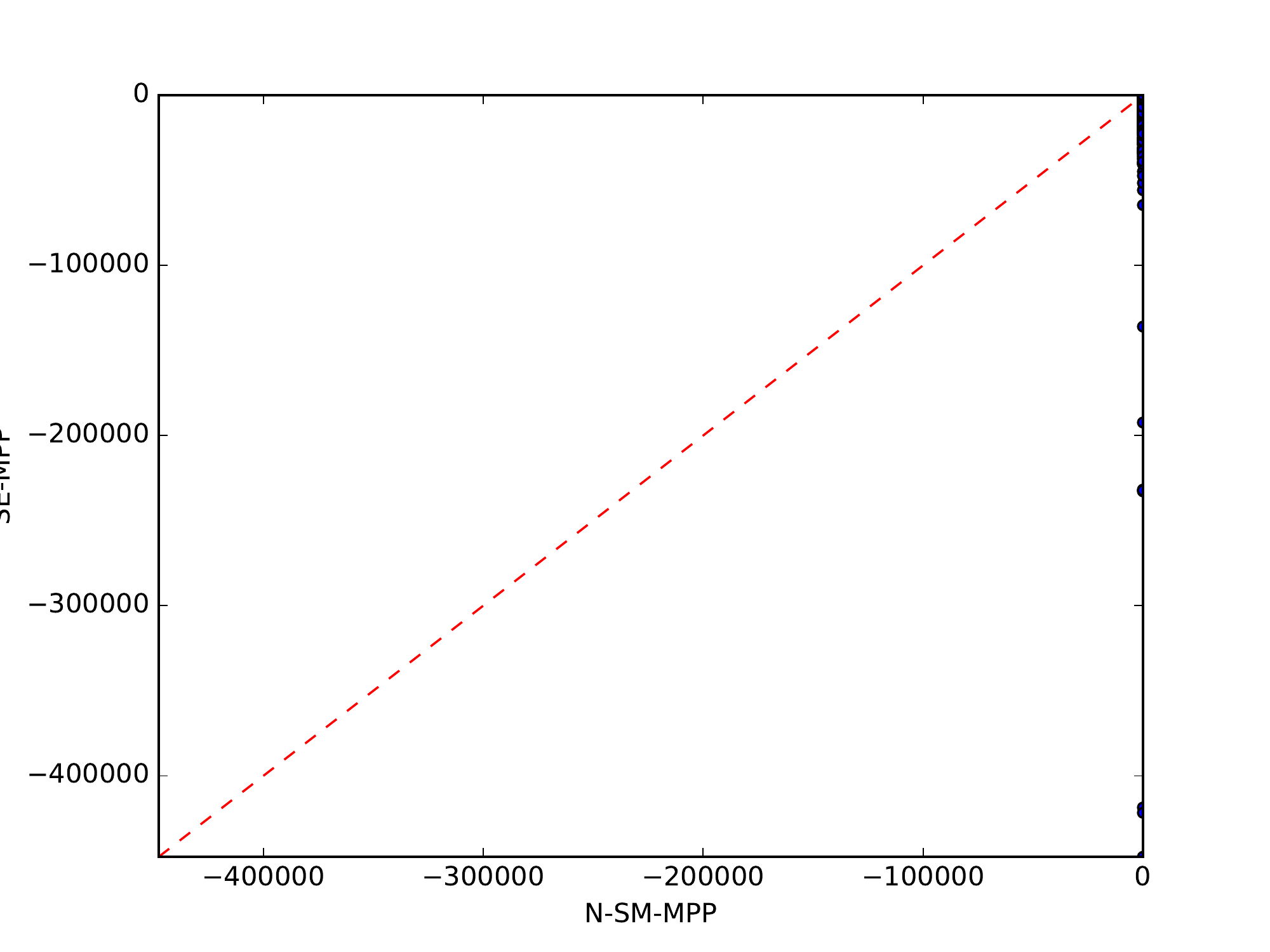}
\caption{
	Scatterplots of N-SM-MPP vs.\@ SE-MPP on MemeTrack.  Same comparison as the left graph of
        \cref{fig:scatter_meme}, but broken down by the log-probabilities of the event types (left graph) and the time intervals (right graph).
}
\label{fig:scatter_meme_with_hawkes_breakdown}
\end{center}
\end{figure}

\cref{fig:curve_meme_breakdown} shows the learning curves of the breakdown of log-likelihood with the same format as~\cref{fig:curve_retweet_breakdown}. \Cref{fig:scatter_meme,fig:scatter_meme_with_hawkes_breakdown} show the scatterplots in the same format as~\Cref{fig:scatter_retweet_copy,fig:scatter_retweet_with_hawkes_breakdown}.

\subsection{Prediction Task Details}
\label{sec:prediction_details}

Finally, we give further details of the prediction experiments from \cref{sec:prediction}.
To avoid tuning on the test data, we split the original training set into a new training set and a held-out dev set. We train our neural model and that of \citet{du-16-recurrent} on the new training set, and choose hyper-parameters on the held-out dev set.
Following \citet{du-16-recurrent}, we consider three datasets, and use five different train-dev-test splits of each dataset to generate the experimental results in~\cref{fig:prediction}.  (None of the test sets' examples were used during
manual development of our system.)

\clearpage

\clearpage

\clearpage
\section{Ongoing and Future Work}
\label{sec:future}
We are currently exploring several extensions to deal with more complex datasets. Based on our survey of existing datasets, we are particularly interested in handling:
\begin{itemize}
\item immediate events ($t_{i-1}=t_i$), as discussed in \cref{fn:immediate}
\item ``baskets'' of events (several events that are recorded as occuring simultaneously but without a specified order, e.g., the purchase of an entire shopping cart)
\item hard constraints on the event type sequence $k_1, k_2, \ldots$
\item marked events\footnote{\label{fn:mark}A ``mark'' is some structured data attached to an event: for example, the textual content associated with a tweet, or the medical records associated with a doctor visit.
    The model should predict the marks from each event and its underlying hidden state, and they should be fed back into the LSTM as additional input.} and annotated events\footnote{Humans may be asked to classify the events in an event stream or the relationships among its events.  Unlike marks, these annotations are not involved in the process that generates the event stream, and so are not fed into the LSTM as input.  Rather, they are assumed to be generated {\em post hoc} by the human from the entire observed stream---and may depend on the human's implicit reconstruction of the hidden states.  We can use any available annotations to help reconstruct the hidden states \citep{zaidan-08-modeling}, if we model them as stochastic functions of the hidden states.  In particular, annotations on the training data serve as side information to improve training of the model.  As a simple example, an
    annotation of the training event $(k_i,t_i)$ could be assumed to depend also on the subsequent LSTM state $\vec{h}(t_i^+) \defeq \lim_{t \rightarrow t_i^+} \vec{h}(t)$.}
\item causation by external events (artificial clock ticks, periodic holidays, weather)
\item richer drift functions\footnote{We expect the exponential drift in \cref{eqn:c_decay} to be expressive enough in most settings.  In principle, however, one might want to allow periodic fluctuation of the intensity between events, say by using a {\em complex} exponential in \eqref{eqn:c_decay}. Another way to increase expressivity would be to compute drift using the LSTM itself, by injecting special ``clock tick'' events into the input stream at regular intervals \citep[compare][]{xiao-17-joint}.  Each clock tick event $(k_i,t_i)$ causes a rich nonlinear update of the LSTM state via \crefrange{eqn:ct_lstm}{eqn:ct_cell}, except that it should always set $\vec{c}_{i+1} = \vec{c}(t_i)$ for continuity.  In this design, the interval between ordinary events is modeled piecewise---it is divided up into short pieces by the clock ticks, with $\vec{c}(t)$ on each piece modeled using our current function family.}
\item hybrid of D-SM-MPP and N-SM-MPP, allowing direct influence from past events
\item multiple agents each with their own state, who observe one another's actions (events)
\end{itemize}

More important, we are interested in modeling causality.  The current model might pick up that a hospital visit elevates the instantaneous probability of death, but this does not imply that a hospital visit {\em causes} death.  (In fact, the severity of an earlier illness is usually the cause of both.)

A model that can predict the result of interventions is called a causal model.  Our model family can naturally be used here: any choice of parameters defines a generative story that follows the arrow of time, which can be interpreted as a causal model in which patterns of earlier events {\em cause} later events to be more likely.  Such a causal model predicts how the distribution over futures would change if we intervened in the stream of events.

In general, one cannot determine the parameters of a causal model based on purely observational data~\citep{pearl-09-overview}.
Thus, in future, we plan to determine such parameters through randomized experiments by deploying our model family as an environment model within reinforcement learning.  A reinforcement learning agent {\em tests} the effect of random interventions to discover their effect (exploration) and thus orchestrate more rewarding futures (exploitation).\looseness=-1

In our setting, the agent is able to stochastically insert or suppress certain event types and observe the effect on subsequent events.  Then our LSTM-based model will discover the causal effects of such actions, and the reinforcement learner will discover what actions it can take to affect future reward. Ultimately this could be a vehicle for personalized medical decision-making.  Beyond the medical domain, a quantified-self smartphone app may intervene by displaying fine-grained advice on eating, sleeping, exercise, and travel; a charitable agency may intervene by sending a social worker to provide timely counseling or material support; a social media website may increase positive engagement by intelligently distributing posts; or a marketer may stimulate consumption by sending more targeted advertisements.
\end{document}